\documentclass[11 pt, oneside]{article}

\usepackage[title]{appendix}
\usepackage[pdftex]{graphicx}
\usepackage {subcaption}

\usepackage{setspace}

\usepackage[paperwidth=8.5in, paperheight=11in, top=1in, bottom=1in, left=1in, right=1in]{geometry}

\usepackage{endnotes}
\let\footnote=\endnote

\usepackage{geometry}
\usepackage{graphicx}
\usepackage{setspace}
\usepackage{natbib}
\usepackage{mathrsfs}
\usepackage{hyperref}
\usepackage{xcolor}
\hypersetup{
    colorlinks,
    linkcolor={black!50!black},
    citecolor={black!50!black},
    urlcolor={blue!80!black}
}

\usepackage[scaled=0.9]{helvet}

\DeclareMathSizes{10}{5}{1}{2}
\usepackage{booktabs}
\usepackage {amsbsy}
\usepackage {amssymb}
\usepackage {amsmath}
\usepackage{amsthm}
\usepackage{multirow}
\usepackage{algorithmic}
\usepackage{algorithm}
\usepackage{enumerate}
\usepackage[affil-sl,auth-sc]{authblk}
\usepackage{color}
\newcommand{\BFcomm}{\textcolor{black}}

\newtheorem{theorem}{Theorem}
\newtheorem{proposition}{Proposition}
\newtheorem{lemma}[theorem]{Lemma}

\usepackage{natbib}
 \bibpunct{(}{)}{;}{a}{,}{,}

\usepackage{color}

\newcommand{\comm}{\textcolor{black}}

\DeclareMathOperator*{\argmax}{arg\,max}

\date{}

\begin{document}

\title{Sparse Pseudo-input Local Kriging for Large Spatial Datasets with Exogenous Variables}

\author{Babak Farmanesh and Arash Pourhabib\thanks{School of Industrial Engineering and Management, Oklahoma State University, Stillwater, Oklahoma, \{babak.farmanesh,arash.pourhabib\}@okstate.edu}}


\maketitle

\begin{abstract}
We study large-scale spatial systems that contain exogenous variables, e.g. environmental factors that are significant predictors in spatial processes. Building predictive models for such processes is challenging because the large numbers of observations present makes it inefficient to apply full Kriging. In order to reduce computational complexity, this paper proposes Sparse Pseudo-input Local Kriging (SPLK), which utilizes hyperplanes to partition a domain into smaller subdomains and then applies a sparse approximation of the full Kriging to each subdomain. We also develop an optimization procedure to find the desired hyperplanes. To alleviate the problem of discontinuity in the global predictor, we impose continuity constraints on the boundaries of the neighboring subdomains. Furthermore, partitioning the domain into smaller subdomains makes it possible to use different parameter values for the covariance function in each region and, therefore, the heterogeneity in the data structure can be effectively captured. Numerical experiments demonstrate that SPLK outperforms, or is comparable to, the algorithms commonly applied to spatial datasets. 

\end{abstract}

\section {Introduction}\label{Sec_Intro}

Advances in data collection technologies for geostatistics have created unprecedented opportunities to build more effective data-driven models.  Of paramount importance in many engineering applications is to build predictive models for spatial processes that include environmental factors such as temperature or irrigation as \comm{significant predictors}~\citep{gao2014estimating,zhangextended}.  We call these environmental factors exogenous variables to distinguish them from simple spatial information such as latitude, longitude, and altitude. 


Kriging~\citep{cressie1990origins}, also known as Gaussian process regression (GPR)~\citep{Ras}, is a powerful tool for modeling spatial processes. Theoretically, GPR can benefit from very large datasets since it is a non-parametric model whose flexibility and performance generally increase by having more data points~\citep{hastie2009elements}. However, the computational complexity of GPR is dominated by the inversion of covariance matrices which is of $\mathcal{O}(N^3)$, where $N$ is the number of data points. 

To reduce GPR computation time, various approaches have been developed to approximate the covariance matrix of GP, resulting in less costly matrix operations. One class of such methods approximates the covariance matrix with sparse matrices, i.e., matrices with many zero elements~\citep{furrer2006covariance,zhang2008covariance,kaufman2008covariance}, and another class of methods finds low-rank approximations of the covariance matrix~\citep{williams2001using,smola2000sparse,Snl,PIC,quinonero2005unifying,pourhabib2014bayesian}. However, these methods do not directly take the heterogeneous structure for the data into account: if the behavior of the response variable strongly depends on the underlying geology~\citep{kim2005analyzing} or the exogenous variables, it is reasonable to assume different values for parameters of a given covariance function (hence, heterogeneity). Although there is a rich body of literature in spatial statistics that proposes different methods to capture inhomogeneous covariance structures (see~\cite{sampson1992nonparametric,schmidt2003bayesian,damian2003variance,paciorek2006spatial,lindgren2011explicit,fuglstad2015exploring}), the application of these methods is generally limited to small datasets with up to two-dimensional input spaces. As such, for large spatial datasets, and especially data with exogenous variables, it is beneficial to allow for different covariance parameters in each region, while addressing the computational challenge of handling the large number of observations.



In order to reduce the computational complexity of GPR, while at the same time improve its ability to tackle inhomogeneous covariance structures for large spatial datasets, one idea is to use a class of local Kriging that assumes distinct covariance functions for each region of the data domain. Local Kriging uses a partitioning policy that decomposes the domain into smaller subdomains and applies local GPR in each subdomain~\citep{haas1990kriging,Park,TGP}. Therefore, local Kriging reduces the total computational complexity to $O(Nn^2)$, where $n$ is the number of local data points, and $n\ll N$. This idea, however, presents two related challenges: discontinuity in prediction on the boundaries of the subdomains and devising an efficient partitioning policy.

To address discontinuity on the boundaries, one category of local Kriging methods uses various averaging techniques to smooth the prediction surface close to the boundaries. Examples in this category include Bayesian committee machine, BCM~\citep{tresp2000bayesian}, mixtures of Gaussian processes, MGP~\citep{rasmussen2002infinite}, treed Gaussian process models, TGP~\citep{urtasun2008sparse,TGP}, bagged Gaussian process, BGP~\citep{chen2009bagging}, and local probabilistic regression, LPR~\citep{urtasun2008sparse}. Such averaging techniques, however, come at the cost of higher computational complexity at prediction time.

Another category of methods to alleviate the discontinuity problem enforces continuity constraints on the boundaries of subdomains. This class of approaches includes domain decomposition method, DDM~\citep{Park}, patching local Gaussian processes, PGP~\citep{Park2016}, and patchwork Kriging, PWK~\citep{park2018patchwork}. Experimental studies suggest that directly imposing continuity constraints generally outperforms the averaging techniques~\citep{Park,Park2016,park2018patchwork}. However, due to the complexity of handling boundary conditions, only PWK can be applied to higher-dimensional domains; DDM and PGP are limited in practice to only two-dimensional domains~\citep{park2018patchwork}.

Furthermore, none of the local Kriging approaches above take the data structure, which is manifested in the covariance function, into account when partitioning the data domain: DDM and PGP use \emph{uniform mesh} that partitions the domain of the input data into rectangles. TGP and PWK, on the other hand, use simple tree based partitioning to iteratively bisect the input domain. Moreover, in order to obtain time efficient algorithms, the number of data points in each subdomain must be kept to a small value, e.g., up to 600 data points in each subdomain~\citep{Park,park2018patchwork}. However, there is a trade-off between the number of subdomains and the accuracy of prediction: as the number of subdomains, and thus boundaries, increases, the prediction accuracy on the boundaries of the subdomains decreases, regardless of the method used to handle the boundary conditions.

To address the limitations of existing local Kriging methods, this paper proposes a new method, Sparse Pseudo-input Local Kriging (SPLK), which utilizes covariance information to partition the data domain into subdomains. The data is partitioned using parallel hyperplanes, and continuity constraints are enforced on the boundaries of subdomains. This partitioning approach minimizes the number of boundaries and simplifies boundary conditions, allowing application to datasets with moderate dimensional spaces. We develop an optimization algorithm to find the desired hyperplanes that result in lower errors for the covariance approximations in each region, and provide theoretical justification for the use of such parallel hyperplanes to create the subdomains based on analysis of the covariance structure. Therefore, SPLK is essentially a hybrid method combining low-rank approximations and local GPR to seamlessly integrate a partitioning policy into local approximations to improve prediction accuracy.

One potential disadvantage to this proposed partitioning is that it can result in large-size subdomains, which makes the application of the full GPR in each subdomain computationally inefficient; this limitation is overcome by using covariance approximation methods for each region. This approximation also has the added benefit of increasing the flexibility of choosing the sizes of the subdomains to further reduce the number of boundaries. While SPLK has a higher computational complexity compared to sparse and low-rank approximation methods due to the handling of the boundary conditions, however, the use of local covariance functions in each subdomain better captures the heterogeneous data structures compared to low-rank approximation methods. Another trade-off is that since the covariance structure of a spatial process can vary in different directions, partitioning in one direction using parallel hyperplanes may not be the most flexible way of capturing such structures. Nonetheless, this simple partitioning of SPLK significantly reduces computation time over existing local Kriging methods while still maintaining acceptable prediction accuracy.

As the dimension of the data domain increases, handling the boundary conditions of SPLK becomes more computationally expensive due to the expansion of the boundary spaces. Therefore, we suggest applying SPLK to spatial datasets with a moderate number of exogenous variables. However, we note that the methodology is general and can be efficiently applied to any large dataset (on the order of hundreds of thousand of data points) with a small number of input variables (say ten or fewer). Our numerical studies demonstrate that SPLK outperforms, or performs as well as, the competing algorithms in terms of computation time or accuracy on two and three-dimensional spatial data, higher-dimensional spatial data with exogenous variables, and a nine-dimensional non-spatial data.



The remainder of this paper is organized as follows. Section~\ref{sec_GP} introduces GPR, and a few approximation techniques that are relevant to this paper. Section~\ref{sec_SPLK} explains the proposed method including domain partitioning, training local models subject to boundary conditions, and choosing directions of cuts. Section~\ref{sec_results} compares the proposed method to commonly used algorithms. Section~\ref{sec_summary} concludes the paper and suggests future research. The supplemental material includes proof of all theorems and other technical details and analyses related to the proposed approach.

\section {Gaussian Process Regression}\label{sec_GP}
Given an index set $\mathbf{T}$, a Gaussian Process (GP) is a stochastic process where for any $\mathbf{T}'=\{t_1,\ldots,t_N\}$ as a finite subset of $\mathbf T$, the random vector $[f_{t_1},\ldots,f_{t_N}]^T$ follows a multivariate normal distribution~\citep{Ras}, where $f_{t_i}$ is a realization of a measurable function $\mathcal{F}:\Omega\subset \mathbb{R}^p \rightarrow \mathbb{R}$ for a given $t_i$. Here, we consider the index set to be a subset of $\mathbb{R}^p$ such that for a given $\{\mathbf x_1,\mathbf x_2,\ldots,\mathbf x_N\}\in\mathbb{R}^p$, the random vector  $\mathbf{f}=[f_1,f_2,\ldots,f_N]^T$ follows a multivariate normal distribution, where $f_i=\mathcal{F}(\mathbf x_i)$ for all $i\in [N]$, and $[N]$ denotes the set of positive integers smaller than or equal to $N$, i.e., $[N]=\{1,\ldots,N\}$.

We say a GP is fully specified when the function $\mathcal{F}$ follows a GP distribution with mean function $\mathcal{M}(\cdot)$ and covariance function $\phi(\cdot,\cdot)$. In other words, given $\mathcal{M}(\cdot)$ and $\phi(\cdot,\cdot)$, the mean vector and the covariance matrix of random vector $\mathbf{f}$ can be calculated, i.e., $\boldsymbol{\mu}=\mathbb{E}(\mathbf f)$ and $\mathbf K=\mathbb{E}\big((\mathbf f-\boldsymbol{\mu})(\mathbf f-\boldsymbol{\mu})^T\big)$, where $\mathbb{E}\{\cdot\}$ denotes the expectation operator. This means that $\mu_i=\mathbb{E}(f_i)=\mathcal{M}(\mathbf x_j)$, $\mu_j=\mathbb{E}(f_j)=\mathcal{M}(\mathbf x_j)$, and $k_{ij}=\mathbb{E}\big((f_i-\mu_i)(f_j-\mu_j)\big)=\phi(\mathbf x_i,\mathbf x_j)$. 

In the context of regression, given a training dataset $\mathbf{D}=\{(\mathbf x_i,y_i)\mid i=1,..,N\}$ consisting of noise contaminated observations, i.e., $y_i=\mathcal{F}(\mathbf x_i)+\epsilon_i$, the Gaussian Process Regression (GPR) seeks $p(f_*|\mathbf y)$, the predictive distribution of $f_*$ at $\mathbf x_*$ given $\mathbf y=[y_1,y_2,\ldots,y_N]^T$. We can derive this predictive distribution directly from the definition of the GP using joint Gaussian distribution
\begin{eqnarray}
\mathbf [y,f_*]^T\sim\mathcal{N}\left(0,
\begin{bmatrix}
\mathbf K_{\mathbf{X}\mathbf{X}}+\sigma^2\mathbf I & \mathbf k_{\mathbf{X}\mathbf x_*}\\
\mathbf k_{\mathbf x_*\mathbf{X}}& k_{\mathbf x_*\mathbf x_*}
\end{bmatrix}
\right),\label{GPJointDist}
\end{eqnarray}
where $\mathbf K_{\mathbf{X}\mathbf{X}}$ is a $N\times N$ covariance matrix of pairwise elements in $\mathbf{X}=\{\mathbf x_1,\ldots,\mathbf x_N\}$, $\mathbf k_{\mathbf{X}\mathbf x_*}$ is a $N\times 1$ vector of covariances between $\mathbf{X}$ and $\mathbf x_*$, and $k_{\mathbf x_*\mathbf x_*}$ is the variance at $\mathbf x_*$. Hence, the GPR predictive distribution can be obtained by conditioning $f_*$ given $\mathbf y$ in~\eqref{GPJointDist}, 
\begin{eqnarray}
f_*|\mathbf y\sim\mathcal{N}\left(\mathbf k_{\mathbf x_*\mathbf{X}}(\mathbf K_{\mathbf{X}\mathbf{X}}+\sigma^2\mathbf I)^{-1} \mathbf y,k_{\mathbf x_*\mathbf x_*}-\mathbf k_{\mathbf x_*\mathbf{X}}(\mathbf K_{\mathbf{X}\mathbf{X}}+\sigma^2\mathbf I)^{-1}\mathbf k_{\mathbf{X}\mathbf x_*}\right).\label{eq:pred_dist}
 \end{eqnarray}

Since calculating~\eqref{eq:pred_dist} entails inverting matrices of size $N$, the computational complexity is of $\mathcal{O}(N^3)$, which is generally too slow for most practical applications, especially spatial statistics. Low-rank covariance approximation methods~\citep{williams2001using,quinonero2005unifying} approximate the original covariance matrix through the Nystr{\"o}m method,
\begin{eqnarray}
\mathbf K_{\mathbf{X}\mathbf{X}}\approx\tilde{\mathbf K}_{\mathbf{X}\mathbf{X}}= \mathbf K_{ \mathbf{X}\tilde{\mathbf{X}}}\mathbf K_{\tilde{\mathbf{X}}\tilde{\mathbf{X}}}^{-1}\mathbf K_{\tilde{\mathbf{X}}\mathbf{X}},\label{eq:Q}
\end{eqnarray}
where $\tilde{\mathbf{X}}$ is either a subset of $\{\mathbf x_1,\mathbf x_2,\ldots,\mathbf x_N\}$ or a set of unobserved \textit{pseudo-inputs}, which are a new set of parameters used to approximate the likelihood of GPR. In particular, Sparse Pseudo-input Gaussian Process (SPGP)~\citep{PIC} assumes that observations $\mathbf y$ are conditionally independent, given the \emph{pseudo-outputs} $\tilde{{\mathbf f}}= [\tilde{f}_1,\ldots,\tilde{f}_m]^T$ defined on pseudo-input set $\tilde{\mathbf{X}}=\{\tilde{\mathbf{x}}_1,\ldots,\tilde{\mathbf{x}}_m\}$. This implies the joint Gaussian likelihood,
\begin{eqnarray}
[\mathbf y,f_*]^T\sim \mathcal{N}\left(0,
\begin{bmatrix}
\tilde{\mathbf K}_{\mathbf{X}\mathbf{X}}+\text{diag}(\mathbf K_{\mathbf{X}\mathbf{X}}- \tilde{\mathbf K}_{\mathbf{X}\mathbf{X}})+\sigma^2 \mathbf I & \tilde{\mathbf k}_{\mathbf{X}\mathbf{x}_*}\\
\tilde{\mathbf k}_{\mathbf{x}_*\mathbf{X}}& k_{\mathbf{x}_*\mathbf{x}_*}
\end{bmatrix}
\right),\label{jointSPGP}
\end{eqnarray}
and the predictive mean and variance,
\begin{eqnarray}
\hat\mu(f_*|\mathbf y)=\tilde{\mathbf k}_{\mathbf{x}_*\mathbf{X}}(\tilde{\mathbf K}_{\mathbf{X}\mathbf{X}}+\text{diag}(\mathbf K_{\mathbf{X}\mathbf{X}}-\tilde{\mathbf K}_{\mathbf{X}\mathbf{X}})+\sigma^2 \mathbf I)^{-1}\mathbf y,\label{eq:SPGP_mean}\\
\hat\sigma^2(f_*| \mathbf y)=k_{\mathbf{x}_*\mathbf{x}_*}-(\tilde{\mathbf K}_{\mathbf{X}\mathbf{X}}+\text{diag}(\mathbf K_{\mathbf{X}\mathbf{X}}-\tilde{\mathbf K}_{\mathbf{X}\mathbf{X}})+\sigma^2 \mathbf I)^{-1}\tilde{\mathbf k}_{\mathbf{X}\mathbf{x}_*},\label{eq:SPGP_var}
\end{eqnarray}
where $\tilde{\mathbf k}_{\mathbf{X}\mathbf{x}_*}$ is the low-rank covariance vector between $\mathbf{X}$ and the test data point $\mathbf{x}_*$ calculated by~\eqref{eq:Q}. Section~\ref{sec_SPLK} explains how the low-rank approximation in SPGP helps us devise a simple but efficient partitioning policy for our proposed local Kriging method.

\section {Sparse Pseudo-input Local Kriging}\label{sec_SPLK}

This section describes our proposed method, Sparse Pseudo-input Local Kriging (SPLK), where we partition the domain of data into smaller subdomains with simple boundaries, train local predictors that utilize a low-rank covariance matrix in each subdomain, and  connect neighboring local predictors on their joint boundaries to obtain a continuous global predictor. To partition the input domain, we use parallel hyperplanes, i.e., $(p-1)$-dimensional linear spaces embedded in a $p$-dimensional space (see Section~\ref{sub_sec_numb_pd} for the details). This partitioning minimizes the number of boundaries, because for $S$  subdomains, we only need $S-1$ parallel hyperplanes regardless of the dimension of the input space. \comm{This reduction in the number of boundaries improves the prediction accuracy, since the accuracy of local models decreases in the regions close to the boundaries regardless of the way the boundary conditions are handled. Moreover, partitioning by parallel hyperplanes creates simple boundary conditions (see Section~\ref{sub_sec_mean_pred}), as each boundary is shared by exactly two subdomains. Hence, each boundary only requires two local predictors} However, the drawback is that the partitioning policy can result in very large subdomains, where a full GPR is computationally inefficient. We overcome this problem by using covariance approximation techniques that utilize pseudo-inputs.

Among the infinite possible ways to partition a domain by parallel hyperplanes, we seek those that improve the accuracy of local predictors, i.e., the covariance approximation in each subdomain has the smallest error. We present two theorems that together determine the policy for creating subdomains. We begin by presenting the local mean and variance calculations, assuming the subdomains have already been determined. Then we discuss justifications for the proposed parallel hyperplanes for creating subdomains. (See Appendix~\ref{sec_practical} for practical aspects of SPLK's implementation such as constructing hyperplanes, hyperparameter learning, and selection of control points).

Any partitioning policy that results in subdomains whose boundaries do not intersect, e.g., concentric hyperspheres, would benefit from having a small number of boundaries and simple boundary conditions. What makes parallel hyperplanes particularly appealing is the fact that the boundary spaces are minimal compared to any other non-intersectional partitioning policy. In addition, the simple structure of parallel hyperplanes allows us to analyze the direction of partitioning based on the underlying covariance structure; this might not be feasible in other partitioning policies.

\subsection{Mean and Variance Prediction}\label{sub_sec_mean_pred}

Let $\Omega\in\mathbb{R}^p$ denote the input domain, i.e., $\mathbf x\in\Omega$. We partition $\Omega$ into $S$ subdomains $\Omega_s$ for  $s\in [S]$ such that $\bigcup_{s=1}^S \Omega_s=\Omega$, and $\Omega_s \cap \Omega_{s'}=\phi$ for $s\neq s'$. We also denote $\mathbf{X}_s=\{\mathbf x_i\in \mathbf{X}\mid \mathbf x_i\in \Omega_s\}$ and $\mathbf{y}_s$ as the vector of observations corresponding to $\mathbf{X}_s$ (see Section~\ref{DirofCuts} for an explanation of determining $\Omega_s$).  
The partitioning scheme explained in Section~\ref{DirofCuts} and Appendix~\ref{sub_sec_numb_pd} can lead to subdomains containing a large number of training data points, which makes the application of a full GPR inefficient. Therefore, for each $\Omega_s$, we use $m_s$ local pseudo-inputs $\tilde{\mathbf{X}}_s=\{\tilde{\mathbf{x}}_1,\ldots,\tilde{\mathbf{x}}_{m_s}\}\in\Omega_s$ to form the local and low-rank covariance approximation,
\begin{eqnarray}
\tilde{\mathbf K}^s_{\mathbf{X}_s\mathbf{X}_s}=\mathbf K_{\mathbf{X}_s\tilde{\mathbf{X}}_s}\mathbf K_{\tilde{\mathbf{X}}_s\tilde{\mathbf{X}}_s}^{-1} \mathbf K_{\tilde{\mathbf{X}}_s\mathbf{X}_s}.\label{eq_low_rank_cov}
\end{eqnarray}

It is easy to verify that among all the linear predictors $\mu(f_*|\mathbf x_*)=\mathbf u(\mathbf x_*)^T\mathbf y$, where $\mathbf u(\mathbf x_*) \in \mathbb{R}^n$ and $[\mathbf y,f_*]^T$ follows distribution~\eqref{GPJointDist}, the GPR mean predictor minimizes the expected squared error, $\mathbb{E} \big((\mu (f_*|\mathbf{x}_*)-f_*)^2\big)$. We extend this idea to find the local and low-rank predictor for each subdomain by assuming that $[\mathbf y_s,f_*]^T$ follows the local version of SPGP's joint likelihood distribution~\eqref{jointSPGP}. As such, we solve 
\begin{eqnarray}
\begin{aligned}
& \min_{\mathbf u_s(\mathbf x_*)}
& &\mathbb{E} \big((\mathbf u_s(\mathbf x_*)^T\mathbf y_s-f_*)^2\big)  \\
& \text{subject to}
&&[\mathbf y_s,f_*]^T\sim \mathcal{N}\left(0,
\begin{bmatrix}
\tilde{\mathbf K}^s_{\mathbf{X}_s\mathbf{X}_s}+\text{diag}(\mathbf K_{\mathbf{X}_s\mathbf{X}_s}-\tilde{\mathbf K}^s_{\mathbf{X}_s\mathbf{X}_s})+\sigma_s^2 \mathbf I_s &  \tilde{\mathbf k}^s_{\mathbf{X}_s\mathbf{x}_*}\\
 \tilde{\mathbf k}^s_{\mathbf{X}_s\mathbf{x}_*}& k_{\mathbf{x}_*\mathbf{x}_*}
\end{bmatrix}
\right),
\end{aligned}\label{UncOpt}
\end{eqnarray}
where $\mathbf{u}_s(\mathbf{x}_*)$ is the local version of $\mathbf{u}(\mathbf{x}_*)$, $\tilde{\mathbf k}^s_{\mathbf{X}_s\mathbf{x}_*}$ is the covariance vector between the test data point $\mathbf{x}_* \in \Omega_s$ and $\mathbf{X}_s$ using low-rank approximation formula~\eqref{eq_low_rank_cov}. 
Expanding the objective function with respect to the constraint in~\eqref{UncOpt} and removing $k_{\mathbf{x}_*\mathbf{x}_*}$, which does not depend on $\mathbf{u}_s(\mathbf{x}_*)$, results in the unconstrained optimization problem for each $\Omega_s$,
\begin{eqnarray}
\min_{ \mathbf{u}_s(\mathbf{x}_*)}\;\;\;\mathscr{M}_s=\mathbf{u}_s(\mathbf{x}_*)^T(\tilde{\mathbf K}^s_{\mathbf{X}_s\mathbf{X}_s}+\text{diag}(\mathbf K_{\mathbf{X}_s\mathbf{X}_s}-\tilde{\mathbf K}^s_{\mathbf{X}_s\mathbf{X}_s})+\sigma_s^2 \mathbf I_s) \mathbf{u}_s(\mathbf{x}_*)-2 \mathbf{u}_s(\mathbf{x}_*)^T \tilde{\mathbf k}^s_{\mathbf{X}_s\mathbf{x}_*}.\label{eq:LocalOBJ}
\end{eqnarray}
Note that  setting $\frac{d \mathscr{M}_s}{d \mathbf{u}_s(\mathbf{x}_*)
} =0$ gives $\mathbf u_s^{\text{opt}}(\mathbf{x}_*)=(\tilde{\mathbf K}_{\mathbf{X}_s\mathbf{X}_s}+\text{diag}(\mathbf K_{\mathbf{X}_s\mathbf{X}_s}- \tilde{\mathbf K}_{\mathbf{X}_s\mathbf{X}_s})+\sigma^2\mathbf I) ^{-1}\tilde{\mathbf k}_{\mathbf{X}_s\mathbf{x}_*}$, which is the SPGP's mean predictor for subdomain $\Omega_s$.  Next, we modify the optimization problem to alleviate the problem of discontinuity in the predictions on the boundaries. 

To impose continuity on the boundaries, we use a small number of \textit{control points} on the boundaries of each subdomain~\citep{park2018patchwork}. Let $\mathbf{B}_s$ be the set of all the control points located on the boundaries of $\Omega_{s}$. We intend to force local predictor $\mathbf{u}_s(\mathbf{x}^*)^T\mathbf{y}_s$ to be equal to the boundary values at the control point locations in $\mathbf{B}_s$,
\begin{eqnarray}
\mathbf{u}_s(\mathbf{b}_i)^T\mathbf{y}_s= \mathcal{R}(\mathbf{b}_i) \;\;\; \forall \mathbf{b}_{i} \in \mathbf{B}_s,\label{eq:ContCons}
\end{eqnarray}
where $\mathcal{R}(\mathbf{b}_{i})$ is a function that evaluates each $\mathbf{b}_{i}$ (see Section~\ref{sub_sec_numb_pd} for the details). Adding constraints~\eqref{eq:ContCons} to local model~\eqref{eq:LocalOBJ} gives the constrained local optimization for each $\Omega_s$,
\begin{eqnarray}
\begin{aligned}
& \underset{\mathbf{u}_s(\mathbf{x}_*)}{\text{min}}
& & \mathscr{M}_s=\mathbf{u}_s(\mathbf{x}_*)^T(\tilde{\mathbf K}^s_{\mathbf{X}_s\mathbf{X}_s}+\text{diag}(\mathbf K_{\mathbf{X}_s\mathbf{X}_s}-\tilde{\mathbf K}^s_{\mathbf{X}_s\mathbf{X}_s})+\sigma_s^2 \mathbf I_s) \mathbf{u}_s(\mathbf{x}_*)-2 \mathbf{u}_s(\mathbf{x}_*)^T\tilde{\mathbf k}^s_{\mathbf{X}_s\mathbf{x}_*} \\
& \text{subject to}
& & \mathbf{u}_s(\mathbf{b}_i)^T\mathbf{y}_s= \mathcal{R}(\mathbf{b}_i) \;\;\; \forall \mathbf{b}_{i} \in \mathbf{B}_s. \label{SPLkOptProb}
\end{aligned}
\end{eqnarray}

The objective function in optimization problem~\eqref{SPLkOptProb} is convex. Considering that the constraints are affine functions, we can solve optimization problem~\eqref{SPLkOptProb} analytically by transforming it into an unconstrained optimization problem using Lagrange duality principle~\citep{bazaraa2013nonlinear}. \comm{Appendix~\ref{SolutionOfLagrangianAPX} in the supplemental material presents the solution procedure.}

Solving optimization problem~\eqref{SPLkOptProb} obtains the optimal solution as
\begin{eqnarray}
\mathbf{u}_s^*(\mathbf{x}_*)=\mathbf{G}_s^{-1}(\tilde{\mathbf k}^s_{\mathbf{X}_s\mathbf{x}_*}+\mathbf{w}_s),
\end{eqnarray}
where $\mathbf{w}_s=0.5(\tilde{\mathbf k}^s_{\mathbf{x}_*\mathbf{X}_s}\tilde{\mathbf k}^s_{\mathbf{X}_s\mathbf{x}_*})^{-1}\mathbf{y}_s \tilde{\mathbf k}^s_{\mathbf{x}_*\mathbf{B}_s} \boldsymbol{\beta}_{s} \tilde{\mathbf K}^s_{\mathbf{B}_s\mathbf{X}_s}\tilde{\mathbf k}^s_{\mathbf{X}_s\mathbf{x}_*}$ and $\mathbf{G}_s=(\tilde{\mathbf K}^s_{\mathbf{X}_s\mathbf{X}_s}+\text{diag}(\mathbf K_{\mathbf{X}_s\mathbf{X}_s}-\tilde{\mathbf K}^s_{\mathbf{X}_s\mathbf{X}_s})+\sigma_s^2 \mathbf I_s)$.
Therefore, the local mean predictor for $\Omega_s$ becomes
\begin {eqnarray}
\hat\mu_s(f_*|\mathbf{x}_*)=\mathbf{u}^*_s(\mathbf{x}_*)^T\mathbf{y}_s= \tilde{\mathbf k}^s_{\mathbf{x}_*\mathbf{X}_s}\mathbf{G}_s^{-1}\mathbf{y}_s +\mathbf{w}_s^T\mathbf{G}_s^{-1}\mathbf{y}_s.\label{eq:PredMean}
\end{eqnarray}

Also, the objective function of local problem~\eqref{eq:LocalOBJ} is in fact the local variance predictor. Therefore plugging $\mathbf{u}_s^*(\mathbf{x}_*)$ into~\eqref{eq:LocalOBJ} obtains the predictive variance for $\Omega_s$, 
\begin{eqnarray}
&&\hat\sigma_s^2(f_*|\mathbf{x}_*)=k_{\mathbf{x_*}\mathbf{x_*}}- \tilde{\mathbf k}^s_{\mathbf{x}_*\mathbf{X}_s} \mathbf{G}_s^{-1}\tilde{\mathbf k}^s_{\mathbf{X}_s\mathbf{x}_*}\label{eq:PredVar}\\
&&+\tilde{\mathbf k}^j_{\mathbf{x}_*\mathbf{X}_s} \mathbf{G}_s^{-1}\mathbf{w}_s+\mathbf{w}_s^T\mathbf{G}_s^{-1}\mathbf{w}_s-\mathbf{w}_s^T\mathbf{G}_s^{-1}\tilde{\mathbf k}^s_{\mathbf{X}_s\mathbf{x}_*},\nonumber
\end{eqnarray}
where $k_{\mathbf{x_*}\mathbf{x_*}}$ is the constant initially removed from the optimization. Note that in both~\eqref{eq:PredMean} and~\eqref{eq:PredVar}, the first term is exactly the predictive mean and variance of local SPGP, and the following terms, which are amplified for local points close to the boundaries, appear to maintain the continuity of the global predictive function.

\subsection{Subdomain selection}\label{DirofCuts}
As mentioned in Section~\ref{sec_SPLK}, for computational efficiency we only consider the subdomains that are separated by parallel hyperplanes. We call these hyperplanes ``cutting hyperplanes,'' because each of them partitions or ``cuts'' $\Omega$ into two non-overlapping sets on different sides of the hyperplane. However, there are infinite ways of choosing the directions of the cutting hyperplanes. In Proposition~\ref{LBPower} of this section, we first provide a criterion to define the meaning of a ``good'' direction of cutting, given a stationary covariance function. Next, using the first and the second theorems that follow, we characterize the direction that optimizes the criterion. Finally, we introduce a constrained optimization that finds the desired direction using a likelihood function of a sample of the training data.

Recall that each subdomain $\Omega_s$ uses a low-rank approximation for its covariance matrix. Therefore, a natural criterion is to look for subdomains such that the error for this approximation is minimized. Therefore, given a symmetric positive semidefinite kernel $\phi(\cdot,\cdot):\Omega_s\times\Omega_s\rightarrow \mathbb{R}$, our objective  is to create subdomain $\Omega_s$ for which the expected covariance approximation error at any $z\in \Omega_s$ using a set of pseudo inputs $\tilde{{\mathbf{X}}}_s$, i.e., 
\begin{eqnarray}\label{ErrorTerm}
\mathbb{E}_{\Omega_s}(h-\mathbf k_{\mathbf z\tilde{\mathbf{X}}_s}\mathbf K_{\tilde{\mathbf{X}}_s\tilde{\mathbf{X}}_s}^{-1} \mathbf k_{\tilde{\mathbf{X}}_s\mathbf z}),
\end{eqnarray} 
where the expectation operator is with
respect to all $\mathbf z$ and $\tilde{{\mathbf{X}}}_s$ over $\Omega_s$ and $h=\phi(\mathbf z,\mathbf z)$, is minimized (\comm{see Appendix~\ref{ErrorFuncAPX} for derivation of covariance approximation error}). However, since the expected error has a complicated form and its direct calculation is a
challenging task, we seek an upper bound for this term and minimize that instead.

\begin{proposition}\label{LBPower}
Let $\phi(\cdot,\cdot)$ denote a stationary covariance function, and $h=\phi(\mathbf t,\mathbf t)\in\mathbb{R}$ be the evaluation of kernel $\phi(\cdot,\cdot)$ at an arbitrary point $\mathbf t\in\Omega_s$. Then, 
$\mathbb{E}_{\Omega_s}(\phi^2(\mathbf{x},\mathbf{x}'))\leq h\mathbb{E}_{\Omega_s}(\mathbf k_{\mathbf z,\tilde{\mathbf{X}}_s}\mathbf K_{\tilde{\mathbf{X}}_s\tilde{\mathbf{X}}_s}^{-1} \mathbf k_{\tilde{\mathbf{X}}_s\mathbf z})$, where $\mathbf{x},\mathbf{x}',\mathbf{z},\tilde{\mathbf{x}}_1,\ldots, \tilde{\mathbf{x}}_{m_s}$ are i.i.d random vectors sampled from subdomain $\Omega_s$ according to some probability distribution $\mathscr{P}$.
\end{proposition}

\begin{proof}
See Appendix \ref{Theorem1APX} in the supplemental material for proofs of all theorems and propositions.
\end{proof}

Propositions~\ref{LBPower} provides an upper bound, i.e., $h(1-\frac{1}{h^2}\mathbb{E}_{\Omega_s}(\phi^2(\mathbf{x},\mathbf{x}')))$, on expected error~\eqref{ErrorTerm} (See Appendix~\ref{SimulAPX} for a simulation study  showing that the relation between $\mathbb{E}_{\Omega_s}(\phi^2(\mathbf{x},\mathbf{x}'))$ and expected error~\eqref{ErrorTerm} is more profound. In fact, under certain conditions by increasing $\mathbb{E}_{\Omega_s}(\phi^2(\mathbf{x},\mathbf{x}'))$, the expected error term itself monotonically decreases). Therefore, we seek to construct the subdomains so that the expected covariance squared function is maximized, i.e., the upper bound of the expected error is minimized. We note that Propositions~\ref{LBPower} makes a stationarity assumption and therefore the results may not hold for other types of covariance functions. However, because we use independent covariance functions in each subdomain, we are still able to handle the heterogeneity, i.e., using different parameters for each local covariance function. 

For our theoretical framework, we consider a general scenario where, after standardizing the data, the domain of data, $\Omega\subset \mathbb{R}^p$, is (or is inscribed in) a hypercube  with edge length $L$, one vertex is on the origin, and all the edges are parallel to one axis of $\mathbb{R}^p$. \comm{The assumption that the domain of the data is inscribed in a hypercube is valid even if each dimension of the original input domain has different length; this is because after standardization, all the dimensions have the same length.} Also we assume that the data points are uniformly sampled from $\Omega$, specifically,
\begin{eqnarray}
x_1,\ldots, x_p \stackrel{i.i.d}{\sim}\mathcal{U}(0,L) & \forall\mathbf{x}\in \Omega.\label{USHDist}
\end{eqnarray}
We call such an $\Omega$ a \emph{uniform straight hypercube}. 

Moreover, we consider the anti-isotropic squared exponential function as the choice of the covariance function, 
\begin{eqnarray}
\mathcal{\phi}(\mathbf{x},\mathbf{x}')=\exp\big(-(\mathbf{x}-\mathbf{x}')^T\boldsymbol{\Gamma }(\mathbf{x}-\mathbf{x}')\big),\label{SEKernel}
\end{eqnarray}
where $\boldsymbol{\Gamma}$ is a diagonal matrix with length-scale parameters $\gamma_1,\ldots,\gamma_p$ on the diagonal, and without loss of generality, assume $\gamma_1\leq\ldots\leq\gamma_p$. We note that the squared function of~\eqref{SEKernel}, i.e., $\mathcal{\phi}^2(\mathbf{x},\mathbf{x}')$, is a new squared exponential covariance function with the length scale parameters $2\gamma_1\leq \ldots \leq 2\gamma_p$. Hence, as $\mathbb{E}_{\Omega_s}\big(\mathcal{\phi}(\mathbf{x},\mathbf{x}')\big)$ increases, $\mathbb{E}_{\Omega_s}\big(\mathcal{\phi}^2(\mathbf{x},\mathbf{x}')\big)$ increases.

Given the $k^{th}$ primary axis and the vector of angles $\boldsymbol{\theta}=\{\theta_1,\ldots,\theta_p\}\backslash\{\theta_k\}$ and assuming that the cutting hyperplanes are equidistant (with distant $W=L\slash S$ from each other), all the subdomains and cutting hyperplanes can be fully characterized (\comm{See Appendix~\ref{Theorem1APX}}). We denote the $s^{\text{th}}\in [S]$ subdomain created on $\Omega$ by $\Omega_{\boldsymbol{\theta},k,W,s}$, where the indices $\boldsymbol{\theta}$, $k$ and $W$ indicate that the cutting hyperplanes are defined by the vector of angels $\boldsymbol{\theta}$, the $k^{th}$ primary axis, and the distance $W$. Note that the cutting hyperplanes are orthogonal to the axis $k$ only if $\boldsymbol{\theta} = \mathbf 0$, that is $\theta_j=0$ for $ j\in [p]\backslash\{k\}$. 

\begin{theorem}\label{TheoremParandRec}
Let $\Omega\subset\mathbb{R}^p$ be a uniform straight hypercube with side length $L$, and let $\phi(\cdot,\cdot)$ denote covariance function~\eqref{SEKernel}. Then, for a fixed $W= L\slash S$, $s \in [S]$, and $k\in [p]$, $\Omega_{\boldsymbol{0},k,W,s}$ gives the maximum expected covariance, i.e., 
\[\argmax_{\boldsymbol{\theta}} \mathbb{E}_{\Omega_{\boldsymbol{\theta},k,W,s}}\big(\phi(\mathbf{x},\mathbf{x}')\big)=\mathbf{0}.\]
\end{theorem}

While Theorem~\ref{TheoremParandRec} shows that cutting orthogonally to the given axis $k\in [p]$, i.e., $\boldsymbol{\theta}=\boldsymbol{0}$, maximizes the expected covariance compared to any other $\boldsymbol{\theta}> \boldsymbol{0}$, Theorem~\ref{TheoremCompMainDirs} further shows that among all the subdomains created by cutting orthogonally to a primary axis, the one created by cutting orthogonally to the axis associated with the fastest direction
of change, i.e., the direction associated with the largest $\gamma$, has the maximum expected covariance

\begin{theorem}\label{TheoremCompMainDirs}
Let $\Omega\subset\mathbb{R}^p$ be a uniform straight hypercube with side length $L$, and let $\phi(\cdot,\cdot)$ denote covariance function~\eqref{SEKernel}. Then for a fixed $W= L\slash S$ and $s \in [S]$, among all the subdomains $\Omega_{\mathbf{0},k,W,s}$ for $k\in[p]$, $\Omega_{\mathbf{0},p,W,s}$ gives the maximum expected covariance, i.e.,
\[\argmax_k \mathbb{E}_{\Omega_{\mathbf{0},k,W,s}}\big(\mathcal{\phi}(\mathbf{x},\mathbf{x}')\big)=p.\]
\end{theorem}

Theorems~\eqref{TheoremParandRec} and~\eqref{TheoremCompMainDirs} along with the property of covariance function~\eqref{SEKernel}, i.e., larger values of $\mathbb{E}_{\Omega_s}\big(\mathcal{\phi}(\mathbf{x},\mathbf{x}')\big)$ imply larger $\mathbb{E}_{\Omega_s}\big(\mathcal{\phi}^2(\mathbf{x},\mathbf{x}')\big)$, provide a partitioning policy for the domain $\Omega$. That is, cutting orthogonal to the direction of the fastest covariance decay reduces the upper bound of expected error~\eqref{ErrorTerm}, and therefore, gives a more accurate covariance approximation in each subdomain. \comm{The policy of cutting orthogonal to the direction of the fastest covariance decay minimizes the correlation between the neighboring subdomains. This is because the covariance on the two sides of each boundary decays faster than any other direction.}

However, we note that the direction of the fastest covariance decay may not necessarily be a primary axis of the input domain. To overcome this drawback,  we relax the restriction of choosing one of the primary axes as the direction of the fastest covariance decay by using a general form of the squared exponential covariance function, $
\mathcal{\phi(\mathbf{x},\mathbf{x}')}=\text{exp}(-(\mathbf{x}-\mathbf{x}')^T \mathbf{M} (\mathbf{x}-\mathbf{x}'))$, where $\mathbf{M}$ is a $p \times p$ positive definite matrix~\citep{Ras}. For the purpose of this discussion, we define $\mathbf{M}$ as $\mathbf{a}\mathbf{a}^T+\gamma \mathbf{I}_p$, where $\mathbf{a}$ is a unit direction vector in the input space with length $p$, and $\gamma$ is a joint length-scale parameter, to obtain  the following covariance function,
\begin{eqnarray}
\phi^{\mathbf{a}}(\mathbf{x},\mathbf{x}')=\text{exp}(-(\mathbf{x}-\mathbf{x}')^T (\mathbf{a}\mathbf{a}^T+\gamma \mathbf{I}_p) (\mathbf{x}-\mathbf{x}')),\label{BestDirCov}
\end{eqnarray}
which involves a dot product $(\mathbf{x}-\mathbf{x}')^T \mathbf{a}$. This means that for a given distance $||\mathbf{x}-\mathbf{x}'||_2$, the angle between $\mathbf{x}-\mathbf{x}'$ and $\mathbf{a}$ determines the covariance. In particular, the direction $ \mathbf{a}$ itself has relatively the highest rate of covariance decay.

Although in practice, direction $\mathbf a$ may not exist, fitting covariance function $\eqref{BestDirCov}$ to the data using Maximum Likelihood Estimation can find the best choice of $\mathbf{a}$ under the MLE criterion. Therefore, under the GP assumptions, we maximize the logarithm of the likelihood function to find the optimal value of vector $\mathbf a$,
\begin{eqnarray}
\max_{\mathbf{a},\gamma,\sigma^2} -\mathbf{y}^T(\mathbf{K}^{\mathbf{a}}+\sigma^2\mathbf{I})^{-1}\mathbf{y}-\text{log}|\mathbf{K}^{\mathbf{a}}+\sigma^2\mathbf{I}|,\label{PrimaryLik}
\end{eqnarray}
where $\mathbf{K}^{\mathbf{a}}$ is the covariance matrix formed based on covariance function~\eqref{BestDirCov}.
 
Here, since we only want to find direction $\mathbf{a}$, the nuisance parameters are the variance and the length scale parameters, $\sigma^2$ and $\gamma$. Therefore, to shrink the parameter space, we set $\sigma^2$ and $\gamma$ to small values after standardizing the data.

Note that optimization problem~\eqref{PrimaryLik} is of $\mathcal{O}(N^3)$, which is the same order of complexity as the original problem. However, since the output of optimization~\eqref{PrimaryLik} is merely used to find a desired direction, and is not used for prediction, we utilize a small subset of data with size  $n\ll N$. Further, since $\mathbf{a}$ is a unit direction vector, we write $\mathbf{a}=[\bar{\mathbf{a}}^T,\sqrt{1-\bar{\mathbf{a}}^T\bar{\mathbf{a}}}]^T$, where  $\bar{\mathbf{a}}=[a_1,\ldots,a_{p-1}]^T$, and add the unity constraint, $\bar{\mathbf{a}}^T\bar{\mathbf{a}}\leq 1$, to the optimization problem. Consequently,
\begin{eqnarray}
\begin{aligned}
&\min_{\bar{\mathbf{a}}}
&&\mathcal{L}(\bar{\mathbf{a}}) =\mathbf{y}^T_n(\mathbf{K}_n^{\bar{\mathbf{a}}}+\sigma^2\mathbf{I_n})^{-1}\mathbf{y}_n+\text{log}|\mathbf{K}_n^{\bar{\mathbf{a}}}+\sigma^2\mathbf{I}_n|\\
&\text{subject to}
&&\bar{\mathbf{a}}^T\bar{\mathbf{a}}\leq 1,
\end{aligned}\label{PreFinalLik}
\end{eqnarray}
where $\mathbf{y}_n$ is the response vector of the small subset of data and $\mathbf{K}_n^{\bar{\mathbf{a}}}$ is the covariance matrix evaluated by covariance function~\eqref{BestDirCov} on the same small subset (See Appendix~\ref{PGDSol} for solving optimization problem~\eqref{PreFinalLik} by using Projected Gradient Descent~\citep{nesterov1994interior}).

\comm{We also note that optimization~\eqref{PreFinalLik} finds the direction of the fastest covariance decay independent of the assumptions stated for Theorems~\ref{TheoremParandRec} and~\ref{TheoremCompMainDirs}. Our experiments in Section~\ref{sec_seb_Direction} show that the directions found by optimization~\eqref{PreFinalLik} can significantly increase the accuracy of SPLK, even if the original input domains are not hypercubes or if the data points are not uniformly distributed.} We refer the reader to Appendix~\ref{SimulAPX} for intuition behind the theoretical results presented above.

\subsection{Computational complexity analysis of SPLK}\label{CompCPLX}

\comm{This section presents the computational complexity analysis for SPLK. To this end, we look at the computational costs of calculating the local mean and variance predictors in Section~\ref{sub_sec_mean_pred}, and finding the direction of cut in Section~\ref{DirofCuts}. In addition, we analyze training the boundary functions presented in Appendix~\ref{sub_sec_numb_pd} and training the local models presented in Appendix~\ref{sub_sec_numb_hyp}.}

Calculating the local mean and variance predictors in each subdomain (explained in Section~\ref{sub_sec_mean_pred} and Appendix~\ref{SolutionOfLagrangianAPX}) requires inverting the low-rank covariance matrix $\mathbf{G}_s$ and the boundary covariance matrix $[(\text{diag}(\tilde{\mathbf K}^s_{\mathbf{B}_s\mathbf{X}_s}\tilde{\mathbf K}^s_{\mathbf{X}_s\mathbf{B}_s}))^{-1}(\tilde{\mathbf K}^s_{\mathbf{B}_s\mathbf{X}_s}\tilde{\mathbf K}^s_{\mathbf{X}_s\mathbf{B}_s})]\circ\mathbf K^s_{\mathbf{B}_s\mathbf{B}_s}$ with sizes $n_s\times n_s$ and $|\mathbf{B}_s|\times |\mathbf{B}_s|$, respectively. Using Woodbury, Sherman and Morrison matrix inversion lemma~\citep{hager1989updating}, the computational complexity of inverting $\mathbf{G}_s$ is of the order of $\mathcal{O}(n_sm_s^2)$, where $m_s\ll n_s$; therefore, the complexity of calculating each local mean and variance predictor becomes $\mathcal{O}(|\mathbf{B}_s|^3+N_sm_s^2)$. Also, training each local model (explained in Appendix~\ref{sub_sec_numb_hyp}) requires maximizing the local likelihood function~\eqref{eq:log_lik}.
\cite{Snl} shows that the cost of finding the derivatives and maximizing~\eqref{eq:log_lik} is of the order of $\mathcal{O}(n_sm_s^2)$. Therefore, denoting $m$ as the average number of pseudo-inputs in each subdomain, and $Q$ as the average number of control points on each boundary, which implies $|B_s|\approx 2Q$, we obtain $\mathcal{O}(2Nm^2+6SQ^3)=\mathcal{O}(Nm^2+SQ^3)$ as the total computation complexity of calculating the local mean and variance predictors and training local models.

Furthermore, since we train the boundary function~\eqref{BoundaryFunc} (explained in Appendix~\ref{sub_sec_numb_pd}) using the full GPR on the set of neighboring data points $\boldsymbol{\Delta}_{\ell}$, the computational complexity of training the boundary functions is of the order of $\mathcal{O}(S\Delta^3)$, where $\Delta$ is the average size of all $\boldsymbol{\Delta}_{\ell}$. Also, solving the optimization~\eqref{PreFinalLik} for finding the direction of cut, through solving optimization problem~\eqref{PreFinalLik}, is dominated by the matrix inversion $(\mathbf{K}_n^{\bar{\mathbf{a}}}+\sigma^2\mathbf{I_n})^{-1}$, which has the order $\mathcal{O}(n^3)$.

Consequently, the total computational complexity of SPLK is of the order of $\mathcal(Nm^2+S(Q^3+\Delta^3)+n^3)$. We note that as the dimension of the training data increases, more control points are required to be located on the boundaries, which implies $Q$ implicitly depends on $p$; however, since the application of the current study focuses on moderate dimensional problems, the complexity of SPLK is dominated by $\mathcal{O}(Nm^2)$, under the assumption that the values of $n\ll N$ and $\Delta\ll N$ are independent of $N$. Section~\ref{ChoicesOFParams} will discuss this assumption and the choice of other tuning parameters.

\subsection{Choice of the tuning parameters of SPLK}\label{ChoicesOFParams}
\comm{This section presents some guidelines for the selection of the tuning parameters of SPLK, which are $S$, $m$, $n$, $\Delta$, $Q$.}

As the average number of local pseudo-inputs in each subdomain, $m$, increases, the accuracy of SPLK increases at the expense of higher computation time. Such a trade-off rules out an ``optimal'' value for $m$.~\cite{williams2002observations} shows that as the eigenspectrum of the underlying covariance function decays more quickly, given a fixed set of pseudo-inputs, the Nystr{\"o}m approximation becomes more accurate. Therefore, the choice of $m$ depends on the covariance structure of the function of interest. However, since SPLK optimizes the distribution of pseudo-inputs in each subdomain using SPGP approximation, SPLK generally requires a smaller number of local pseudo-inputs compared to other approximation methods that use ad-hoc selection of pseudo-inputs~\citep{Snl}. In order to have a computationally efficient algorithm, we suggest setting $m$ of the order of $\mathcal{O}(\sqrt{N})$, i.e., $m=\kappa\sqrt{N}$, where $\kappa$ is a tuning parameter that determines the density of pseudo-inputs in each subdomain. Our experiments in section~\ref{sec_results} show that setting $1<\kappa<9$ results in efficient computation time and relatively high accuracy. Alternatively, we note that a Bayesian approach can also be used for the selection of $m$ ~\citep{pourhabib2014bayesian}, however the computation time is significantly increased by the Markov Chain Monte Carlo sampling that is required in the Bayesian approach.

Similar to $m$, a trade-off exists between the accuracy and computation time for $S$, the number of subdomains. As mentioned earlier, regardless of the approach used for handling the boundary conditions, a larger  number of subdomains reduces the computation time as well as the prediction accuracy; smaller local models can be trained more efficiently but result in a larger number of boundaries, which in turn reduces the overall accuracy. This is because the accuracy of the local models decreases in regions close to their boundaries. We suggest choosing $S$ such that each subdomain contains between $500$ and $5000$ data points. Based on our experiments, choosing an $S$ that results in subdomains with more than $5000$ data points makes the parameter estimation of each local model computationally inefficient. On the other hand,  a value of $S$ that results in subdomains with less than $500$ data points creates too many boundaries, which reduces the accuracy. 

Next, we discuss $Q$, the number of control points on each boundary. As the dimension of the input domain increases, we need to locate more control points to efficiently handle the boundary conditions. We suggest setting $Q$ proportional to the dimension of the boundary to effectively cover the boundary spaces. Specifically, we use
$Q=q^{p-1}$, where $p$ is the dimension of the domain of data, and $q$ determines the density of control points on each boundary space. Moreover, in order to balance the computation time between training the subdomains, which is of $\mathcal{O}(Nm^2)$, and handling boundary conditions, which is of $\mathcal{O}(SQ^3)$, we suggest $(\frac{\kappa^2 N^2}{S})^{\frac{1}{3p-3}}$ as an upper bound for $q$, which enforces $\mathcal{O}(SQ^3)<\mathcal{O}(Nm^2)$. Theoretically, SPLK can be applied to even higher dimensional spaces, but as the dimension of the input domain increases, the upper bound for $q$ decreases, which means a more sparse distribution of the control points~\citep{park2018patchwork}. \BFcomm{Also, SPLK uses uniform distribution of control points on the boundaries (see Appendix~\ref{sub_sec_numb_pd}), which might not be efficient in higher dimensions due to the sparsity of the control points}. Therefore, we do not recommend the application of SPLK to very high dimensional spaces. Our experiments in Section~\ref{sec_seb_cpdensity} show that choosing $q\in [2,3]$ provides satisfactory results in terms of both computation time and accuracy. 

As mentioned in Section~\ref{DirofCuts}, a small subset of data with size $n$ is used to merely find a desired direction for applying the cutting hyperplanes, and as such we suggest $n\ll N$. For our experiments in Section~\ref{sec_results}, we choose $n=1000$ to find the cutting direction through solving optimization problem~\eqref{PreFinalLik}, which resulted in a small computational overhead. 

Finally, for the choice of $\Delta$, the average number of neighboring data points of each boundary, we suggest setting $\delta=0.1 L$, where $L$ is the width of each subdomain, and $\delta$ is the maximum distance of the neighboring data points to their associated boundary (see Section~\ref{sub_sec_numb_pd}). This choice of $\delta$ ensures that the local data points reasonably close to the boundaries when training the boundary functions. Moreover, assuming data points are uniformly disturbed in the input domain, we set $100<\Delta<1000$ for subdomains with sizes ranging between 500 and 5000, which reduces computational overhead when training the boundary functions.

\section{Experimental results}\label{sec_results}
In this section, we apply SPLK to four real datasets and compare its performance with local probabilistic regression (LPR) ~\citep{urtasun2008sparse},  Bayesian committee machine (BCM)~\citep{tresp2000bayesian}, bagged Gaussian process (BGP)~\citep{chen2009bagging}, partial independent conditional GP (PIC)~\citep{PIC}, DDM~\citep{park2012gplp}, \comm{and PWK~\citep{park2018patchwork}}. We use the BGP, LPR, BCM, and DDM implementations in the GPLP toolbox~\citep{park2012gplp}, PWK and BCM implementations provided by the authors of~\citep{park2018patchwork} and~\cite{schwaighofer2003transductive} respectively. We  also conduct sensitivity analysis of the parameters in SPLK and propose some guidelines for their selection.

\subsection{Datasets and evaluation criteria}

We  implement  SPLK in MATLAB and test it on four real datasets:
\begin{enumerate}
\item The spatial dataset, TCO, which contains 65000 observations, collected by the NIMBUS7 satellite for NASA's Total Ozone Mapping Spectrometer (TOMS) project (\url{https://www.nodc.noaa.gov}). The global measurement was conducted on a two-dimensional grid, i.e., latitude and longitude, from 1978 to 2003 on a daily basis. We select the measurements of ``total column of ozone'' on this grid for the data collected on January 1, 2003. The dataset is highly non-stationary and an appropriate dataset for comparing SPLK and DDM because it is constructed on a two-dimensional input space,

\item The spatial dataset, Levitus, which contains 56000 observations, is a part of the world ocean atlas that measures the annual means of major ocean parameters (\url{http://iridl.ldeo.columbia.edu/SOURCES/.LEVITUS94}). 
The global measurement was conducted on a three-dimensional grid, i.e., latitude, longitude, and depth, in 1994. We select the ``apparent oxygen utilization'' as the response variable on this grid.
\end{enumerate}
Recalling  that handling exogenous variables in spatial datasets also motivates this paper, we use a third real dataset. 
\begin{enumerate}
\setcounter{enumi}{2}
\item The spatial dataset, Dasilva, which contains 70000 observations, is a part of a five-volume atlas series of Surface Marine Data (\url{http://iridl.ldeo.columbia.edu/SOURCES/.DASILVA/.SMD94/.halfbyhalf/.climatology/}). The global measurement was conducted on a two-dimensional grid, i.e., latitude and longitude, on a monthly basis in 1994. We select three exogenous variables, ``constrained outgoing heat flux'', ``zonal heat flux'', and ``sea minus air temperature'', and the objective is to predict ``long wave Chi sensitivity'' based on the data collected on January 1994. 
\end{enumerate}

Although SPLK was developed to handle spatial datasets, the methodology is general and can be efficiently applied to non-spatial data that have a moderate number (say ten or fewer) of exogenous variables. We use a fourth dataset to demonstrate the performance of SPLK on non-spatial data.
\begin{enumerate}
\setcounter{enumi}{3}
\item The non-spatial dataset, Protein, which contains 46000 measurements, is a collection of Physicochemical Properties of Protein Tertiary Structure (\url{http://archive.ics.uci.edu/ml/datasets/Physicochemical+Properties+of+Protein+Tertiary+Structure}). This dataset contains nine ``physicochemical properties'' of proteins as explanatory variables and ``size of the residue'' as the response variable.
\end{enumerate}

We randomly partition each dataset into $90\%$ for training and $10\%$ for testing. We use three measures to evaluate the performance of each method. The first one is the measure of prediction accuracy, which is assessed by the Mean Squared Error (MSE),
\begin{eqnarray}
\text{MSE}=\frac{1}{T} \sum_{i=1}^{T} (y_*^i-\mu_*^i)^2,
\end{eqnarray}
where $y_*^i$ is the noisy observation of the test location $\mathbf x_*$ and $\mu_*^i$ is the mean prediction of this test location. The second measure is the Negative
Log Predictive Density (NLPD) that takes into account uncertainty in prediction in addition to accuracy, specifically 
\begin{eqnarray}
\text{NLPD}=\frac{1}{T} \sum_{i=1}^{T} \frac{(y_*^i-\mu_*^i)^2}{2(\sigma_*^i)^2}+0.5\log (2\pi{\sigma_*^i}^2),
\end{eqnarray} 
where $\sigma_*^2$ is variance of the predictor at the test location $\mathbf x_*$. The third measure is the computation time, i.e., training plus testing time, that evaluates the success of SPLK in speeding up GPR. Note that the computation time on its own is not an appropriate measure, and the corresponding MSE or NLPD must also be taken into account, as a reduction in training time without an accurate prediction is not useful. \comm{Finally, variable selection is beyond the scope of the current study, as we assume that the input variables in each dataset are significant predictors which have passed the variable selection process based on the domain knowledge or a statistical procedure.}

\subsection {Computation time and prediction accuracy}\label{Alg_Comp_sec}
Here, we compare the computation time and the prediction accuracy of SPLK with those of the competing algorithms. Specifically, we consider the MSE and NLPD as functions of the computation time and plot the set of best results for each algorithm. Under this criterion, the algorithm associated with the curve closest to the origin will be superior. The parameter selection for each algorithm is as follows.

For SPLK, we solve optimization problem~\eqref{PreFinalLik} for each dataset to find the direction of the cuts. It turns out that for the spatial dataset the best direction, based on the criteria of optimization problem~\eqref{PreFinalLik}, is one of the primary axes of the dataset domains: For dataset TCO, the best direction is the direction of the first primary axis (i.e., latitude), for dataset Levitus, it is the direction of the third primary axis (i.e., depth), and for dataset Dasilva, it is the direction of the first primary axis (i.e., latitude). For dataset Protein, which is not a spatial dataset, the best direction is not the direction of any of the primary axes of the input domain (see Section~\ref{sec_seb_Direction} for a discussion of cuts in other directions). It is insightful to observe that for the spatial datasets used in this study the solution to optimization problem~\eqref{PreFinalLik} is aligned with one of the primary axes, which may reflect a relationship between the response surface and the underlying geology. For example, for measuring ``long wave Chi sensitivity'' in dataset Dasilva the direction of the fastest change is the same as latitude; or for dataset Levitus, the covariance decays fastest when we change the depth of the measurement for ``apparent oxygen utilization.''

\begin{figure}[ht]
\begin{center}
	\begin{subfigure}{0.45\textwidth}
		\includegraphics[height=4cm,width=7.5cm]{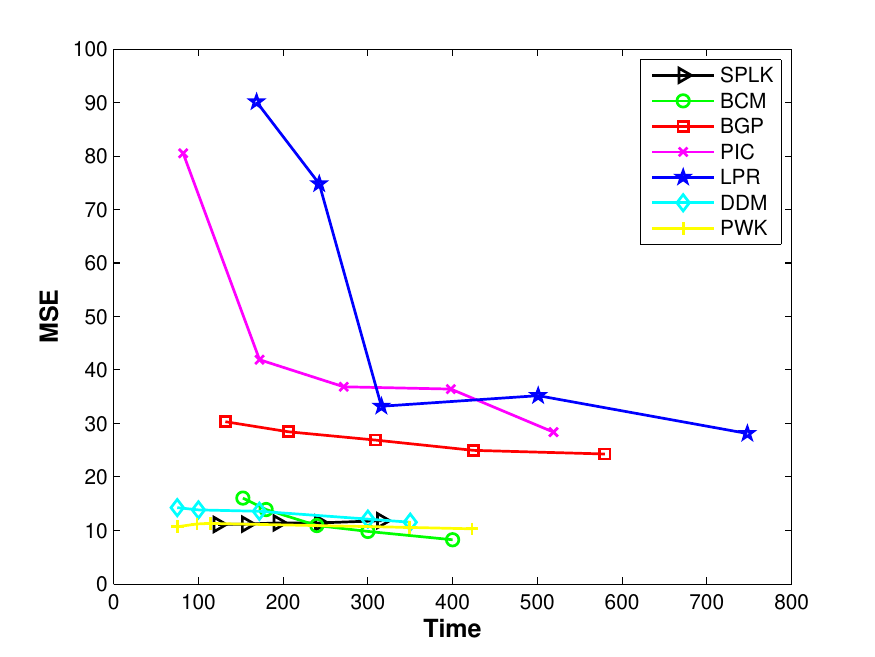}
		\caption {\tiny{TCO}:$\kappa=6, S \in \{20,30,40,50,60\},q=3$}
\label{fig:TimeVsMSETCO}
	\end{subfigure}
	\begin{subfigure}{0.45\textwidth}
		\includegraphics[height=4cm,width=7.5cm]{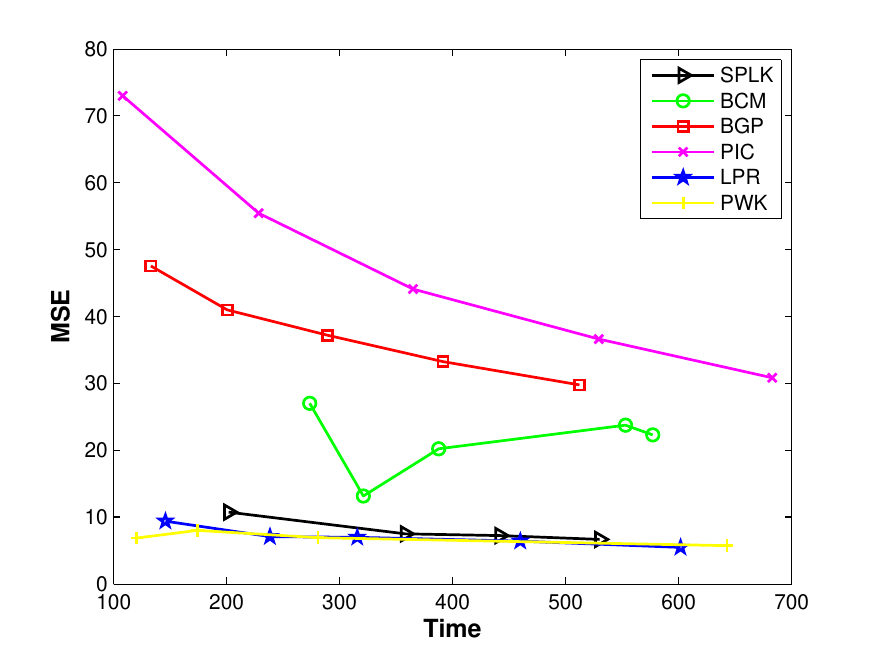}
		\caption {\tiny{Levitus}:$\kappa=8, N \in \{8,11,16,33\},q=3$ }
\label{fig:TimeVsMSELevitus}
	\end{subfigure}
\begin{subfigure}{0.45\textwidth}
		\includegraphics[height=4cm,width=7.5cm]{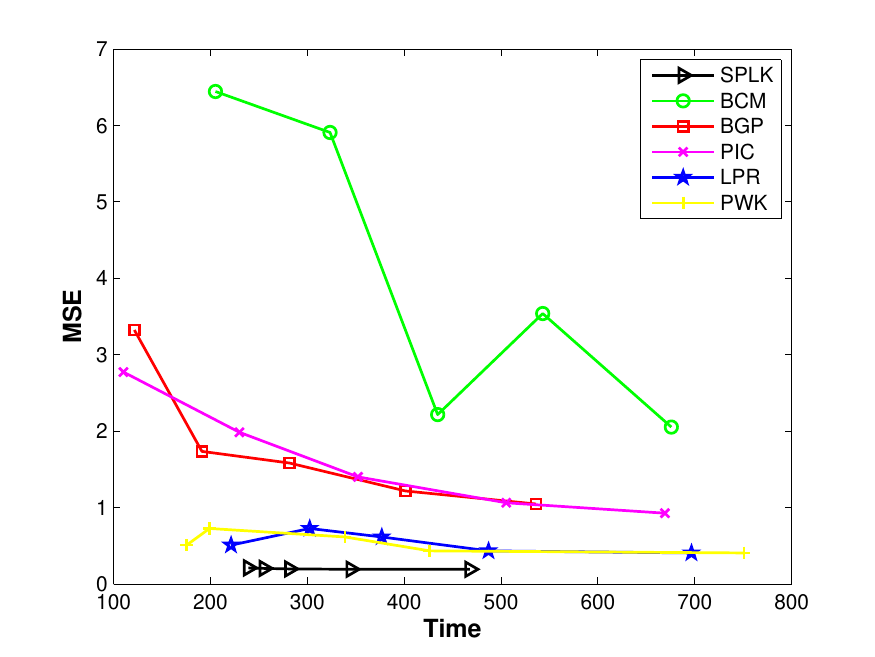}
		\caption {\tiny{Dasilva}:$k=8, N \in \{20,30,40,50,60\},q=3$ }
\label{fig:TimeVsMSEDasilva}
	\end{subfigure}
\begin{subfigure}{0.45\textwidth}
		\includegraphics[height=4cm,width=7.5cm]{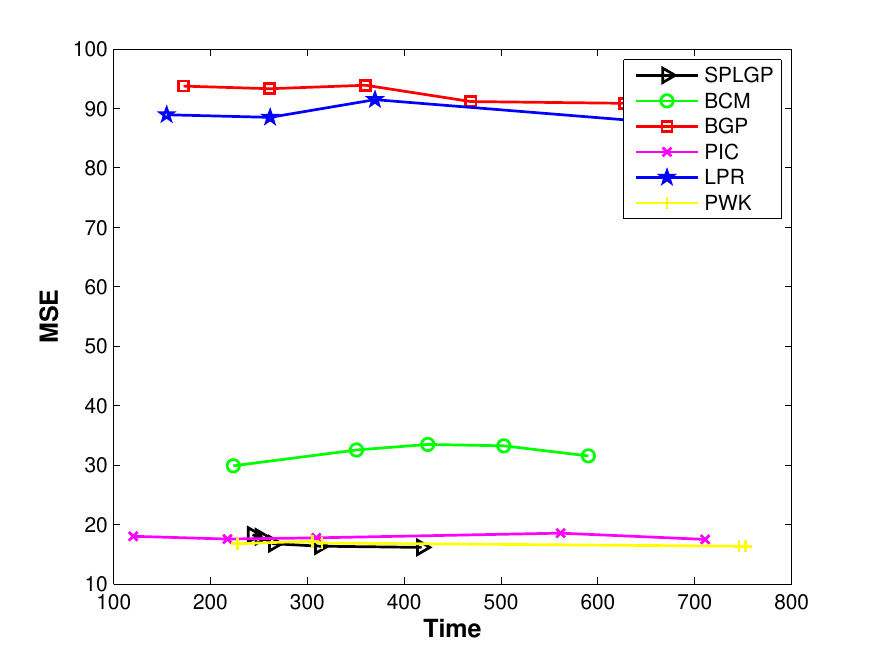}
		\caption {\tiny{Protein}:$k=8, N \in \{20,30,40,50,60\},q=2.2$}
\label{fig:TimeVsMSEProtein}
	\end{subfigure}
\end{center}
\caption{MSE versus computation time. For DDM, $Q=3$ and $S\in \{100,200,300,400,500\}$; for PWK, $(Q,S)\in \{3,5,7\}\otimes\{100,200,300,400,500\}$; for PIC, $S=500$ and $m\in \{100,200,300,400,500,600\}$; for BCM, $S\in \{200,300,400,500,600,700\}$; for LPR, $(S,m,R) \in \{5,10,15,20\}\otimes\{100,200,300\}\otimes\{500,1000,1500\}$; and for BGP, $S=40$ and $m\in \{500,600,700,800,900\}$}
\label{fig:MSEVSOther}
\end{figure}

\begin{figure}[ht]
\begin{center}
	\begin{subfigure}{0.45\textwidth}
		\includegraphics[height=4cm,width=7.5cm]{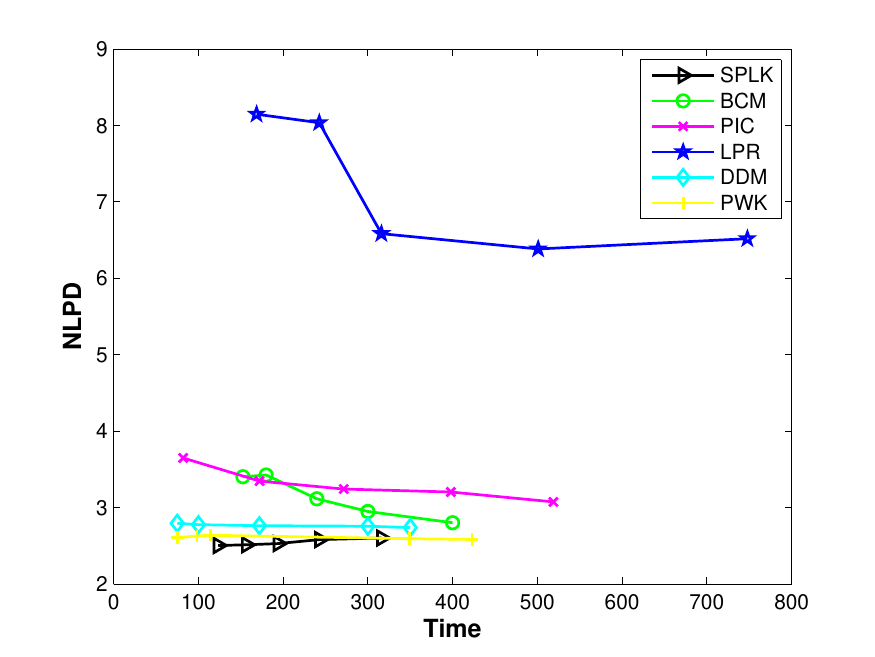}
\caption {\tiny{TCO}:$\kappa=6, S \in \{20,30,40,50,60\},q=3$}
\label{fig:TimeVsNLPDTCO}
	\end{subfigure}
	\begin{subfigure}{0.45\textwidth}
		\includegraphics[height=4cm,width=7.5cm]{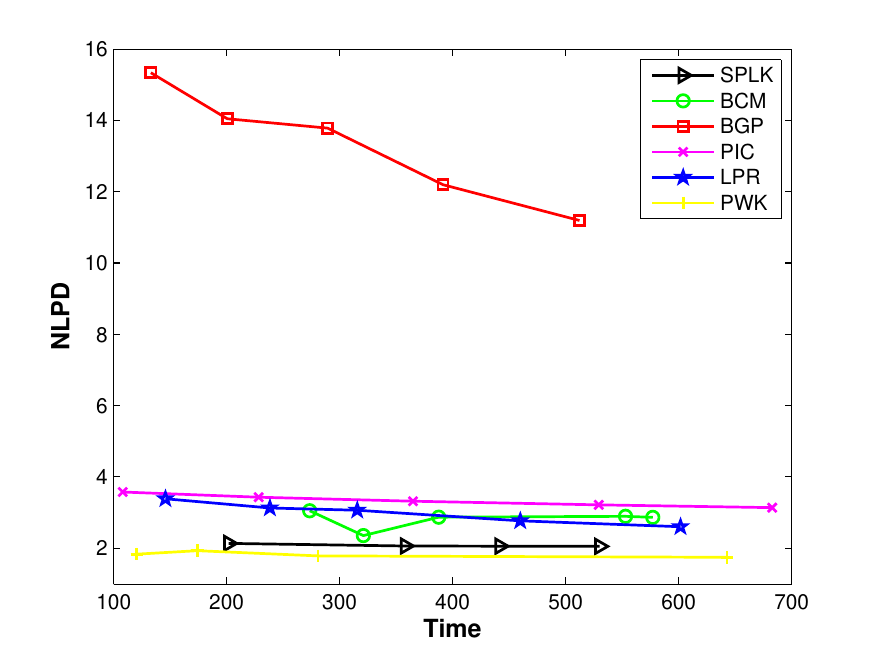}
	\caption {\tiny{Levitus}:$\kappa=8, N \in \{8,11,16,33\},q=3$ }
\label{fig:TimeVsNLPDLevitus}
	\end{subfigure}
\begin{subfigure}{0.45\textwidth}
		\includegraphics[height=4cm,width=7.5cm]{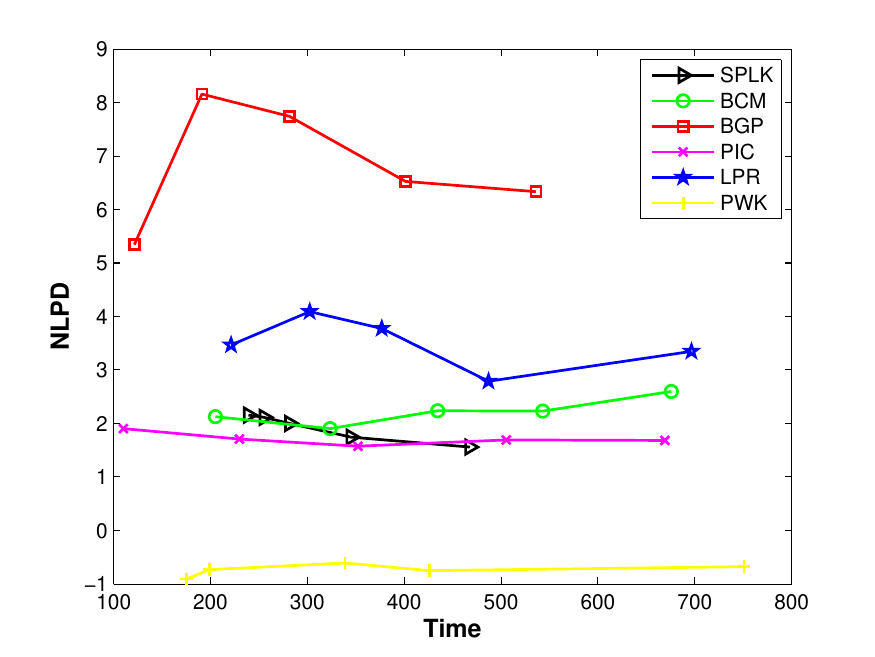}
			\caption {\tiny{Dasilva}:$k=8, N \in \{20,30,40,50,60\},q=3$ }
\label{fig:TimeVsNLPDDasilva}
	\end{subfigure}
\begin{subfigure}{0.45\textwidth}
		\includegraphics[height=4cm,width=7.5cm]{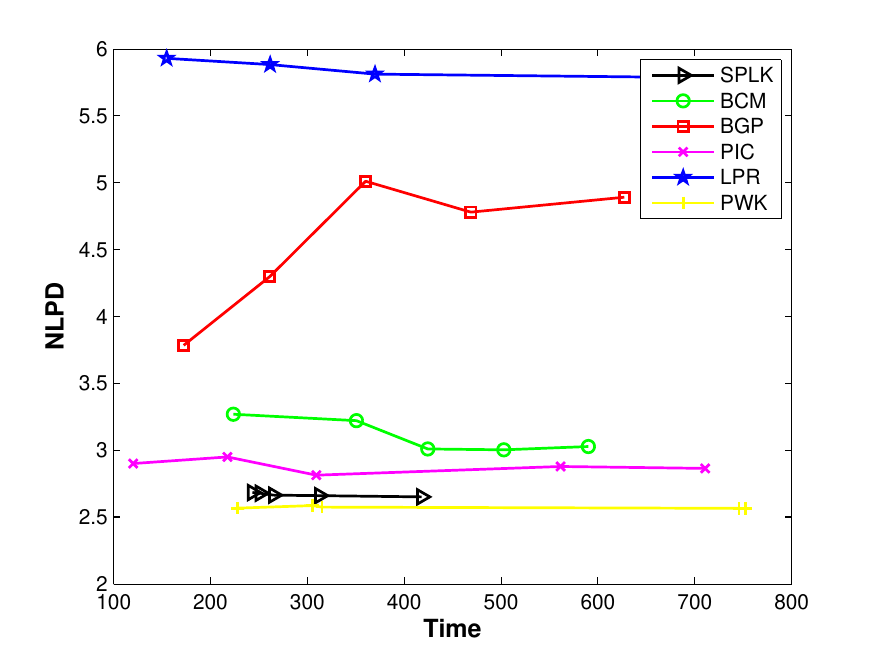}
		\caption {\tiny{Protein}:$k=8, N \in \{20,30,40,50,60\},q=2.2$}
\label{fig:TimeVsNLPDProtein}
	\end{subfigure}
\end{center}
\caption{NLPD versus computation time. For DDM, $Q=3$ and $S\in \{100,200,300,400,500\}$; for PWK, $(Q,S)\in \{3,5,7\}\otimes\{100,200,300,400,500\}$; for PIC, $S=500$ and $m\in \{100,200,300,400,500,600\}$; for BCM, $S\in \{200,300,400,500,600,700\}$; for LPR, $(S,m,R) \in \{5,10,15,20\}\otimes\{100,200,300\}\otimes\{500,1000,1500\}$; and for BGP, $S=40$ and $m\in \{500,600,700,800,900\}$}
\label{fig:NLPDVSOther}
\end{figure}

We use the guidelines discussed in Section~\ref{ChoicesOFParams} for choosing the tuning parameters. We choose $S$ from the set $\{20,30,40,50,60\}$, except for the dataset Levitus, to keep the number of local data point in each subdomain between $500$ and $5000$. For Levitus, since we cut the domain of data from the third direction with 33 distinct levels, we choose $S$ from the set $\{8,11,16,33\}$. Our experiments in Section~\ref{sec_seb_cpdensity} suggest that setting $q$ to small values results in a good performance and increasing it does not affect the algorithm's accuracy much. Therefore, we set $q=3$, for datasets TCO, Levitus, and Dasilva, and $q=2.2$ for dataset Protein. We also fix $\kappa=8$ for all the datasets (see Section~\ref{sub_sec_sen} for a discussion of varying values of $\kappa$). Note that as $S$ increases, computation time decreases, so the points with smaller computation times belong to larger values of $S$ in Figures~\ref{fig:MSEVSOther} and~\ref{fig:NLPDVSOther}.

The tuning parameters for DDM are $Q$, the number of control points on each boundary, and $S$, the number of subdomains. For the two-dimensional dataset TCO, we set $Q=3$ and choose  $S$ from the set $\{100,200,300,400,500,600\}$  to keep the average size of the subdomains between $100$ and $600$ as instructed in~\citep{Park}. As expected, for smaller values of $S$, i.e., larger subdomains, the efficiency of the algorithm deteriorates in terms of computation time; therefore, the points with higher computation times belong to smaller values of $S$ in Figures~\ref{fig:MSEVSOther} and~\ref{fig:NLPDVSOther}.

\comm{For PWK, the major tuning parameters are the number of boundary pseudo-observations, $Q$, and the number of subdomains, $S$.} Similar to DDM, PWK suggests keeping the average size of the subdomains between 100 and 600; therefore, we choose the values of $S$ from $\{100,200,300,400,500,$  $600\}$. We also choose the value of $Q$ from the set $\{3,5,7\}$ as suggested in~\citep{park2018patchwork}. Among the 18 possible combinations of $S$ and $Q$, we choose five combinations that have different computation times for the sake of clear demonstration. In Figures~\ref{fig:MSEVSOther} and~\ref{fig:NLPDVSOther}, those points with higher computation times belong to smaller values of $S$.

PIC, which is the localized version of SPGP, has two tuning parameters, the number of local models, $S$, and the number of pseudo-inputs, $m$. We use $k$-means clustering to partition the domain of data into $S$ local models. We note that $m$ is the major tuning parameter that affects the algorithm's computation time. Therefore, we fix $S$ to a reasonable value and choose the values of $m$ from the set $\{100, 200, 300, 400, 500,600\}$. After testing various values of $S$ in the range  of 100 to 800, we find that $S=500$ is a reasonable choice for our experiments. Therefore, we set $S=500$ for all the four datasets. In Figures~\ref{fig:MSEVSOther} and~\ref{fig:NLPDVSOther}, those points with higher computation times belong to larger values of $m$.

For BCM, we use $k$-means clustering to partition the domain of data into $S$ local experts similar to PIC and choose the values of $S$ from the set $\{200,300,400,500,600,700\}$. The points with higher computation times belong to larger values of $S$ in Figures~\ref{fig:MSEVSOther} and~\ref{fig:NLPDVSOther}. 

LPR has three major tuning parameters, which are $S$, the number of local experts; $m$, the size of each local expert; and $R$, the size of the subset used for local hyperparameter learning. The location of $R$ data points used for local hyperparameter learning can be chosen randomly or by clustering; however, for the sake of fair comparison, we use clustering to choose these locations. Moreover, we choose the values of $S$, $m$, and $R$ from the sets $\{5,10,15,20\}$, $\{100,200,300\}$, and $\{500,1000,1500\}$, respectively. For each dataset, we fix $S$ to a value that results in better performance in terms of computation time and MSE, and choose five combinations out of the nine possible combinations of $m$ and $R$ that have different computation times. 

Last, BGP has two tuning parameters, the number of bags, $S$, and the number of data points assigned to each bag, $m$. Based on our experiments, $m$ is the major tuning parameter affecting the algorithm's computation time; therefore, we vary the values of $m$ from the set $\{500,600,700,800,900\}$ and fix the value of $S$ to a reasonable number. After varying the values of $S$ in range 10 to 80, we chose 40 as the fixed value of $S$. In Figures~\ref{fig:MSEVSOther} and~\ref{fig:NLPDVSOther}, those data points with higher computation times belong to larger size bags.

For two-dimensional dataset TCO, SPLK, DDM, PWK, and BCM perform almost the same, but they are faster and more accurate than the other algorithms as shown in Figure~\ref{fig:TimeVsMSETCO}. However, in terms of NLPD, SPLK, DDM, and PWK perform better than BCM as shown in Figure~\ref{fig:TimeVsNLPDTCO}. We attribute the BCM's higher NLPD values to underestimating the predictive variance in the BCM algorithm. Also, despite the fact that SPLK uses a low-rank covariance approximation, it performs as efficient as DDM and PWK, mainly because it creates fewer boundaries thus compensating for the inaccuracy of the low-rank approximations in the subdomains. Note that for the other datasets, we cannot compare the performance of DDM with the other competing algorithms, because DDM's implementation is restricted to one- or two-dimensional spaces. 

For three-dimensional dataset Levitus, SPLK, LPR, and PWK outperform the other algorithms in terms of MSE as shown in Figure~\ref{fig:TimeVsMSELevitus}. However, in terms of NLPD, performance of SPLK and PWK are superior (Figure~\ref{fig:TimeVsNLPDLevitus}) meaning that SPLK and PWK obtain a better goodness of fit compared to LPR.

For the five-dimensional dataset Dasilva, SPLK, PWK, and LPR outperform other competing algorithms as shown in Figures~\ref{fig:TimeVsMSEDasilva} and~\ref{fig:TimeVsNLPDDasilva}. Comparing these two algorithms however indicates that SPLK can reach higher level of accuracy in terms of MSE, while the lower predictive variance gives PWK better NLPD values. \BFcomm{ The performance of SPLK for this dataset can be better understood by noting that as the covariance decays faster in one direction, which means as $\gamma$ increases, partitioning parallel to that direction reduces the prediction accuracy close to the boundaries. This is due to the fact that the short range of covariance allows a higher degree of mismatch on the boundaries. This has been shown through a simulation study in Section 5.1 of the paper by~\cite{park2018patchwork}. However, because SPLK avoids partitioning along the direction of the largest $\gamma$, it partially reduces the degree of mismatches on the boundaries. This becomes particularly helpful when the rates of covariance decay highly differ in various directions, and as such SPLK performs better compared to the other algorithms that do not consider covariance structure in partitioning the domain. In fact, for the dataset Dasilva,  the third, fourth, and fifth directions have relatively much lower rates of covariance decay compared to the first two directions. We further investigate this hypothesis by comparing the performance of the competing algorithms on a simulated dataset having a similar covariance structure to Dasilva in Appendix~\ref{App_Dasilva_Simulation}.}

Finally, for nine-dimensional dataset Protein, SPLK, PIC, and PWK perform much better than the other algorithms as shown in Figures~\ref{fig:TimeVsMSEProtein} and~\ref{fig:TimeVsNLPDProtein}. However, similar to the analysis of TCO and Levitus, the lower NLPD values of SPLK and PWK make them more desirable than PIC.  We note that unlike the other datasets in this study, we do not set the density parameter $\kappa$ to 3, since $3^8$ control points on each boundary slow down the SPLK's performance without having a significant effect on  accuracy (see Section~\ref{sec_seb_cpdensity}). Therefore, we set $q$ to a smaller value of 2.2.
 
\subsection{Sensitivity analysis}\label{sub_sec_sen}
This section describes the sensitivity analysis we conduct on the tuning parameters of SPLK. Section~\ref{sub_sec_numberofcuts} discusses some guidelines for selecting the size of the subdomains and the density of local pseudo-inputs. Section~\ref{sec_seb_Direction} explains the significance of cutting from various directions. We discuss  the effect of the number of control points in Section~\ref{sec_seb_cpdensity}.
 
\subsubsection{Number of cuts and local pseudo-inputs}\label{sub_sec_numberofcuts}
\begin{figure}[ht]
\begin{center}
\begin{subfigure}{0.45\textwidth}
	\includegraphics[height=4cm,width=7cm]{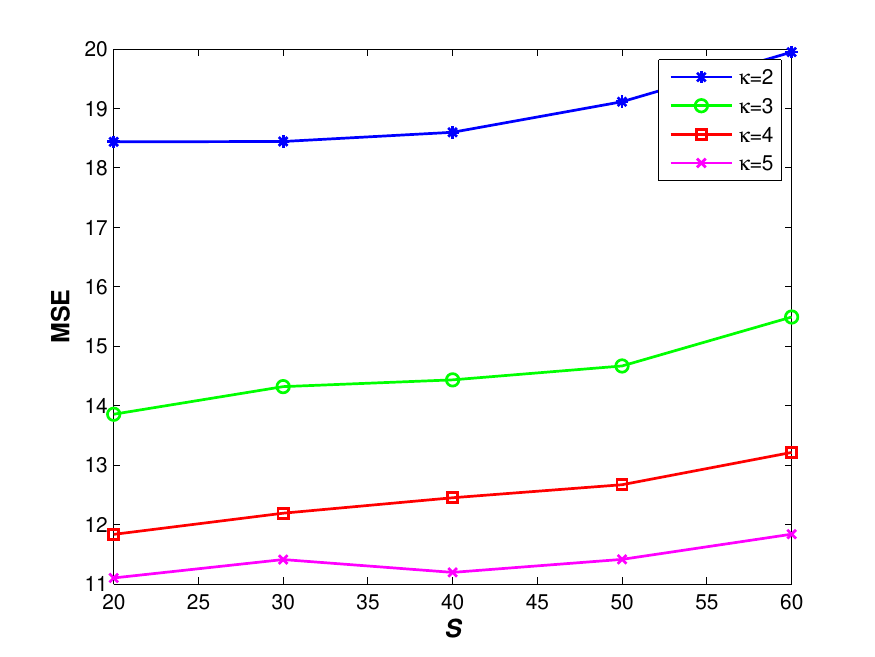}
		\caption {\tiny{TCO (MSE)}}
\label{fig:SPLKMSETCO}
\end{subfigure}
\begin{subfigure}{0.45\textwidth}
	\includegraphics[height=4cm,width=7cm]{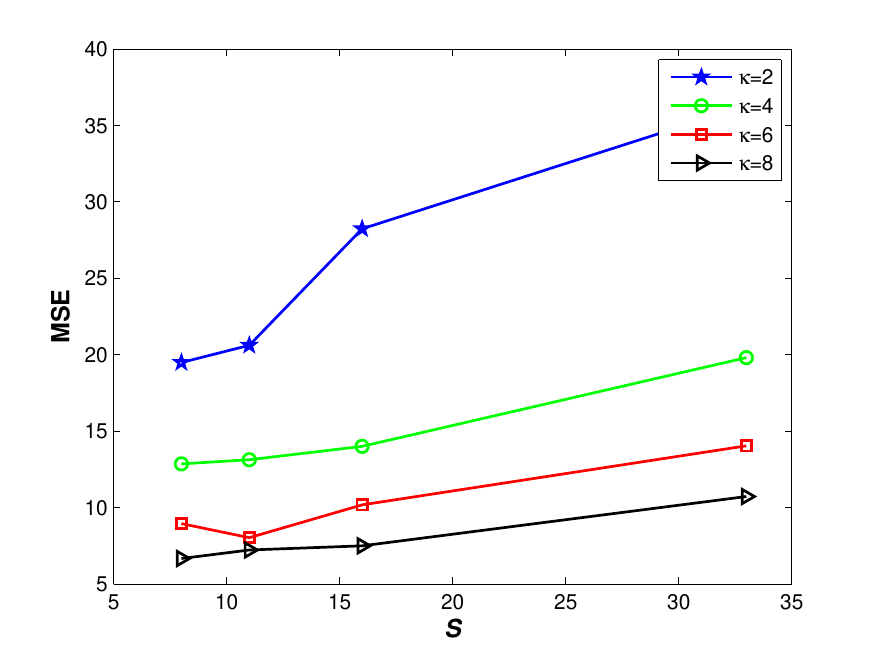}
		\caption {\tiny{Levitus (MSE)}}
\label{fig:SPLKMSELevitus}
	\end{subfigure}
	\begin{subfigure}{0.45\textwidth}
		\includegraphics[height=4cm,width=7cm]{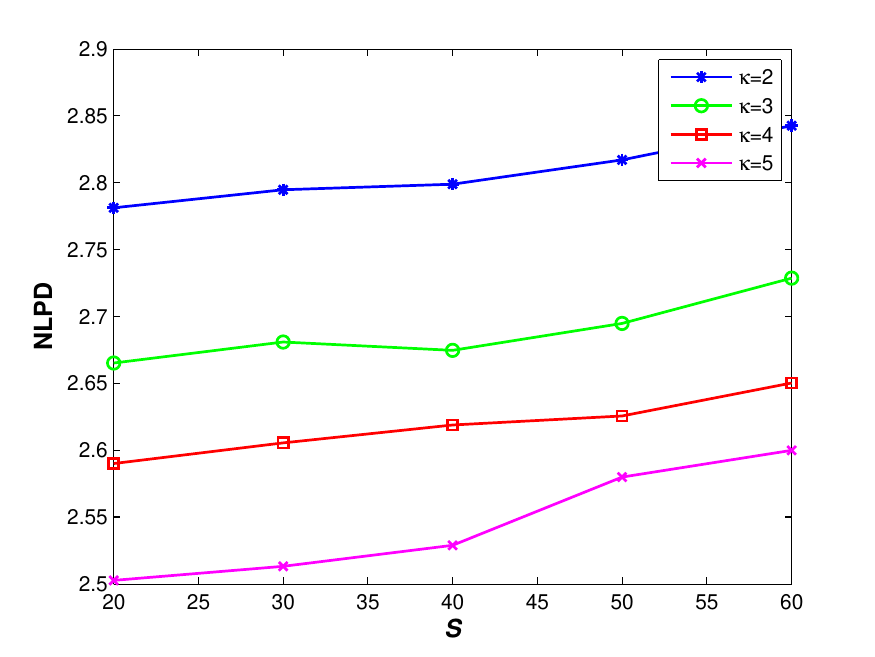}
        \caption{\tiny{TCO (NLPD)}}
\label{fig:SPLKNLPDTCO}
	\end{subfigure}
	\begin{subfigure}{0.45\textwidth}
		\includegraphics[height=4cm,width=7cm]{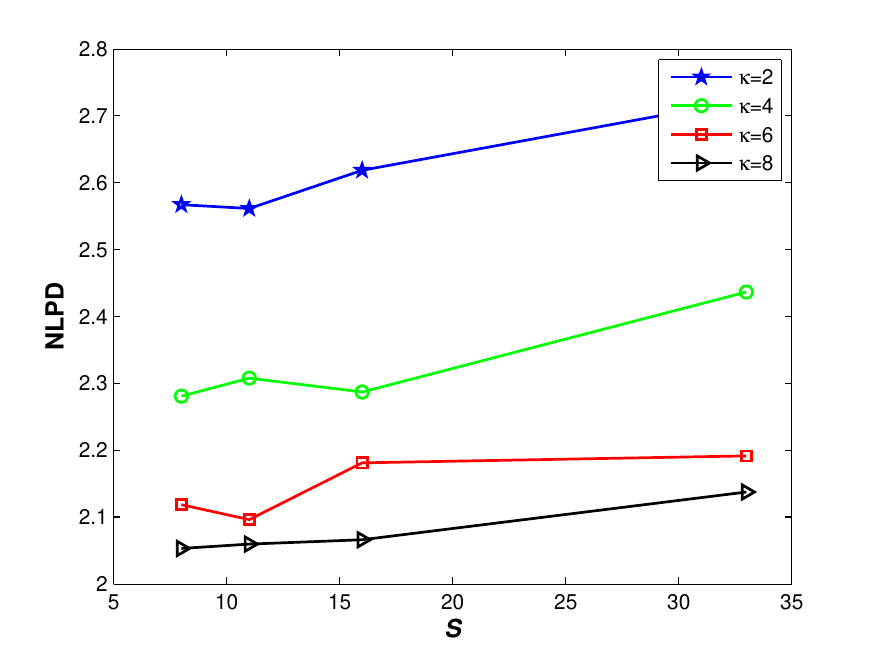}
		\caption {\tiny{Levitus (NLPD)}}
\label{fig:SPLKNLPDLevitus}
	\end{subfigure}
	\begin{subfigure}{0.45\textwidth}
		\includegraphics[height=4cm,width=7cm]{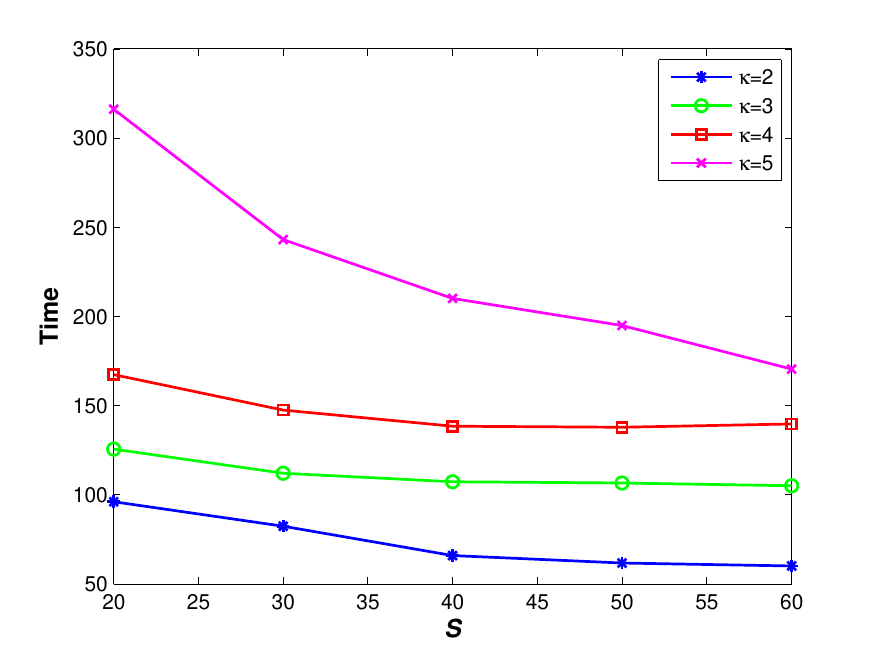}
		\caption {\tiny{TCO (Time)} }
\label{fig:SPLKTimeTCO}
	\end{subfigure}
	\begin{subfigure}{0.45\textwidth}
		\includegraphics[height=4cm,width=7cm]{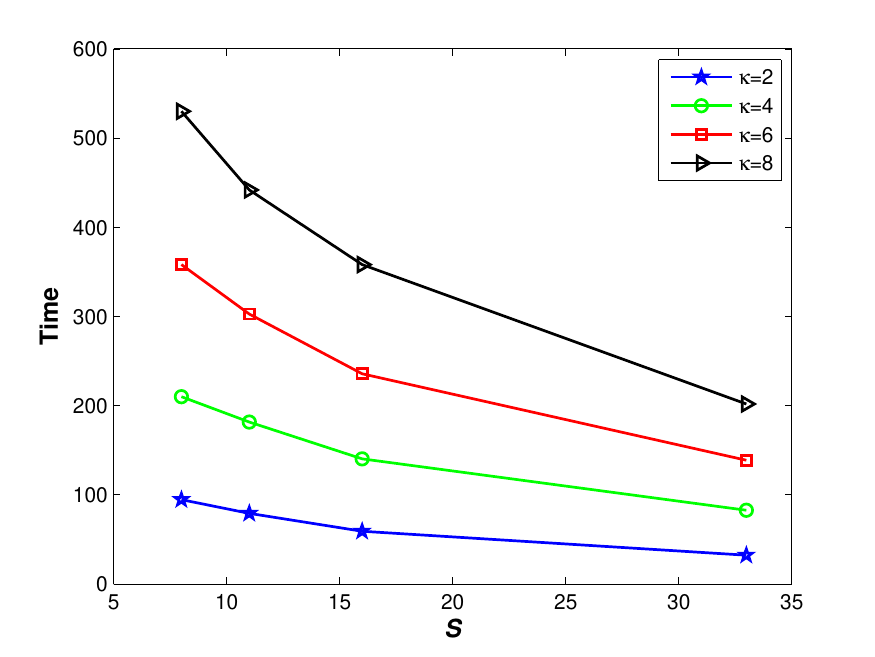}
		\caption {\tiny{Levitus (Time)}}
\label{fig:SPLKTimeLevitus}
	\end{subfigure}
\end{center}

\caption{MSE, NLPD, and computation time versus $S$. Each curve is associated with a value of $\kappa$.}
\label{fig:NS1}
\end{figure}

\begin{figure}[ht]
\begin{center}
\begin{subfigure}{0.45\textwidth}
		\includegraphics[height=4cm,width=7cm]{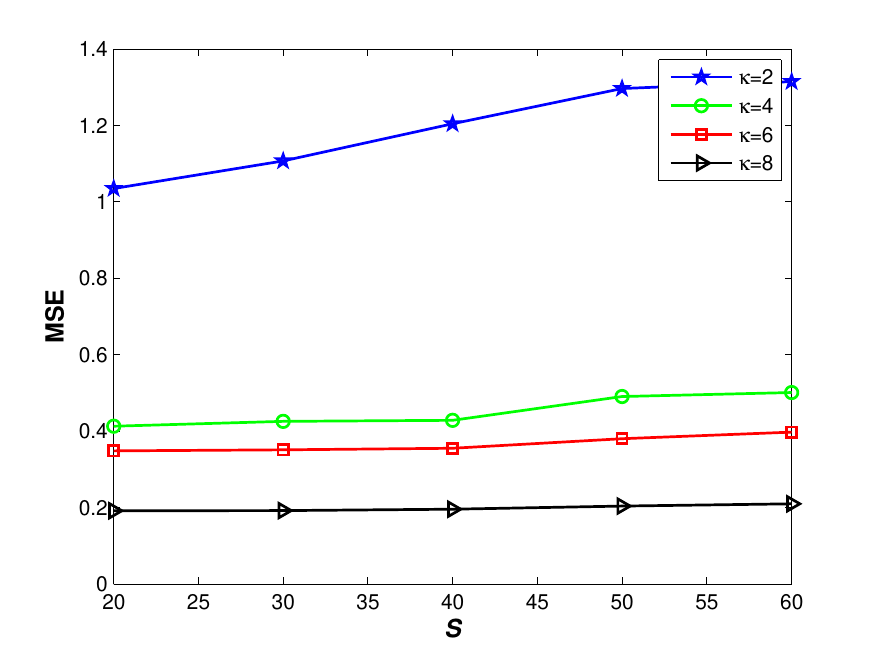}
		\caption {\tiny{Dasilva (MSE)}}
\label{fig:SPLKMSEDasilva}
	\end{subfigure}
	\begin{subfigure}{0.45\textwidth}
		\includegraphics[height=4cm,width=7cm]{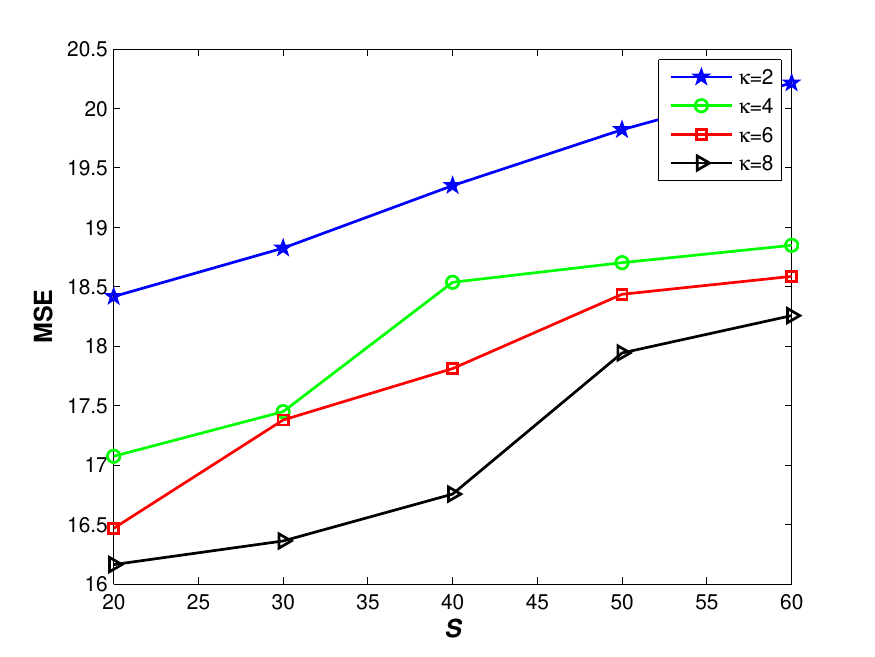}
		\caption {\tiny{Protein (MSE)}}
\label{fig:SPLKMSEProtein}
	\end{subfigure}
	\begin{subfigure}{0.45\textwidth}
		\includegraphics[height=4cm,width=7cm]{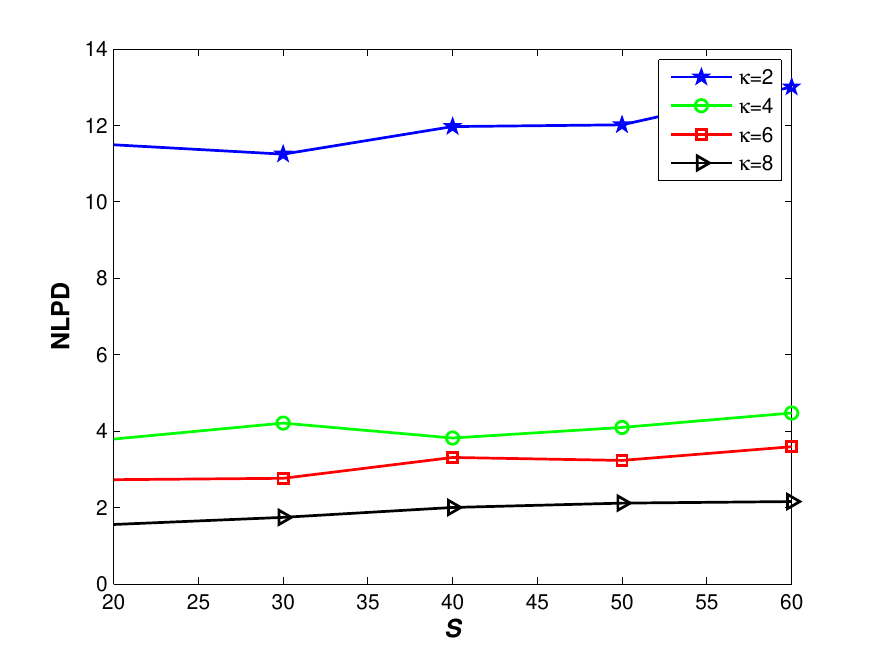}
		\caption {\tiny{Dasilva (NLPD)}}
\label{fig:SPLKNLPDDasilva}
	\end{subfigure}
	\begin{subfigure}{0.45\textwidth}
		\includegraphics[height=4cm,width=7cm]{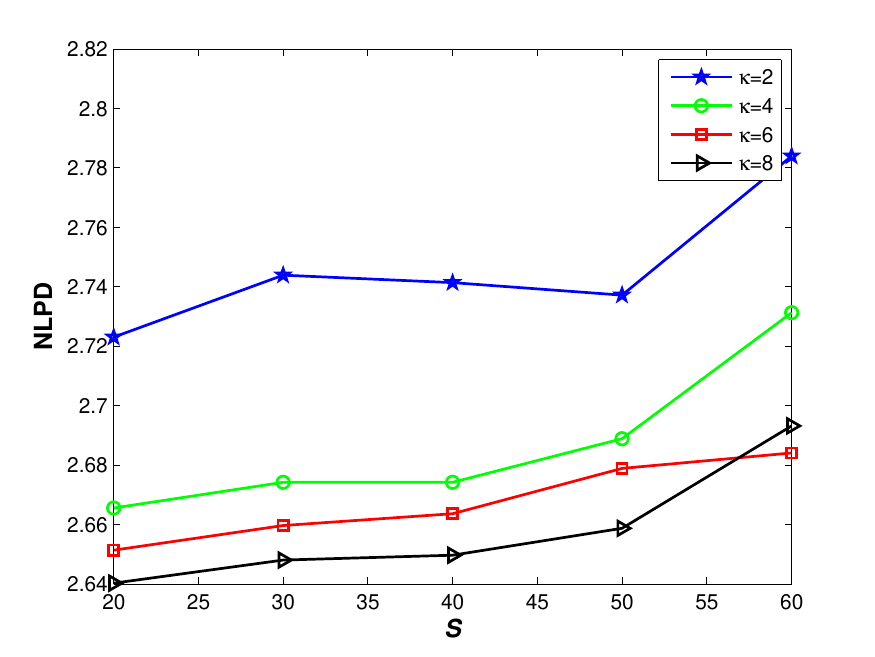}
		\caption {\tiny{Protein (NLPD)}}
\label{fig:SPLKNLPDProtein}
	\end{subfigure}
	\begin{subfigure}{0.45\textwidth}
		\includegraphics[height=4cm,width=7cm]{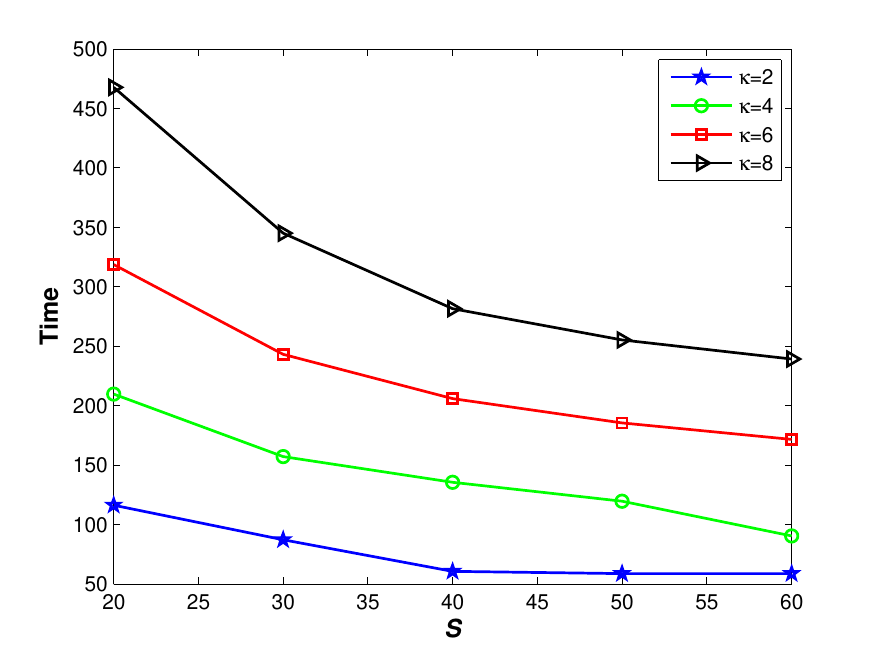}
		\caption {\tiny{Dasilva (Time)}}
\label{fig:SPLKTimeDasilva}
	\end{subfigure}
	\begin{subfigure}{0.45\textwidth}
		\includegraphics[height=4cm,width=7cm]{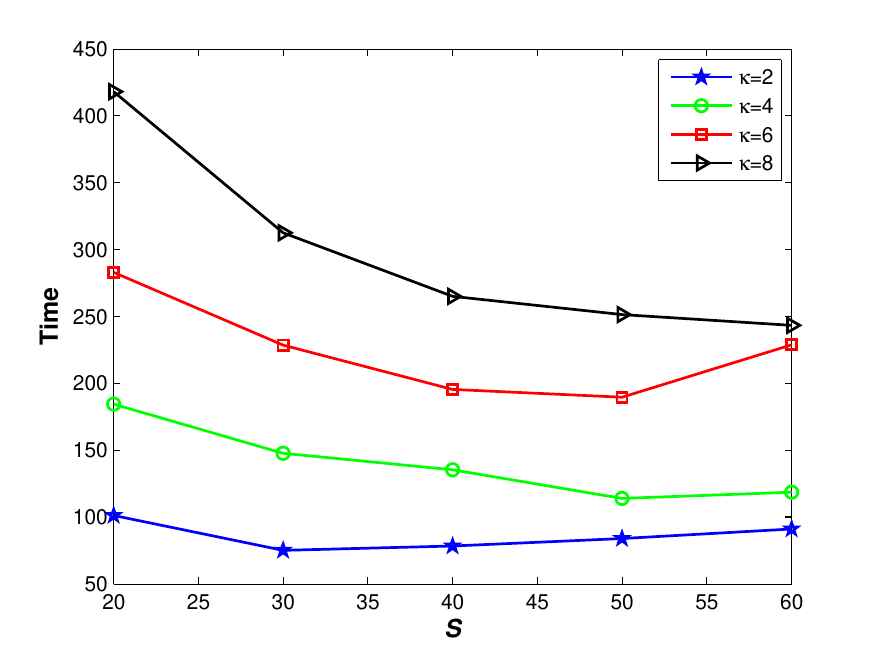}
		\caption {\tiny{Protein(Time)}}
\label{fig:SPLKTimeProtein}
	\end{subfigure}
\end{center}

\caption{MSE, NLPD, and computation time versus $S$. Each curve is associated with a value of $\kappa$.}
\label{fig:NS2}
\end{figure}

In this section, we show the trade-off between accuracy and computation time for the choices of $m$ and $S$. In our experiment, for each dataset, we vary the number of subdomains, $S$, and the density of local pseudo-inputs, $\kappa$, and use the values of MSE, NLPD, and computation time as the measures of efficiency. To illustrate the effect of various settings on the algorithm's efficiency, we plot the values of MSE, NLPD, and computation time for varying $S$ and a fixed $k$ as shown by the curves in Figures~\ref{fig:NS1} and~\ref{fig:NS2}.

Figures~\ref{fig:SPLKTimeTCO},~\ref{fig:SPLKTimeLevitus},~\ref{fig:SPLKTimeDasilva}, and~\ref{fig:SPLKTimeProtein} show that as $S$ increases, i.e., the size of the subdomains decreases, SPLK performs faster for a fixed value of $\kappa$. Moreover, the curves belonging to smaller values of $\kappa$ are always below the curves with larger values of $\kappa$, meaning that as the density of the pseudo-inputs increases, the algorithm becomes slower. Consequently, the algorithm takes longer to run by increasing the size of subdomains or the number of local pseudo-inputs.

On the other hand, Figures~\ref{fig:SPLKMSETCO},~\ref{fig:SPLKMSELevitus},~\ref{fig:SPLKMSEDasilva}, and~\ref{fig:SPLKMSEProtein} show a positive correlation between $S$ and MSE, i.e., by fixing the value of $\kappa$, SPLK performs more accurately in terms of MSE, as the size of the subdomains increases. Moreover, the curves belonging to larger values of $\kappa$ are always above the curves with lower values of $\kappa$, i.e., as  $\kappa$  increases, SPLK becomes more accurate for a fixed value of $S$. Figures~\ref{fig:SPLKNLPDTCO},~\ref{fig:SPLKNLPDLevitus},~\ref{fig:SPLKNLPDDasilva}, and~\ref{fig:SPLKNLPDProtein} show the same trend for the values of NLPD. Therefore, we conclude that our algorithm attains higher accuracy in terms of MSE and NLPD by increasing the density of local pseudo-inputs or enlarging the size of the subdomains.

In summary, by increasing the size of the subdomains or the density of local pseudo-inputs,  the algorithm’s accuracy improves, but computation time increases. 
Therefore,  we suggest using sufficiently large values of $\kappa$ in smaller subdomains, because, as shown in Figures~\ref{fig:NS1} and~\ref{fig:NS2}, the MSEs are small even with a large number of subdomains and computation times stay relatively low.

\subsubsection{Direction of cuts}\label{sec_seb_Direction}
\begin{figure}[ht]
\begin{center}
	\begin{subfigure}{0.45\textwidth}
		\includegraphics[height=4cm,width=7.5cm]{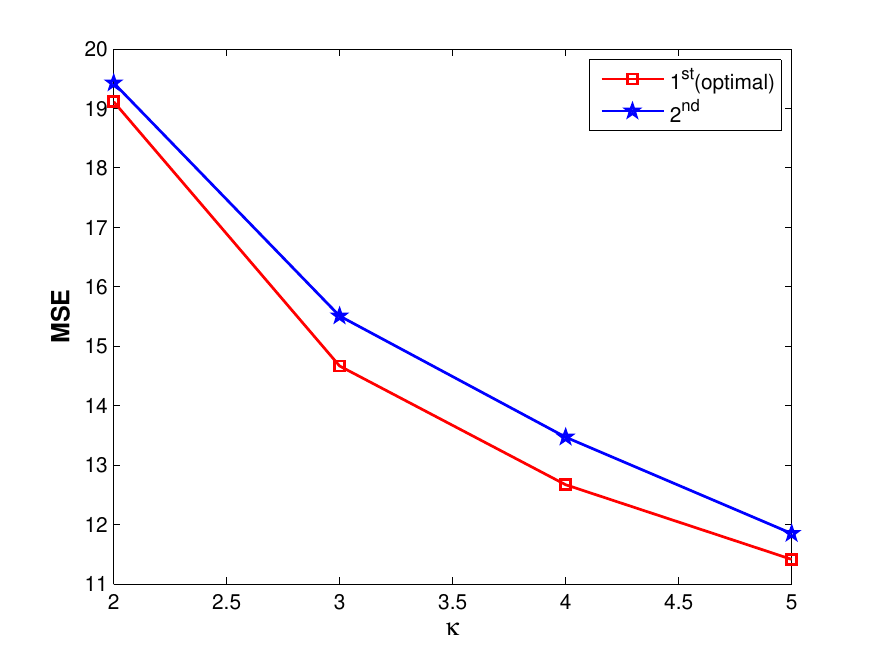}
		\caption {\tiny{TCO}:$S=30,Q=3$}
\label{fig:DirsTCO}
	\end{subfigure}
	\begin{subfigure}{0.45\textwidth}
		\includegraphics[height=4cm,width=7.5cm]{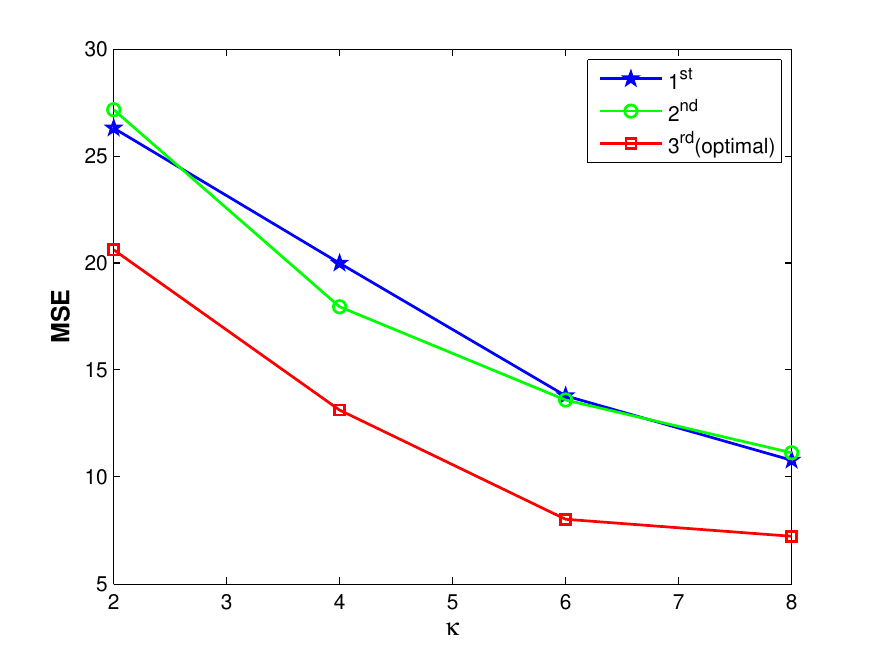}
		\caption {\tiny{Levitus}:$S=11,Q=9$ }
\label{fig:DirsLevitus}
	\end{subfigure}
\begin{subfigure}{0.45\textwidth}
		\includegraphics[height=4cm,width=7.5cm]{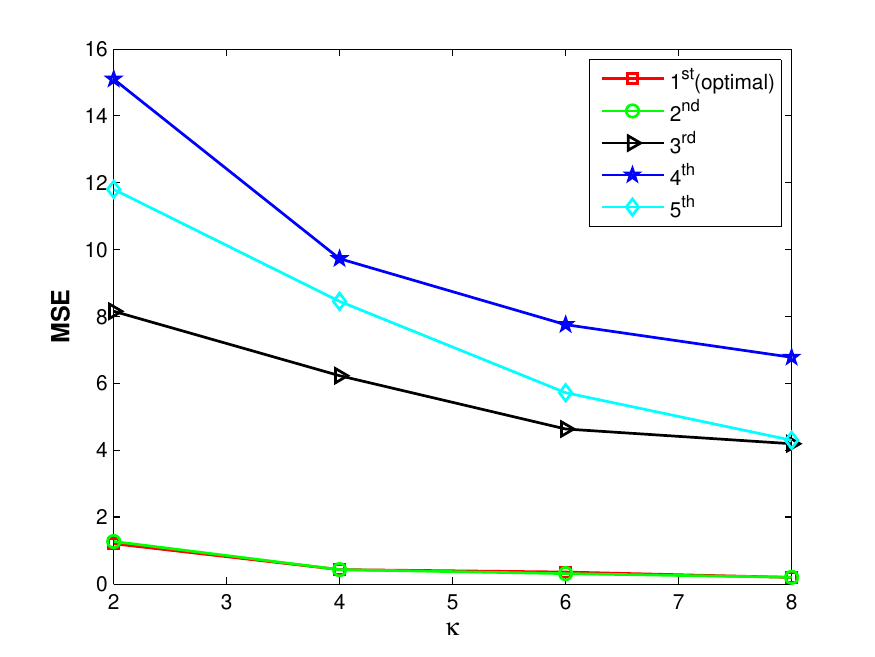}
		\caption {\tiny{Dasilva}:$S=40,Q=81$ }
\label{fig:DirsDasilva}
	\end{subfigure}
\begin{subfigure}{0.45\textwidth}
		\includegraphics[height=4cm,width=7.5cm]{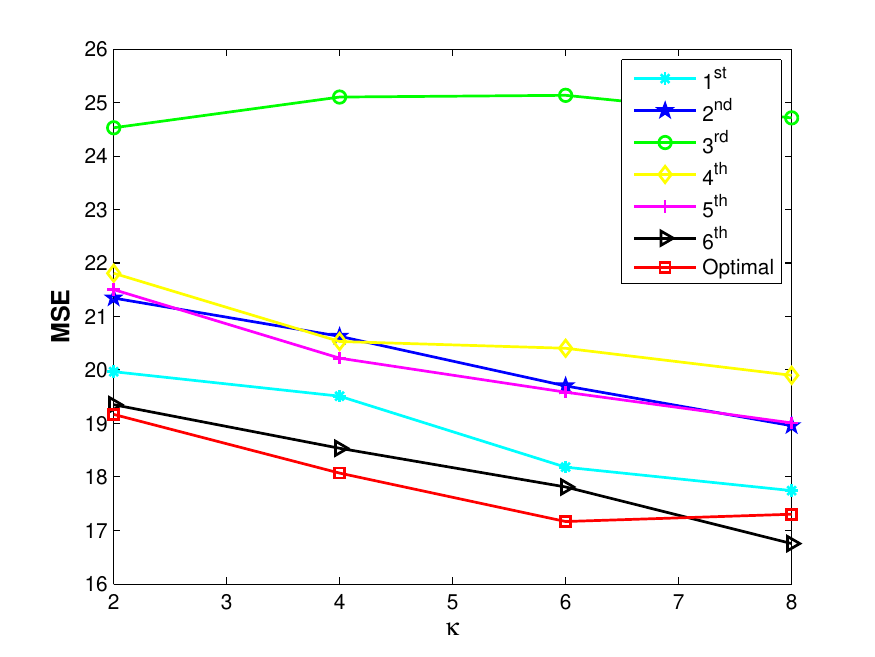}
		\caption {\tiny{Protein}:$S=40,Q=500$}
\label{fig:DirsProtein}
	\end{subfigure}
\end{center}
\caption{Effects of cutting directions on MSE for the four datasets}
\label{fig:Direction}
\end{figure}

This section demonstrates how cutting from different directions affects SPLK's performance. To discuss the significance of cutting from the direction obtained from optimization~\eqref{PreFinalLik}, we fix the value of $S$ and vary the values of $\kappa$ and the direction of cuts for each dataset, and measure the accuracy of prediction in terms of MSE. Note that since there is an infinite number of directions of cuts, for the sake of comparison, we only consider the best direction, i.e., the direction found through solving optimization problem~\eqref{PreFinalLik}, along with the directions of primary axes of the input space for each dataset. In Figure~\ref{fig:Direction}, each curve shows the trend of changes in MSE for a particular direction and the varying values of $\kappa$.

For dataset TCO, the direction of cuts is the direction of the first primary axis as shown in Figure~\ref{fig:DirsTCO}. Cutting from this direction attains higher accuracy for the varying values of $\kappa$ compared to the direction of the second primary axis.

For dataset Levitus, the direction of cuts is the direction of the third primary axis as shown in Figure~\ref{fig:DirsLevitus}. Cutting from this direction attains higher accuracy compared to the directions of the other primary axes.

For dataset Dasilva, the direction of cuts is the direction of the first primary axis as shown in Figure~\ref{fig:DirsDasilva}. Cutting from this direction attains a much higher accuracy compared to the directions of the third, forth, and fifth primary axes. However, the performance of cutting from the direction of the second primary axis is almost the same as the direction that we find through solving optimization problem~\eqref{PreFinalLik}. This can be justified by considering the objective values of optimization~\eqref{PreFinalLik} for these two directions. In fact, the objective values for the directions of the first and the second primary axes are very close and much smaller than the other directions. Therefore, we observe such a similar and much accurate performance by cutting from these two directions compared to the other directions.

Finally, for dataset Protein, the direction of cuts, which is not the direction of one of the primary axes of the input domain, is compared with the  directions of the first six primary axes as shown in Figure~\ref{fig:DirsProtein}. Cutting from the direction found by solving optimization problem~\eqref{PreFinalLik} attains a much higher accuracy compared to the directions of the second and the third primary axes, and slightly better than the direction of the sixth primary axis.

\section{Summary}\label{sec_summary}
GPR is a powerful tool in the analysis of spatial systems, but it does not scale efficiently to large datasets. In addition, many spatial datasets have highly heterogeneous covariance structures which cannot be modeled effectively with a single covariance function, and the problem is exacerbated when the spatial data contains environmental variables. This paper proposed Sparse Pseudo-input Local Kriging (SPLK), which  simultaneously addressed scalability and heterogeneity by partitioning the data domain into subdomains. The partitioning used parallel hyperplanes to create non-overlapping subdomains and fitted a sparse GPR to the data within each subdomain, which allowed the selected partitions to have large numbers of data points.  Two theorems were proposed, and an algorithm was developed to find the desired hyperplanes, which resulted in more accurate approximations of the covariance structures in each subdomain. SPLK also alleviated the discontinuity of the overall prediction surface by putting control points on the boundary of neighboring partitions. SPLK was applied to a spatial dataset with exogenous variables, two spatial datasets without exogenous variables, and a non-spatial dataset. The latter demonstrated that the methodology was general and was not restricted to spatial datasets. The results showed that SPLK maintained a good balance between prediction accuracy and computation time.

The limitations of SPLK could be better understood by using a larger number of real spatial datasets with exogenous variables. We also suggest four paths for future research. First, more flexible cuts, such as parallel hyper-curves or concentric hyper-spheres which give the same number of boundaries created by parallel hyperplanes, could be used. Second, from a theoretical perspective, theories that provide the relationship between the expected covariance function and expected error in the low-rank covariance approximation should be developed under less restrictive assumptions. Third, the value of $\kappa$, the tuning parameter that determines the density of pseudo-inputs in each subdomain, could potentially be determined with more rigorous approaches such as using an estimated rate of eigenspectrum reduction of the covariance matrix. \BFcomm{Finally, the proposed method could benefit from more sophisticated techniques for sampling control points, as opposed to using a uniform distribution. This would especially improve the model’s performance on higher-dimensional problems, in which the density of control points decreases close to the boundaries. }

\onehalfspacing
\bibliographystyle{chicago}
\bibliography{FP2015}
\newpage
\doublespacing

\begin{appendices}
\section{Solving optimization problem~\eqref{SPLkOptProb} }\label{SolutionOfLagrangianAPX}
Due to the convex objective function and affine constraints of optimization problem~\eqref{SPLkOptProb}, the duality gap between the primal and dual problems of~\eqref{SPLkOptProb} is zero by Lagrange duality principle~\citep{bazaraa2013nonlinear}. This allows us to transform the optimization problem~\eqref{SPLkOptProb} to an unconstrained optimization problem and maximize the Lagrangian of~\eqref{SPLkOptProb} instead,
\begin{eqnarray}
\max_{\mathbf{u}_s(\mathbf{x}_*),\boldsymbol{
\lambda}_s(\mathbf{x}_*)}\mathcal{L}(\mathbf{u}_s(\mathbf{x}_*),\boldsymbol{
\lambda}_s(\mathbf{x}_*))=\mathbf{u}_s(\mathbf{x}_*)^T(\tilde{\mathbf K}^s_{\mathbf{X}_s\mathbf{X}_s}+\text{diag}(\mathbf K_{\mathbf{X}_s\mathbf{X}_s}-\tilde{\mathbf K}^s_{\mathbf{X}_s\mathbf{X}_s})+\sigma_s^2 \mathbf I_s) \mathbf{u}_s(\mathbf{x}_*)\label{PreLagrange}\\-2 \mathbf{u}_s(\mathbf{x}_*)^T \tilde{\mathbf k}^s_{\mathbf{X}_s\mathbf{x}_*}
-\sum_{i=1:|\mathbf{B}_s|} \lambda_{is}(\mathbf{x}_*) (\mathbf{u}_s(\mathbf{b}_i)^T\mathbf{y}_s- \mathcal{R}(\mathbf{b}_i)),\nonumber
\end{eqnarray}
where $|\mathbf{B}_s|$ is the number of all the control points located on the boundaries of subdomain $\Omega_s$, and $\boldsymbol{
\lambda}_s(\mathbf{x}_*)=[ \lambda_{1s}(\mathbf{x}_*),\ldots, \lambda_{|\mathbf{B}_s|s}(\mathbf{x}_*)]^T$ is the vector of the Lagrange multipliers.

Assuming $\mathbf{u}_s(\mathbf{x}_*)$ depends on the covariance between $\mathbf{x}_*$ and $\mathbf{X}_s$, and $\lambda_{is}(\mathbf{x}_*)$ depends on the covariance of $\mathbf{b}_i$ and $\mathbf{x}_*$, we write $\mathbf{u}_s(\mathbf{x}_*)= \mathbf{H}_s\tilde{\mathbf k}^s_{\mathbf{X}_s\mathbf{x}_*}$ and $\lambda_{is}(\mathbf{x}^*)=\beta_{is} \tilde{ k}^s_{\mathbf{b}_i\mathbf{x}_*}$ as suggested in~\citep{Park}, where $\mathbf{H}_j$ is a squared matrix with size equal to the number of data points in $\Omega_s$, and $\beta_{is}$ is the Lagrange parameter associated with $\lambda_{is}$ that does not depend on $\mathbf{x}_*$. Consequently, we rewrite Lagrangian~\eqref{PreLagrange} as
\begin{eqnarray}
\max_{\mathbf{H}_s,\boldsymbol{\beta}_s}\mathcal{L}(\mathbf{H}_s,\boldsymbol{\beta}_s )=\tilde{\mathbf k}^s_{\mathbf{x}_*\mathbf{X}_s}\mathbf{H}_s^T(\tilde{\mathbf K}^s_{\mathbf{X}_s\mathbf{X}_s}+\text{diag}(\mathbf K_{\mathbf{X}_s\mathbf{X}_s}-\tilde{\mathbf K}^s_{\mathbf{X}_s\mathbf{X}_s})+\sigma_s^2 \mathbf I_s)\mathbf{H}_s\tilde{\mathbf k}^s_{\mathbf{X}_s\mathbf{x}_*}\label{lagrangian}\\
-2 \tilde{\mathbf k}^s_{\mathbf{x}_*\mathbf{X}_s}\mathbf{H}_s^T \tilde{\mathbf k}^s_{\mathbf{X}_s\mathbf{x}_*}
- \tilde{\mathbf k}^s_{\mathbf{x}_*\mathbf{B}_s} \boldsymbol{\beta}_{s} (\tilde{\mathbf K}^s_{\mathbf{B}_s\mathbf{X}_s}\mathbf{H}_s^T\mathbf{y}_s- \mathbf{r}_s),\nonumber
\end{eqnarray}
where $\boldsymbol{\beta}_s$ is a diagonal matrix with diagonal elements $\beta_{1s},\ldots,\beta_{|\mathbf{B}_s|s}$, and $\mathbf{r}_s=[\mathcal{R}(\mathbf{b}_1),\ldots,\mathcal{R}(\mathbf{b}_{|\mathbf{B}_s|})]^T$ is the vectors of boundary values of $\Omega_s$. 

Due to convexity of function~\eqref{lagrangian} we can calculate the optimal values of $\mathbf{H}_s$ and $\boldsymbol{\beta}_s$ analytically by writing out the first order necessary conditions,
\begin{eqnarray}
&&\frac{d\mathcal{L}(\mathbf{H}_s,\boldsymbol{\beta}_s )}{d\mathbf{H}_s}=2(\mathbf{G}_s\mathbf{H}_s-\mathbf{I}_s)\tilde{\mathbf k}^s_{\mathbf{X}_s\mathbf{x}_*}\tilde{\mathbf k}^s_{\mathbf{x}_*\mathbf{X}_s}- \mathbf{y}_s \tilde{\mathbf k}^s_{\mathbf{x}_*\mathbf{B}_s} \boldsymbol{\beta}_{s} \tilde{\mathbf K}^s_{\mathbf{B}_s\mathbf{X}_s}=0,\label{condition1}\\
&&\frac{d\mathcal{L}(\mathbf{H}_s,\boldsymbol{\beta}_s )}{d\beta_{is}}=\tilde{\mathbf k}_{\mathbf{b}_i\mathbf{X}_s}\mathbf{H}_s^T\mathbf{y}_j-r_{is}=0 \;\;\; \forall i\in [|\mathbf{B}_s|],\label{condition2}
\end{eqnarray}
where $\mathbf{G}_s=(\tilde{\mathbf K}^s_{\mathbf{X}_s\mathbf{X}_s}+\text{diag}(\mathbf K_{\mathbf{X}_s\mathbf{X}_s}-\tilde{\mathbf K}^s_{\mathbf{X}_s\mathbf{X}_s})+\sigma_s^2 \mathbf I_s)$, and $r_{is}$ is the $i^{\text{th}}$ element of the vector $\mathbf{r}_s$. Reordering  equation~\eqref{condition1},
\begin{eqnarray}
(\tilde{\mathbf k}^s_{\mathbf{x}_*\mathbf{X}_s}+0.5(\tilde{\mathbf k}^j_{\mathbf{x}_*\mathbf{X}_s}\tilde{\mathbf k}^j_{\mathbf{X}_s\mathbf{x}_*})^{-1}\tilde{\mathbf k}^j_{\mathbf{x}_*\mathbf{X}_s}\tilde{\mathbf K}^j_{\mathbf{X}_s\mathbf{B}_s}\boldsymbol{\beta}_{s}\tilde{\mathbf k}^j_{\mathbf{B}_s\mathbf{x}_*}\mathbf{y}_s^T)\mathbf{G}_s^{-1}\mathbf{y}_s=\tilde{\mathbf k}^s_{\mathbf{x}_*\mathbf{X}_s}\mathbf{H}_s^T\mathbf{y}_s,\label{condition1Reorder}
\end{eqnarray}
and evaluating it at the boundary locations gives the system of equations with $|\mathbf{B}_s|$ equations and Lagrangian parameters,
\begin{eqnarray}
(\tilde{\mathbf k}^s_{\mathbf{b}_i\mathbf{X}_s}+0.5(\tilde{\mathbf k}^s_{\mathbf{b}_i\mathbf{X}_s}\tilde{\mathbf k}^s_{\mathbf{X}_s\mathbf{b}_i})^{-1}\tilde{\mathbf k}^s_{\mathbf{b}_i\mathbf{X}_s}\tilde{\mathbf K}^s_{\mathbf{X}_s\mathbf{B}_s}\boldsymbol{\beta}_{s}\tilde{\mathbf k}^s_{\mathbf{B}_s\mathbf{b}_i}\mathbf{y}_s^T)\mathbf{G}_s^{-1}\mathbf{y}_s=r_{is}\;\; \forall i\in [|\mathbf{B}_s|].\label{LagrangeSysEQ}
\end{eqnarray}

After some simple matrix algebra, we obtain the solution to the system of linear equations~\eqref{LagrangeSysEQ},
\begin{eqnarray}
\boldsymbol{\beta}_{s}=\frac{\mathbf{I}_s(\mathbf{r}_s-\tilde{\mathbf K}^s_{\mathbf{B}_s\mathbf{X}_s}\mathbf{G}_s^{-1}\mathbf{y}_s)\{[(\text{diag}(\tilde{\mathbf K}^s_{\mathbf{B}_s\mathbf{X}_s}\tilde{\mathbf K}^s_{\mathbf{X}_s\mathbf{B}_s}))^{-1}(\tilde{\mathbf K}^s_{\mathbf{B}_s\mathbf{X}_s}\tilde{\mathbf K}^s_{\mathbf{X}_s\mathbf{B}_s})]\circ\mathbf K^s_{\mathbf{B}_s\mathbf{B}_s}\}^{-1}}
{0.5\mathbf{y}_s^T\mathbf{G}_s^{-1}\mathbf{y}_s}.\label{LagrangeSolution}
\end{eqnarray}

Using the values of $\boldsymbol{\beta}_{s}$ from~\eqref{LagrangeSolution}, we can easily obtain the solution to $\mathbf{u}(\mathbf{x}_*)$ from~\eqref{condition1},
\begin{eqnarray}
\mathbf{u}^*_s(\mathbf{x}_*)=\mathbf{H}_s\tilde{\mathbf k}^s_{\mathbf{X}_s\mathbf{x}_*}=\mathbf{G}_s^{-1}(\tilde{\mathbf k}^s_{\mathbf{X}_s\mathbf{x}_*}+\mathbf{w}_s),
\end{eqnarray}
where $\mathbf{w}_s=0.5(\tilde{\mathbf k}^s_{\mathbf{x}_*\mathbf{X}_s}\tilde{\mathbf k}^s_{\mathbf{X}_s\mathbf{x}_*})^{-1}\mathbf{y}_s \tilde{\mathbf k}^s_{\mathbf{x}_*\mathbf{B}_s} \boldsymbol{\beta}_{s} \tilde{\mathbf K}^s_{\mathbf{B}_s\mathbf{X}_s}\tilde{\mathbf k}^s_{\mathbf{X}_s\mathbf{x}_*}$.

\section{Derivation of low-rank covariance approximation error}\label{ErrorFuncAPX}
We follow the procedure proposed in~\citep{smola2000sparse} to derive the low-rank covariance approximation error in each subdomain $\Omega_s$. In this derivation, given the covariance function $\phi(\cdot,\cdot):\Omega_s\times\Omega_s\rightarrow\mathbb{R}$ as a symmetric positive semidefinite kernel, we intend to approximate the kernel $\phi(\mathbf{x},\cdot):\Omega_s\rightarrow\mathbb{R}^{\Omega_s}$ centered at $\mathbf z\in \Omega_s$ as a linear combination of kernels centered at each element of $\mathbf{X}_s$, i.e.,
\begin{eqnarray}
\phi(\mathbf{z},\cdot)\approx\sum_{i\in [m_s]}c_i \phi(\tilde{\mathbf x}_i,\cdot).\label{ApproxByLinComb}
\end{eqnarray}
To this end, 
let $\mathcal{H}$ be a reproducing kernel Hilbert space (RKHS) that is defined as the space of functions constructed by the span of $\phi(\mathbf{x},\cdot)$ centered at a finite number of elements of $\Omega_s$, i.e.,
\[
\Bigg\{\sum_{i\in [n] }\alpha_i \phi({\mathbf x}_i,\cdot): n\in \mathbb{N}, \mathbf{x}_i \in \Omega_s, c_i \in \mathbf{R} 
\Bigg\}.\]
$\mathcal{H}$ is also equipped with the inner product
\begin{eqnarray}
\bigg<\sum_{i\in [n_1] }\alpha_i \phi({\mathbf x}_i,\cdot),\sum_{j\in [n_2] }\beta_j \phi({\mathbf x}_j,\cdot)\bigg>_{\mathcal{H}}=\sum_{i\in [n_1] }\sum_{j\in [n_2] }\alpha_i\beta_j\phi(\mathbf{x}_i,\mathbf{x}_j),\label{InnerProd}
\end{eqnarray}
which, for any function $f\in \mathcal{H}$, induces the norm
\begin{eqnarray}
||f||_{\mathcal{H}}^2=<f,f>_{\mathcal{H}}.\label{NormOfRKHS}
\end{eqnarray}
Given such $\mathcal{H}$, a natural criterion to find an approximation for the covariance function is to minimize the norm of function $\phi(\mathbf{z},\cdot)-\sum_{i\in [m_s]}c_i \phi(\tilde{\mathbf x}_i,\cdot)$, which belongs to $\mathcal{H}$, that is 
\begin{eqnarray}
\min_{\mathbf c}\left\| \phi(\mathbf z,\cdot) - \sum_{i\in [m_s]}c_i \phi(\tilde{\mathbf x}_i,\cdot)\right\|^2_{\mathcal{H}},\label{MinimizingCovApproxEr}
\end{eqnarray}
where $\mathbf{c}=[c_1,\ldots,c_{m_s}]^T$.
Assuming $\phi(\mathbf{z},\mathbf{z})=h$, objective function~\eqref{MinimizingCovApproxEr} can be expanded after plugging in~\eqref{InnerProd} and~\eqref{NormOfRKHS} as
\[\min_{\mathbf c}\;h-2\mathbf{c}^T\mathbf k_{\tilde{\mathbf{X}}_s\mathbf z}+\mathbf{c}^T \mathbf K_{\tilde{\mathbf{X}}_s\tilde{\mathbf{X}}_s} \mathbf{c},\]
which has the solution $\mathbf c^*=\mathbf K_{\tilde{\mathbf{X}}_s\tilde{\mathbf{X}}_s}^{-1} \mathbf k_{\tilde{\mathbf{X}}_s\mathbf z}$. Therfore, the approximation of $\phi(\mathbf{z},\cdot)$ becomes $\mathbf k_{\mathbf z\tilde{\mathbf{X}}_s}\mathbf K_{\tilde{\mathbf{X}}_s\tilde{\mathbf{X}}_s}^{-1} \mathbf k_{\tilde{\mathbf{X}}_s\mathbf z}$, and the error of covariance approximation becomes
\[h-\mathbf k_{\mathbf z\tilde{\mathbf{X}}_s}\mathbf K_{\tilde{\mathbf{X}}_s\tilde{\mathbf{X}}_s}^{-1} \mathbf k_{\tilde{\mathbf{X}}_s\mathbf z}.\]

We finally note that using $\mathbf z=\mathbf x_i$ for all $\mathbf x_i \in \mathbf X_s$ in objective function~\eqref{MinimizingCovApproxEr} and minimizing the sum over all terms obtains $\mathbf K_{\mathbf{X}_s\tilde{\mathbf{X}}_s}\mathbf K_{\tilde{\mathbf{X}}_s\tilde{\mathbf{X}}_s}^{-1} \mathbf K_{\tilde{\mathbf{X}}_s\mathbf{X}_s}$, which is the low-rank approximation of $\mathbf K_{\mathbf{X}_s\mathbf{X}_s}$ in equation~\eqref{eq_low_rank_cov}. 

\section{Proof of Theorems }\label{Theorem1APX}

\subsection{Proof of Proposition~\ref{LBPower}}
\begin{proof}
For any $i\in [m_s]$, let $\mathbf{u}_i$ denote the covariance vector between $\mathbf{z}$ and the first $i$ elements of $\tilde{\mathbf{X}}_s$, and let $\mathbf{v}_i$ denote the covariance vector between the $(i+1)^{\textrm{th}}$ element of $\tilde{\mathbf{X}}_s$ and the first $i$ elements of $\tilde{\mathbf{X}}_s$. That is, $\mathbf{u}_i=[\phi(\mathbf{z},\tilde{\mathbf{x}_1}),\ldots,\phi(\mathbf{z},\tilde{\mathbf{x}}_i)]^T$, and $\mathbf{v}_i=[\phi(\tilde{\mathbf{x}}_{i+1},\tilde{\mathbf{x}}_1),\ldots,\phi(\tilde{\mathbf{x}}_{i+1},\tilde{\mathbf{x}}_i)]^T$. Also let $\mathbf{K}_i$ denote the covariance matrix between the first $i$ elements of $\tilde{\mathbf{X}}_s$ themselves.
We now prove by induction on $i$. For the base case, i.e, $i=1$, the claim clearly holds, 
\begin{eqnarray}\label{Basecase}
\mathbb{E}_{\Omega_s}(\mathbf u^T_1\mathbf K_1^{-1} \mathbf u_1)=\mathbb{E}_{\Omega_s}(\phi(\mathbf z,\tilde{\mathbf{x}}_1)\phi (\tilde{\mathbf{x}}_1,\tilde{\mathbf{x}}_1)^{-1}\phi(\mathbf z,\tilde{\mathbf{x}}_1))=\frac{1}{h}\mathbb{E}_{\Omega_s}(\phi^2(\mathbf z,\tilde{\mathbf{x}}_1))=\frac{1}{h}\mathbb{E}_{\Omega_s}(\phi^2(\mathbf{x},\mathbf{x}')).
\end{eqnarray}
Suppose the claim holds for $m_s-1$, we show that it also holds for $m_s$. Expanding $\mathbf u^T_{m_s}\mathbf K_{m_s}^{-1} \mathbf u_{m_s}$ gives
\begin{subequations}\label{PowerFunExpanded}
\begin{align}
&\mathbf u^T_{m_s}\mathbf K_{m_s}^{-1} \mathbf u_{m_s}=
\begin{bmatrix}
\mathbf u^T_{m_s-1}&
\phi(\mathbf{z},\tilde{\mathbf{x}}_{m_s})
\end{bmatrix}
\begin{bmatrix}
\mathbf K_{m_s-1} & \mathbf v_{m_s-1}\\
\mathbf v^T_{m_s-1}& h
\end{bmatrix}^{-1}
\begin{bmatrix}
\mathbf u_{m_s-1}\\
\phi(\mathbf{z},\tilde{\mathbf{x}}_{m_s})
\end{bmatrix}\\
&=\begin{bmatrix}
\mathbf u^T_{m_s-1}&
\phi(\mathbf{z},\tilde{\mathbf{x}}_{m_s})
\end{bmatrix}
\begin{bmatrix}
\mathbf K_{m_s-1}^{-1}+c\mathbf K_{m_s-1}^{-1}\mathbf v_{m_s-1}\mathbf v^T_{m_s-1}\mathbf K_{m_s-1}^{-1} & -c \mathbf K_{m_s-1}^{-1} \mathbf v_{m_s-1}\\
-c\mathbf v^T_{m_s-1} \mathbf K_{m_s-1}^{-1}& c
\end{bmatrix}
\begin{bmatrix}
\mathbf u^T_{m_s-1}\\
\phi(\mathbf{z},\tilde{\mathbf{x}}_{m_s})
\end{bmatrix}\label{InvBlockPowerFun}\\
&=\mathbf u^T_{m_s-1}\mathbf K_{m_s-1}^{-1} \mathbf u_{m_s-1}+\frac{(\mathbf v^T_{m_s-1}\mathbf K_{m_s-1}^{-1} \mathbf u_{m_s-1})^2+\phi^2(\mathbf{z},\tilde{\mathbf{x}}_{m_s})-2\mathbf v^T_{m_s-1}\mathbf K_{m_s-1}^{-1} \mathbf u_{m_s-1}\phi(\mathbf{z},\tilde{\mathbf{x}}_{m_s})}{c}\\
&=\mathbf u^T_{m_s-1}\mathbf K_{m_s-1}^{-1} \mathbf u_{m_s-1}+\frac{(\mathbf v^T_{m_s-1}\mathbf K_{m_s-1}^{-1} \mathbf u_{m_s-1}-\phi(\mathbf{z},\tilde{\mathbf{x}}_{m_s}))^2}{c}\\
&\geq \mathbf u^T_{m_s-1}\mathbf K_{m_s-1}^{-1} \mathbf u_{m_s-1}.
\end{align}
\end{subequations}
where equality~\eqref{InvBlockPowerFun} follows from the block matrix inversion lemma~\citep{hager1989updating}, and $c=(h-\mathbf v^T_{m_s-1}\mathbf K_{m_s-1}^{-1}\mathbf v_{m_s-1})^{-1}$, which is always non-negative. 

By~\eqref{PowerFunExpanded} and the induction step,
\begin{eqnarray}
\mathbb{E}_{\Omega_s}(\mathbf u^T_{m_s}\mathbf K_{m_s}^{-1} \mathbf u_{m_s})\geq \mathbb{E}_{\Omega_s}(\mathbf u^T_{m_s-1}\mathbf K_{m_s-1}^{-1} \mathbf u_{m_s-1})\geq \frac{1}{h}\mathbb{E}_{\Omega_s}(\phi^2(\mathbf{x},\mathbf{x}')).
\end{eqnarray}
\end{proof}

\subsection{Proof of Theorem~\ref{TheoremParandRec}}
First, we prove the following lemma 
\begin{lemma}\label{OrderingLemma}
For the random variables $z_1\sim \mathcal{U}(a,a+e)$ and $z_2\sim \mathcal{U}(b,b+e)$, where $a\leq b$ and $a,b,e\geq 0$, define $v=(z_1-z_2)^2$. Then $\mathbb{E}_v(\exp(-cv))\leq \mathbb{E}_{v}(\exp(-cv)\mid a=b)$ for any $c>0$.
\end{lemma}
\begin{proof}
Let $z=z_1-z_2$, then by convolution of probability distributions, we have:
\begin{eqnarray}
f_z(t)=\int^{+\infty}_{-\infty}f_{z_1}(t+z_2)f_{z_2}(z_2)dz_2=\frac{1}{e}\int^{b+e}_{b}f_{z_1}(t+z_2)dz_2,\label{Conv1}
\end{eqnarray}
where the last equation follows from the fact that $f_{z_2}=\frac{1}{e}$ if $b \leq z_2\leq b+e$. Note that the integrand $f_{z_1}(z+z_2)$ is zero unless $a\leq t+z_2\leq a+e$, which implies $a-t\leq z_2 \leq a+e-t$. Figure~\ref{fig:ConvPic} shows the region defined by $a-t\leq z_2 \leq a+e-t$ and  $b \leq z_2\leq b+e$, for the case that $a+b<e$ and $a+e>b$. In the both cases, integration~\eqref{Conv1} can be calculated as follows:
\begin{eqnarray}
\begin{aligned}
 &   f_z(t)= 
\begin{cases}
  \frac{1}{e^2}\int_{a-e-d}^{t}dz_2 &  a-b-e\leq t < a-b\\
 \frac{1}{e^2}\int^{a-e}_{t}dz_2& a-b\leq t \leq a-b+e
\end{cases} 
 & = 
\begin{cases}
  \frac{1}{e^2}(t+b-a+e)&  a-b-e\leq t < a-b\\
 \frac{-1}{e^2}(t+b-a-e)& a-b\leq t \leq a-b+e.
\end{cases}
\end{aligned}
\end{eqnarray}

\begin{figure}[H]
\begin{center}
	\begin{subfigure}{0.8\textwidth}
\includegraphics[height=5cm,width=15cm]{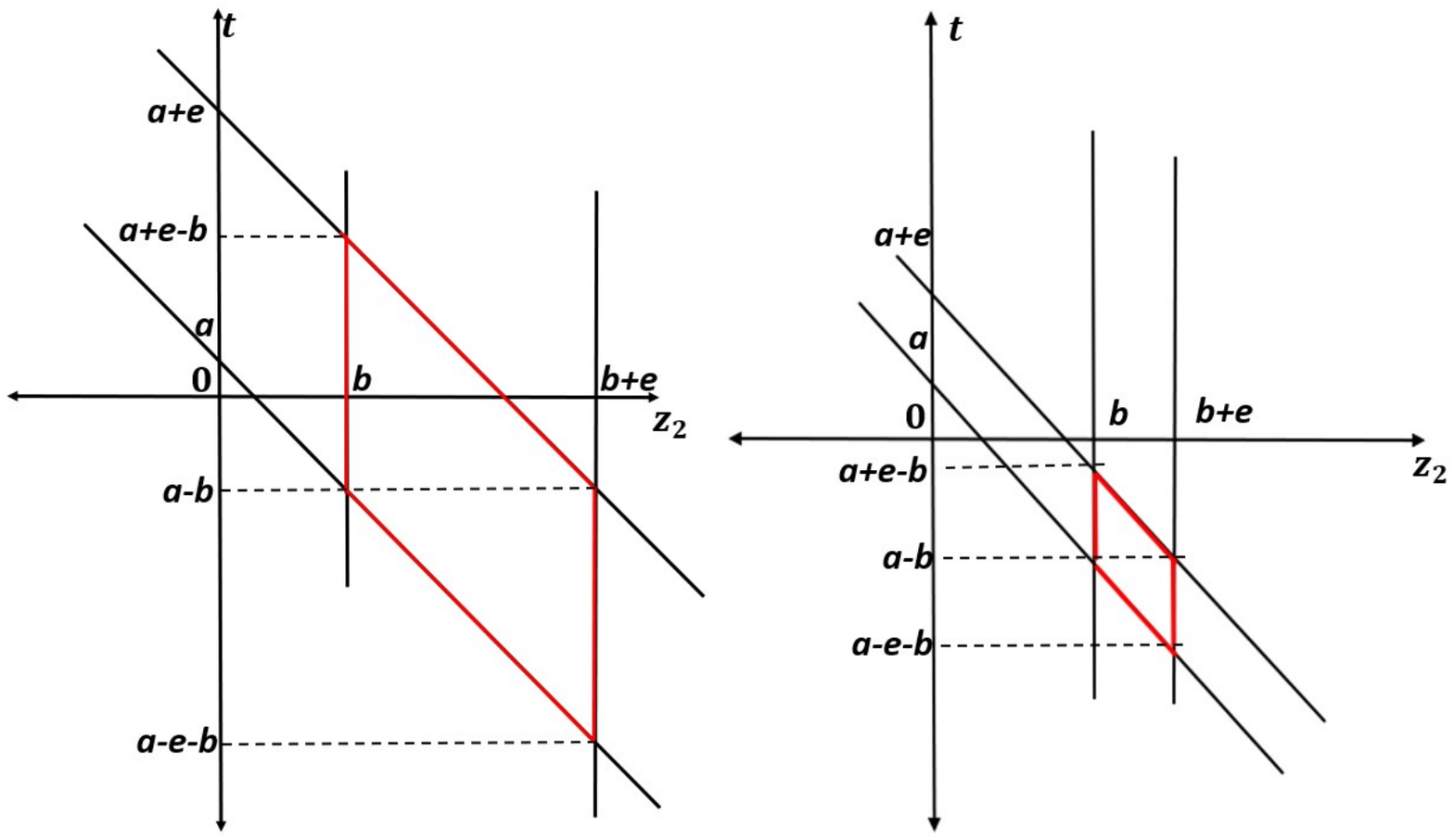}
	\end{subfigure}
\end{center}
\caption{The region defined by $a-t\leq z_2 \leq a+e-t$ and  $b \leq z_2\leq b+e$. Left panel corresponds to the case when $a+b>e$ and right panel corresponds to the case when $a+e<b$. }
\label{fig:ConvPic}
\end{figure}

Hence, $F_v(t)=p(v\leq t)=p(z^2\leq t)=p(\sqrt{t}\leq z \leq \sqrt{t})$ can be written as
\begin{eqnarray}
\begin{aligned}
 &&&   F_v(t)= 
\begin{cases}
  \frac{2\sqrt{t}}{e^2}(a-b+e)&  0\leq \sqrt{t}< b-a,\\
 \frac{1}{e^2}(2\sqrt{t}e-t-(a-b)^2)&  b-a\leq \sqrt{t} < a-b+e,\\
1-\frac{1}{2e^2}(\sqrt{t}+a-b-e)^2 & a-b+e\leq \sqrt{t}\leq b-a+e.
\end{cases} 
\end{aligned}\label{DistS}
\end{eqnarray}
Moreover, $G_{v}(t)=p(v\leq t\mid a=b)=p(z^2\leq t\mid a=b)=p(\sqrt{t}\leq z \leq \sqrt{t}\mid a=b)$ can be derived by setting $a=b$ in CDF~\eqref{DistS}
\begin{eqnarray}
G_{v}(t)=\frac{1}{e^2}(2\sqrt{t}e-t)\;\;0\leq \sqrt{t}\leq e. \label{DistSRed}
\end{eqnarray}
Comparing $G_{v}(t)$ and $F_{v}(t)$ for all possible values of $t$ gives
\begin{itemize}
\item $\sqrt{t}<  0$: $G_v(t)=F_{v}(t)=0$.
\item $0\leq \sqrt{t}< b-a$: then $F_{v}(t)-G_v(t) =\frac{1}{e^2}(2\sqrt{t}(a-b)+t)$. Since $\sqrt{t}< b-a\Rightarrow  t < \sqrt{t}(b-a)\Rightarrow t+\sqrt{t}(a-b)< 0 \Rightarrow t+2\sqrt{t}(a-b)< 0 \Rightarrow F_{v}(t)-G_v(t)< 0 \Rightarrow F_{v}(t)< G_v(t)$.
\item $b-a \leq \sqrt{t} < a-b+e$: then $F_{v}(t)-G_v(t) =-\frac{(a-b)^2}{e^2}<0\Rightarrow F_{v}(t)-G_v(t)< 0 \Rightarrow F_{v}(t)< G_v(t)$.
\item  $a-b+e\leq \sqrt{t}< e$: then $F_{v}(t)-G_v(t)=1-\frac{1}{2e^2}(\sqrt{t}+a-b-e)^2+\frac{1}{e^2}(t-2\sqrt{t}e).$ 

Note that $\frac{e (F_{v}(t)-G_v(t))}{e t}=\frac{1}{2e^2}(1-\frac{a-b+e}{\sqrt{t}})>0$, and therefore, $F_{v}(t)-G_v(t)$ is a monotonically increasing function. Due to the monotonicity of $F_{v}(t)-G_v(t)$, the maximum occurs at $e$, so $\max_{t} F_{v}(t)-G_v(t)=F_{v}(e)-G_v(e)=-\frac{(a-b)^2}{e^2}<0$. Therefore, $F_{v}(t)-G_v(t)\leq F_{v}(e)-G_v(e)<0\Rightarrow F_{v}(t)\leq G_v(t)$.
\item $e \leq \sqrt{t}<  b-a+e$: in this case $G_v(t)$ is always 1, hence, $F_{v}(t)\leq G_v(t)$.
\item $b-a+e\leq \sqrt{t}$: in this case $G_v(t)=F_{v}(t)=1$
\end{itemize}

Therefore, we can conclude that \[p(v\leq t)\leq p(v\leq t\mid a=b) \;\forall t \in \mathbb{R} \Rightarrow p(-cv\geq t')\leq p(-cv \geq t'\mid a=b) \;\forall t' \in \mathbb{R}\text{ and } c>0, \]
which implies that random variable $(-cv)$ is stochastically less than random variable $(-cv\mid a=b)$, i.e., $-cv\preceq_{st} -cv\mid a=b$. Consequently, the expectation of any non-decreasing function of these two variables are ordered, i.e., $\mathbb{E}_v(\exp(-cv))\leq \mathbb{E}_{v}(\exp(-cv)\mid a=b)$ for any $c>0$.
\end{proof}

To proceed to the proof of Theorem~\ref{TheoremParandRec}, we use the following characterization for the cutting hyperplanes and subdomains. Assuming that the cutting hyperplanes are equidistant with distant $W=L\slash S$ from each other, we can characterize the $\ell^{\text{th}}\in [S-1]$ cutting hyperplane on $\Omega$ with respect to $k^{\text{th}}$ primary axis of $\mathbb{R}^p$ using the vector of angles $\boldsymbol{\theta}=\{\theta_1,\ldots,\theta_p\}\backslash\{\theta_k\}$, 
\begin{eqnarray}
H_{\boldsymbol{\theta},k,W,\ell}=\{\mathbf x\in \Omega \mid x_k-\sum_{ j\in [p]\backslash\{k\}}\tan(\theta_j)x_j-\ell W=0\}& \forall \ell \in  [S-1].\label{MidHP}
\end{eqnarray}
Note that this cutting hyperplane is orthogonal to the axis $k$ only if $\boldsymbol{\theta} = \mathbf 0$, that is $\theta_j=0$ for $ j\in [p]\backslash\{k\}$.

Denoting, respectively, the hyperplanes containing the ``bottom'' and the ``top'' faces of $\Omega$ as 
\[H_{\boldsymbol{\theta},k,W,0}=\{\mathbf x\in \Omega \mid x_k=0\}   \quad \text{and} \quad H_{\boldsymbol{\theta},k,W,S}=\{\mathbf x\in \Omega \mid x_k-L=0\},\]
we define the $s^{\text{th}}$ subdomain as the intersection of area between two consecutive hyperplanes and $\Omega$, specifically, 
\begin{eqnarray}
\Omega_{\boldsymbol{\theta},k,W,s}=\{\mathbf x \in \Omega \mid \min_{\mathbf x'\in H_{\boldsymbol{\theta},k,W,s-1}}||\mathbf x-\mathbf x'||_2 \leq W \quad \text{and} \quad\min_{\mathbf x'\in H_{\boldsymbol{\theta},k,W,s}}||\mathbf x-\mathbf x'||_2\leq W \},\label{quadruple}
\end{eqnarray} 
where $\|\cdot\|_2$ denotes the Euclidean norm.

\begin{proof}[\textbf{Proof of Theorem~\ref{TheoremParandRec}}]
Let $\mathbf{x}_{\{k\}}=\{x_1,\ldots,x_p\}\backslash\{x_k\}$ for any $\mathbf{x}\in \Omega$. Then, based on how each $\Omega_{\boldsymbol{\theta},k,W,s}$ in~\eqref{quadruple} is constructed and considering the distribution of the data points in $\Omega$ according to~\eqref{USHDist}, all variables $x_j \in \mathbf{x}_{\{i\}} $ are independent and have the uniform distribution $\mathcal{U}(0,L)$. Moreover, by the definition of the hyperplanes in~\eqref{MidHP}, and given $\mathbf{x}_{\{k\}},$ the corresponding values of the variable $x_k$ on the hyperplanes $H_{\boldsymbol{\theta},k,W,s-1}$ and $H_{\boldsymbol{\theta},k,W,s}$ are
\begin{eqnarray}
\sum_{j\in [p]\backslash \{k\}} \tan(\theta_j)x_j+(s-1)w \quad\& \sum_{j\in [p]\backslash \{k\}} \tan(\theta_j)x_j+sw.\label{CondBound}
\end{eqnarray}
Therefore, the conditional distribution $x_k|\mathbf{x}_{\{k\}}$ in the parallelogram subdomain $\Omega_{\boldsymbol{\theta},k,W,s}$ has a uniform distribution whose support is bounded by the values calculated in~\eqref{CondBound}. 
Consequently, given a parallelogram subdomain $\Omega_{\boldsymbol{\theta},k,W,s}$, for any $\mathbf{x} \in \Omega_{\boldsymbol{\theta},k,W,s-1}$,
\begin{subequations}\label{SubdomainDist}
\begin{align}
& x_j \sim \mathcal{U}(0,L) \quad  \forall j\in [p]\backslash \{k\},\\
&x_i|\mathbf{x}^i \sim \mathcal{U}\bigg(\sum_{j\in [p]\backslash \{k\}} \tan(\theta_j)x_j+(s-1)w,\sum_{j\in [p]\backslash \{k\}} \tan(\theta_j)x_j+sw\bigg).
\end{align}
\end{subequations}
Now that we have the distribution~\eqref{SubdomainDist}, we expand $\mathbb{E}_{\Omega_{\boldsymbol{\theta},k,W,s}}\big(\mathcal{\phi}(\mathbf{x},\mathbf{x}')\big)$ by conditioning, that is
\begin{subequations}
\begin{align}
&\mathbb{E}_{\Omega_{\boldsymbol{\theta},k,W,s}}\big(\mathcal{\phi}(\mathbf{x},\mathbf{x}')\big)=\mathbb{E}_{\mathbf{x}_{\{k\}},\mathbf{x}'_{\{k\}}}\bigg(\mathbb{E}_{x_k,x'_k}\big(\mathcal{\phi}(\mathbf{x},\mathbf{x}')\mid \mathbf{x}_{\{k\}},\mathbf{x}'_{\{k\}}\big)\bigg)\label{ConditioningEQ}\\
&=\mathbb{E}_{\mathbf{x}_{\{k\}},\mathbf{x}'_{\{k\}}}\bigg(\exp\bigg(-\sum_{j\in [p]\backslash\{k\}}\gamma_j(x_j-x'_j)^2\bigg) \mathbb{E}_{x_k,x'_k}\bigg(\exp\big(-\gamma_k(x_k-x'_k)^2\big)\mid \mathbf{x}_{\{k\}},\mathbf{x}'_{\{k\}}\bigg)\bigg)\\
&=\mathbb{E}_{\mathbf{x}_{\{k\}},\mathbf{x}'_{\{k\}}}\big(g(\mathbf{x}_{\{k\}},\mathbf{x}'_{\{k\}})h(\mathbf{x}_{\{k\}},\mathbf{x}'_{\{k\}})\big).
\end{align}
\end{subequations}

Note that the function $g(\mathbf{x}_{\{k\}},\mathbf{x}'_{\{k\}})$ is always positive and independent of $\boldsymbol{\theta}$, and  function $h(\mathbf{x}_{\{k\}},\mathbf{x}'_{\{k\}})$ is positive that attains its maximum for any given $\mathbf{x}_{\{k\}},\mathbf{x}'_{\{k\}}$ at $\boldsymbol{\theta}=\mathbf{0}$ by Lemma~\eqref{OrderingLemma}. Therefore, $\boldsymbol{\theta}=\mathbf{0}$, \[g(\mathbf{x}_{\{k\}},\mathbf{x}'_{\{k\}})h(\mathbf{x}_{\{k\}},\mathbf{x}'_{\{k\}})\leq g(\mathbf{x}_{\{k\}},\mathbf{x}'_{\{k\}})h(\mathbf{x}_{\{k\}},\mathbf{x}'_{\{k\}}\mid \boldsymbol{\theta}=\mathbf{0})\quad \forall \mathbf{x}_{\{k\}},\mathbf{x}'_{\{k\}},\]
which results in \[\mathbb{E}_{\Omega_{\boldsymbol{\theta},k,W,s}}\big(\mathcal{\phi}(\mathbf{x},\mathbf{x}')\big)\leq\mathbb{E}_{\Omega_{\boldsymbol{\theta},k,W,s}}\big(\mathcal{\phi}(\mathbf{x},\mathbf{x}')\mid \boldsymbol{\theta}=\mathbf{0}\big)\] \[\Rightarrow\argmax_{\boldsymbol{\theta}} \mathbb{E}_{\Omega_{\boldsymbol{\theta},k,W,s}}\big(\mathcal{\phi}(\mathbf{x},\mathbf{x}')\big)=\mathbf{0}.\]
\end{proof}

 \subsection{Proof of Theorem~\ref{TheoremCompMainDirs}}
First, we prove the following lemma 
 \begin{lemma}
$\mathbb{E}_{z_1,z_2}\bigg(\exp\big(-c(z_1-z_2)^2\big) \bigg)=\int_0^{b^2}\exp(-c t)(\frac{1}{b\sqrt{t}}-\frac{1}{b^2})dt $, where $z_1,z_2\stackrel{i.i.d}{\sim}\mathcal{U}(a,a+b)$.\label{DistLemma}
\end{lemma}

\begin{proof}
Let $v=(z_1-z_2)^2$, then~\cite{philip2007probability} shows that $v$ has the following PDF: \[f_v(t)=\frac{1}{\sqrt{t}b}-\frac{1}{b^2}\quad \forall\; 0\leq t\leq b^2; \] therefore, 
\[\mathbb{E}_{z_1,z_2}\bigg(\exp\big(-c(z_1-z_2)^2\big)\bigg)=\mathbb{E}_v\bigg(\exp(-c v)\bigg)=\]\[
\int_0^{b^2}\exp(-c t)f_s(t)dt=\int_0^{b^2}\exp(-c t)(\frac{1}{b\sqrt{t}}-\frac{1}{b^2})dt.\]\end{proof}

\begin{proof}[\textbf{Proof of Theorem~\ref{TheoremCompMainDirs}}]
By the assumptions of uniform distribution of points in $\Omega$~\eqref{USHDist}, and independence of the dimensions due to geometry of $\Omega_{\mathbf{0},k,W,s}$, for any $\mathbf x \in \Omega_{\mathbf{0},i,W,s}$, 
\begin{eqnarray}\label{DistStarightSD}
x_k \sim\mathcal{U}\big((s-1)W,sW\big)\quad \& \quad x_j \sim \mathcal{U}\big(0,L\big) \quad \forall j\in[p]\backslash\{k\}.
\end{eqnarray}
Letting $G_k=\mathbb{E}_{\Omega_{\mathbf{0},k,W,s}}(\mathcal{\phi}(\mathbf{x},\mathbf{x}'))$, and using distribution~\eqref{DistStarightSD}, 
\begin{subequations}
\begin{align}
&G_k=\mathbb{E}_{x_k}\bigg(\exp\big(-\gamma_k(x_k-x'_k)^2\big)\bigg)\prod_{j\in [p]\backslash\{k\}}\mathbb{E}_{x_j}\bigg(\exp\big(-\gamma_j(x_j-x'_j)^2\big)\bigg)\label{SumToProdEq}\\
&=\mathbb{E}_{v_k}\bigg(\exp(-\gamma_k v_k)\bigg)\prod_{j\in [p]\backslash\{k\}}\mathbb{E}_{v_j}\bigg(\exp(-\gamma_j v_j)\bigg)\label{CngVarEq1}\\
&=\bigg(\int_0^{W^2}\exp(-\gamma_k t)(\frac{1}{W\sqrt{t}}-\frac{1}{W^2})dt\bigg)\Bigg(\prod_{j\in [p]\backslash\{k\}}\bigg(\int_0^{L^2}\exp(-\gamma_j t)(\frac{1}{L\sqrt{t}}-\frac{1}{L^2})dt\bigg)\Bigg)\label{CngVarEq2}\\
&=\Bigg(\int_{0}^{W^2}g^W_{k}(t)dt\Bigg)\Bigg(\prod_{j\in [p]\backslash\{k\}}\bigg(\int_{0}^{L^2}g^L_{j}(t)dt\bigg)\Bigg),\label{CngVarEq3}
\end{align}
\end{subequations}
where equality~\eqref{SumToProdEq} follows from the independence of dimensions in each $\Omega_{\mathbf{0},i,W,s}$,  equalities~\eqref{CngVarEq1} and~\eqref{CngVarEq2} follow from Lemma~\eqref{DistLemma} with $f_{v_k}(t)=\frac{1}{\sqrt{t}W}-\frac{1}{W^2}\;\; 0\leq t\leq W^2$ and $ f_{v_j}(t)=\frac{1}{\sqrt{t}L}-\frac{1}{L^2}\; \;0\leq t\leq L^2$, and $g^m_{\ell}(t)= \exp(-\gamma_{\ell} t)(\frac{1}{m\sqrt{t}}-\frac{1}{m^2})$ in~\eqref{CngVarEq3}.

To show that $G_p-G_k\geq 0$ for any $k\in [p]$, We first expand $G_p-G_k$,
{\fontsize{0.32cm}{.5cm}\selectfont
\[G_p-G_k=\Bigg(\int_{0}^{W^2}g^W_{p}(t)dt\Bigg)\Bigg(\prod_{j\in [p]\backslash\{p\}}\bigg(\int_{0}^{L^2}g^L_{j}(t)dt\bigg)\Bigg)-\Bigg(\int_{0}^{W^2}g^W_{k}(t)dt\Bigg)\Bigg(\prod_{j\in [p]\backslash\{k\}}\bigg(\int_{0}^{L^2}g^L_{j}(t)dt\bigg)\Bigg)\]
\[=\Bigg(\prod_{j\in [p]\backslash\{k,p\}}\bigg(\int_{0}^{L^2}g^L_{j}(t)dt\bigg)\Bigg)\Bigg(\int_{0}^{W^2}g^W_{p}(t)dt\int_{0}^{L^2}g^L_{k}(t)dt-\int_{0}^{W^2}g^W_{k}(t)dt\int_{0}^{L^2}g^L_{p}(t)dt\Bigg)=A*B.\]}
Note that $A$ is always positive, since each $\int_{0}^{L^2}g^L_{j}(t)dt$ is the expectation of the random variable $\exp(-\gamma_jv_j)$ which is positive. Hence, it is enough to show that $B$ is positive. Expanding B further,
\begin{subequations}
{\fontsize{0.32cm}{.5cm}\selectfont
\begin{align}
&B=\bigg(\int_{0}^{W^2}g^W_{p}(t)dt\bigg)\bigg(\int_{0}^{W^2}g^L_{k}(t)dt+\int_{W^2}^{L^2}g^L_{k}(t)dt\bigg)-\bigg(\int_{0}^{W^2}g^W_{k}(t)dt\bigg)\bigg(\int_{0}^{W^2}g^L_{p}(t)dt+\int_{W^2}^{L^2}g^L_{p}(t)dt\bigg)\\
&=\int_{t_k:0}^{W^2}\int_{t_p:0}^{W^2}g^W_{p}(t_k)g^L_{k}(t_p)dt_kdt_p+\int_{t_k:0}^{W^2}\int_{t_p:W^2}^{L^2}g^W_{p}(t_k)g^L_{k}(t_p)dt_kdt_p\nonumber\\
&-\int_{t_k:0}^{w^2}\int_{t_p:0}^{w^2}g^w_{k}(t_k)g^L_{p}(t_p)dt_kdt_p-\int_{t_k:0}^{w^2}\int_{t_p:w^2}^{L^2}g^w_{k}(t_k)g^L_{p}(t_p)dt_kdt_p\\
&=\int_{t_k:0}^{W^2}\int_{t_p:0}^{W^2}\big(\exp(-\gamma_pt_k-\gamma_kt_p)-\exp(-\gamma_kt_k-\gamma_pt_p)\big)\big(\frac{1}{W\sqrt{t_k}}-\frac{1}{W^2}\big)\big(\frac{1}{L\sqrt{t_p}}-\frac{1}{L^2}\big)dt_kdt_p\nonumber\\
&+\int_{t_k:0}^{W^2}\int_{t_p:W^2}^{L^2}\big(\exp(-\gamma_pt_k-\gamma_kt_p)-\exp(-\gamma_kt_k-\gamma_pt_p)\big)\big(\frac{1}{W\sqrt{t_k}}-\frac{1}{W^2}\big)\big(\frac{1}{L\sqrt{t_p}}-\frac{1}{L^2}\big)dt_kdt_p\\
&=\int_{t_k:0}^{W^2}\int_{t_p:0}^{W^2}c(t_k,t_p)dt_kdt_p+\int_{t_k:0}^{W^2}\int_{t_p:W^2}^{L^2}c(t_k,t_p)dt_kdt_p.\label{SecondExpansion}
\end{align}
}
\end{subequations}
Note that for any member of set 
\begin{eqnarray}\label{A1}
\{(W,L,t_p,t_k,\gamma_p,\gamma_k) \mid 0<W<L,\; 0<\gamma_k<\gamma_p,\;0\leq t_k\leq W^2,\; W^2\leq t_p\leq L^2\},
\end{eqnarray}
we have
\begin{eqnarray}\label{A2}
\big(\frac{1}{w\sqrt{t_k}}-\frac{1}{w^2}\big)\big(\frac{1}{L\sqrt{t_p}}-\frac{1}{L^2}\big)>0,
\end{eqnarray} 
and also 
\begin{eqnarray}
(-\gamma_pt_k-\gamma_kt_p)-(-\gamma_kt_k-\gamma_pt_p)=(\gamma_p-\gamma_k)(t_p-t_k)>0\label{ExponentInEq1},
\end{eqnarray}
where the latter results in
\begin{eqnarray}
\exp(-\gamma_pt_k-\gamma_kt_p)-\exp(-\gamma_kt_k-\gamma_pt_p)>0\label{ExponentInEq2}.
\end{eqnarray}

Therefore, by~\eqref{A2} and~\eqref{ExponentInEq2}, the integrand $c(t_k,t_p)$ in~\eqref{SecondExpansion} is positive for any member of set~\eqref{A1}, so is integral $\int_{t_k:0}^{W^2}\int_{t_p:W^2}^{L^2}c(t_k,t_p)dt_kdt_p$. Hence, to complete the proof we need to show integral $\int_{t_k:0}^{w^2}\int_{t_p:0}^{w^2}c(t_k,t_p)dt_kdt_p$ in~\eqref{SecondExpansion} is also positive. To show this, we expand the integral,
\begin{subequations}
{\fontsize{0.32cm}{.5cm}\selectfont
\begin{align}
&\int_{t_k:0}^{W^2}\int_{t_p:0}^{W^2}c(t_k,t_p)dt_kdt_p=\int_{t_k:0}^{W^2}\int_{t_p:t_k}^{W^2}c(t_k,t_p)dt_kdt_p+\int_{t_p:0}^{W^2}\int_{t_k:t_p}^{W^2}c(t_k,t_p)dt_pdt_k\\
&=\int_{t_k:0}^{W^2}\int_{t_p:t_k}^{W^2}c(t_k,t_p)dt_kdt_p+\int_{t_k:0}^{W^2}\int_{t_p:t_k}^{W^2}c(t_p,t_k)dt_kdt_p=\int_{t_k:0}^{W^2}\int_{t_p:t_k}^{W^2}\big(c(t_k,t_p)+c(t_p,t_k)\big)dt_kdt_p\\
&=\frac{1}{wL}\int_{t_k:0}^{W^2}\int_{t_p:t_k}^{W^2}\big(\exp(-\gamma_pt_k-\gamma_kt_p)-\exp(-\gamma_kt_k-\gamma_pt_p)\big)(\frac{1}{\sqrt{t_k}}-\frac{1}{\sqrt{t_p}})(\frac{1}{W}-\frac{1}{L})dt_kdt_p\label{LastExpansion}.
\end{align}
}%
\end{subequations}
Similar to~\eqref{A1}-\eqref{ExponentInEq2}, for any member of set 
\begin{eqnarray}\label{A3}
\{(W,L,t_p,t_k,\gamma_p,\gamma_k) \mid 0<W<L,\; 0<\gamma_k<\gamma_p,\;0 \leq t_k\leq W^2,\;t_k\leq t_p\leq W^2\},
\end{eqnarray}
we have  $(\frac{1}{\sqrt{t_k}}-\frac{1}{\sqrt{t_p}})>0$, $(\frac{1}{W}-\frac{1}{L})>0$, and $\big(\exp(-\gamma_pt_k-\gamma_kt_p)-\exp(-\gamma_kt_k-\gamma_pt_p)\big)>0$. Hence the integrand in~\eqref{LastExpansion} is positive for any member of set~\eqref{A3}, so is integral~\eqref{LastExpansion}, and the proof is complete.

\end{proof}

\section{A simulation study on the relation between expected error~\eqref{ErrorTerm} and $\mathbb{E}_{\Omega_s}(\phi^2(\mathbf{x},\mathbf{x}'))$}\label{SimulAPX}

Consider the squared exponential Gaussian kernel $\phi(x,x')=exp(-\gamma (x-x')^2)$ with $\gamma>0$ defined on 
\begin{eqnarray}
\Omega_s=\{x\in \mathbb{R}|a \leq x\leq a+b\}\label{SimDomain}
\end{eqnarray}
 with uniform sampling distribution
 \begin{eqnarray}
 x\sim\mathcal{U}(a,a+b)\quad \forall x\in \Omega_s.\label{SimDist}
 \end{eqnarray}
To have a general simulation study, we need the following lemma.

\begin{lemma}
$\mathbb{E}_{z_1,z_2}\bigg(\exp\big(-c(z_1-z_2)^2\big)\bigg)$, where $z_1,z_2\stackrel{i.i.d}{\sim}\mathcal{U}(a,a+b)$, is a monotonically decreasing function of $c$ and $b$.\label{MonotoneLemma}
\end{lemma}
\begin{proof}
We need to show that $\nabla g(b,c)=[\frac{\partial g(b,c)}{\partial {b}},\frac{\partial g(b,c)}{\partial c}]^T<0$ for all $[b,c]^T>0$, where \[g(b,c)=\mathbb{E}_{z_1,z_2}\bigg(\exp\big(-c(z_1-z_2)^2\big)\bigg)=\int_0^{b^2}\exp(-c t)(\frac{1}{b\sqrt{t}}-\frac{1}{b^2})dt\]
by Lemma~\ref{DistLemma}.

We can write $\frac{\partial g(b,c)}{\partial b}$ as
\begin{subequations}
\begin{align}
&\frac{\partial g(b,c)}{\partial b}=
\frac{1}{b^2}\int_0^{b^2}\exp(-c t)(\frac{2}{b}-\frac{1}{\sqrt{t}})dt\label{LeibnizDiff}\\
&=\frac{1}{b^2}\bigg(\bigg[\exp(-c t)(\frac{2t}{b}-2\sqrt{t})\bigg]_0^{b^2}-\int_0^{b^2}-c\exp(-c t)(\frac{2t}{b}-2\sqrt{t})\bigg)\label{IntegByPart}\\
&=\frac{2c}{b^2}\int_0^{b^2}\exp(-c t)(\frac{t}{b}-\sqrt{t}),
\end{align}
\end{subequations}
where equalities~\eqref{LeibnizDiff} and~\eqref{IntegByPart} follow from the Leibniz integral differentiation and the integration by part rules, respectively. It is easy to check that integrand $\exp(-c t)(\frac{t}{b}-\sqrt{t})$ is always negative for any member of set $\{(b,c,t)\mid 0<b , 0<c, 0\leq t\leq b^2\}$; therefore, we always have $\frac{\partial g(b,c)}{\partial b}<0$.

Moreover, for $\frac{\partial g(b,c)}{\partial c}$,
\[
\frac{\partial g(b,c)}{\partial c}=
\int_0^{b^2}-t\exp(-ct)(\frac{1}{b\sqrt{t}}-\frac{1}{b^2})dt=\frac{-1}{b}\int_0^{b^2}t\exp(-c t)(\frac{1}{\sqrt{t}}-\frac{1}{b})dt.\]
It is again easy to check that the integrand $t\exp(-c t)(\frac{1}{\sqrt{t}}-\frac{1}{b})$ is positive for any member of set $\{(b,c,t)\mid 0<b , 0<c, 0\leq t\leq b^2\}$. Therefore, $\frac{\partial g(b,c)}{\partial c}$ is always negative. 
\end{proof}

\begin{figure}[H]

\begin{center}
	\begin{subfigure}{0.45\textwidth}

\includegraphics[height=6cm,width=7.5cm]{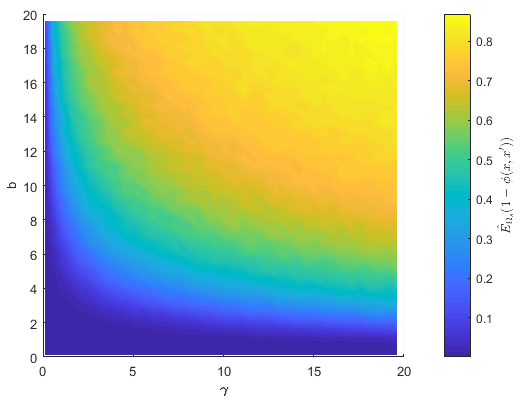}
	\end{subfigure}
\end{center}
\caption{Heat map of the approximation of expected error function~\eqref{ErrorTerm} on domain~\eqref{SimDomain} and sampling distribution~\eqref{SimDist} for varying values of $\gamma$ and $b$ and a fixed value of $m_s$}
\label{fig:Error}
\end{figure}

By Lemma~\ref{MonotoneLemma}, expectation function 
\begin{eqnarray}\label{OneDimError}
\mathbb{E}_{\Omega_s}(\phi^2(\mathbf{x},\mathbf{x}'))=\mathbb{E}_{x,x'}(\exp(-2\gamma(x-x')^2)
\end{eqnarray} is a monotonically decreasing function of $\gamma$ and $b$. This means that there are only two ways to increase expectation $\mathbb{E}_{\Omega_s}(\phi^2(\mathbf{x},\mathbf{x}'))$, which are either decreasing $\gamma$ or decreasing $b$. The approximation of expected error function~\eqref{ErrorTerm} on domain~\eqref{SimDomain} and sampling distribution~\eqref{SimDist} for varying values of $\gamma$ and $b$ and a fixed value of $m_s$ using a heat map plot is shown in Figure~\ref{fig:Error}. We observe that as the values of $\gamma$ or $b$ decrease, or equivalently, $\mathbb{E}_{\Omega_s}(\phi^2(\mathbf{x},\mathbf{x}'))$ increases, the approximation of the expected error function decreases.

Our simulation study can be used to infer a more general case. Consider the covariance function as  $\phi(\mathbf x,\mathbf x')=\exp(-\sum_{i=1}^{p} \gamma_k(x_k-x'_k))$ defined on $\Omega_s$ as a $p$-dimensional hyper-rectangle with side lengths $b_1, \ldots, b_p$ with a uniform sampling distribution, i.e., $x_k \sim \mathcal{U}(a_k,a_k+b_k)\quad \forall\mathbf{x}\in \Omega_s$. With this setup, we can write 
\begin{eqnarray}
\mathbb{E}_{\Omega_s}(\phi^2(\mathbf{x},\mathbf{x}'))=\prod_{i=1}^{p} \mathbb{E}_{x_k,x_k'}(\exp(-2\gamma_k(x_k-x'_k)),
\end{eqnarray}
which is a monotonic function in each $b_i$ and $\gamma_i$ by lemma~\ref{MonotoneLemma}. Therefore, our simulation results are valid for this generalized case as well.

\begin{figure}[H]
\begin{center}
	\begin{subfigure}{0.3\textwidth}
     \includegraphics[width=\textwidth]       {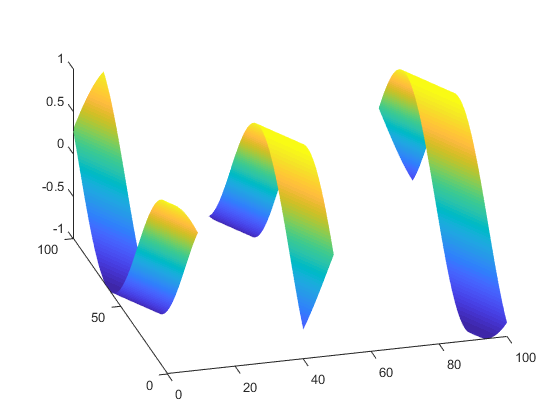}
	   \caption {$\mathbf{a}=[1,0]^t$}
     \label{fig:DirCutSyn0}
	\end{subfigure}
    \begin{subfigure}{0.3\textwidth}
     \includegraphics[width=\textwidth]       {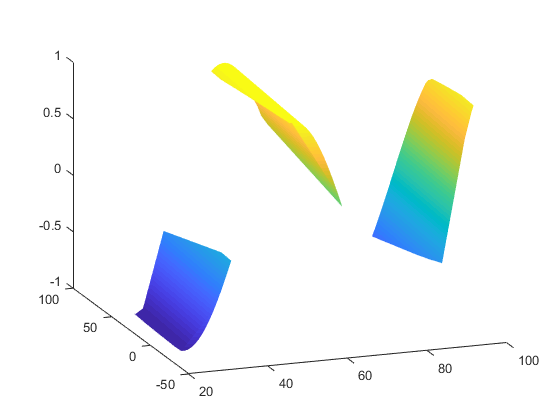}
	   \caption {$\mathbf{a}=[0.43,0.9]^t$}
     \label{fig:DirCutSyn30}
	\end{subfigure}
    \begin{subfigure}{0.3\textwidth}
     \includegraphics[width=\textwidth]       {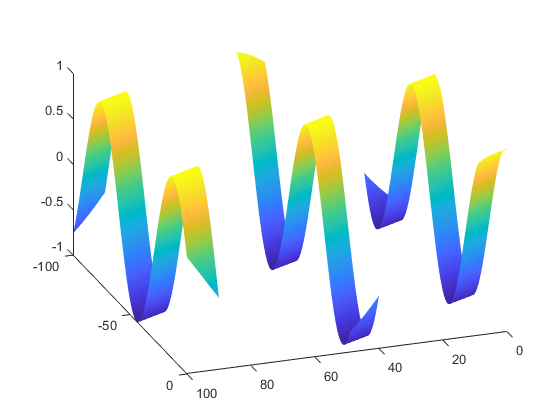}
	   \caption {$\mathbf{a}=[0,1]^t$}
     \label{fig:DirCutSyn90}
	\end{subfigure}
	
\end{center}
\caption{Local functions created by cutting orthogonal to directions $[1,0]$, $[0.43,0.9]$ (solution of~\eqref{PrimaryLik}), and $[0,1]$ on a synthetic dataset}
\label{fig:DirCut}
\end{figure}

Finally, we present some intuition behind the theoretical results in Section~\ref{sec_SPLK}. The reason why the direction $\mathbf a$, found by solving optimization problem~\eqref{PreFinalLik}, results in a better covariance approximation in each subdomain can be visually perceived for a two-dimensional domain. Suppose we can partition the domain of two-dimensional function $f(\mathbf{x})=\cos(0.05x_1+0.1x_2)$ by cutting orthogonal to either of three directions $[1,0]$, $[0.43,0.9]$, or $[0,1]$, where direction $[0.43,0.9]$ is the direction of the fastest covariance decay obtained by optimizing~\eqref{PreFinalLik}. Figure~\ref{fig:DirCut} shows the 3-D presentations of three local functions created by cutting orthogonal to each direction. \comm{We observe that the local functions created by cutting orthogonal to the desired direction have a less fluctuating  behaviour compared to those of directions $[1,0]$ and $[0,1]$.} That the function has less fluctuation allows a random point on the local functions of Figure~\ref{fig:DirCutSyn30} to have (on average) higher correlation to its neighboring data points. Therefore, we can obtain a better approximation of local covariance structures by using the same number of pseudo data points located in each subdomain.

\section{Solving optimization problem~\eqref{PreFinalLik}}\label{PGDSol}
Let first write the partial derivatives of objective function in~\eqref{PreFinalLik}, 
\begin{eqnarray}
\frac{\partial\mathcal{L}(\bar{\mathbf{a}})}{\partial a_k}=-\mathbf{y}^T_n(\mathbf{K}_n^{\bar{\mathbf{a}}}+\sigma^2\mathbf{I_n})^{-1} \frac{\partial\mathbf{K}_n^{\bar{\mathbf{a}}}}{\partial a_k}(\mathbf{K}_n^{\bar{\mathbf{a}}}+\sigma^2\mathbf{I_n})^{-1}\mathbf{y}_n+\text{tr}((\mathbf{K}_n^{\bar{\mathbf{a}}}+\sigma^2\mathbf{I_n})^{-1} \frac{\partial\mathbf{K}_n^{\bar{\mathbf{a}}}}{\partial a_k}),
\end{eqnarray}
where  $\frac{\partial\mathbf{K}_n^{\bar{\mathbf{a}}}}{\partial a_k}
$ is the matrix of element-wise derivatives with respect to the $k^{\text{th}}$ element of $\bar{\mathbf{a}}$. Note that each element of $\frac{\partial\mathbf{K}_n^{\bar{\mathbf{a}}}}{\partial a_k}
$ involves the term  $\frac{1}{\sqrt{1-\bar{\mathbf{a}}^T\bar{\mathbf{a}}}}$. Therefore, the gradient of the objective function in~\eqref{PreFinalLik} does not exist on the boundary of the feasible region, i.e., $\nabla \mathcal{L}(\bar{\mathbf{a}})\rightarrow \infty$ as $\bar{\mathbf{a}}^T\bar{\mathbf{a}}\rightarrow1$. Therefore, to avoid an undefined gradient on the boundary, we modify the optimization by making the feasible region slightly tighter, i.e., 

\begin{eqnarray}
\begin{aligned}
&\min_{\bar{\mathbf{a}}}
&&\mathcal{L}(\bar{\mathbf{a}})= \mathbf{y}^T_n(\mathbf{K}_n^{\bar{\mathbf{a}}}+\sigma^2\mathbf{I_n})^{-1}\mathbf{y}_n+\text{log}|\mathbf{K}_n^{\bar{\mathbf{a}}}+\sigma^2\mathbf{I}_n|\\
&\text{subject to}
&&\bar{\mathbf{a}}^T\bar{\mathbf{a}}\leq 1-\epsilon,
\end{aligned}\label{FinalLik}
\end{eqnarray}
where $\epsilon$ is a very small number. In our experiments, we set  $\epsilon=0.001$.

Due to the simple convex structure of constraint $\bar{\mathbf{a}}^T\bar{\mathbf{a}}\leq 1-\epsilon$, i.e., a $d-1$-dimensional hypersphere, optimization~\eqref{FinalLik} can be solved by the Projected Gradient Descent algorithm~\citep{nesterov1994interior}. In this projection algorithm, the $(j+1)^{\text{th}}$ decent step is defined by 
\begin{eqnarray}
\bar{\mathbf{a}}^{j+1}=\mathcal{P}\big(\bar{\mathbf{a}}^j-\frac{\alpha}{||\nabla \mathcal{L}(\bar{\mathbf{a}}^{j})||}\nabla \mathcal{L}(\bar{\mathbf{a}}^{j})\big),
\end{eqnarray}
where $\frac{\alpha}{||\nabla \mathcal{L}(\bar{\mathbf{a}}^{j})||}$ is a normalized length step, and
\begin{eqnarray}
\begin{aligned}
&\mathcal{P}(\mathbf{z})= &&\text{argmin}_{\mathbf{w}} ||\mathbf{w}-\mathbf{z}||\\
&\text{subject to}
&&\mathbf{w}^T\mathbf{w}\leq 1-\epsilon.
\end{aligned}
\end{eqnarray}

$\mathcal{P}(\mathbf{z})=\mathbf{z}$ when $\mathbf{z}^T\mathbf{z}\leq 1-\epsilon$, otherwise the solution to $\mathcal{P}(\mathbf{z})$  occurs at the point that the line defined by $\mathbf{z}$ and the center of the hypersphere, ($\mathbf{0}$), crosses the boundary of the hypersphere, i.e, intersection of $\frac{w_1}{z_1}=\frac{w_2}{z_2}=\ldots=\frac{w_{p-1}}{z_{p-1}}$ and $\mathbf{w}^T\mathbf{w}=1-\epsilon$. Therefore, the solution to $\mathcal{P}(\mathbf{z})$ has the closed form,
\begin{eqnarray}
\mathcal{P}(\mathbf{z})=
\begin{cases}
\mathbf{z}&\mathbf{z}^T\mathbf{z}\leq 1-\epsilon\\
[\frac{z_1}{\sqrt{\mathbf{z}^T\mathbf{z}}},\ldots,\frac{z_{p-1}}{\sqrt{\mathbf{z}^T\mathbf{z}}}]^T &\mathbf{z}^T\mathbf{z}> 1-\epsilon.
\end{cases}
\end{eqnarray}

\section{Practical Considerations}\label{sec_practical}
\subsection{Creating boundaries, control points, and boundary functions}\label{sub_sec_numb_pd}

The focus of this section is on the practical implementation of SPLK, and therefore, the characterization of cutting hyperplanes differs from the discussion in Section ~\ref{DirofCuts}. Here, instead of using a vector of angles corresponding to primary axes of input space, we use a given direction, which can be the solution to optimization~\eqref{PreFinalLik} or any other arbitrary direction, to define the cutting hyperplanes. 

Recall that in our partitioning policy all the cutting hyperplanes are parallel to each other, and therefore, orthogonal to a unique direction, which is characterized by a vector $\mathbf{a}=[a_1,\ldots,a_p]^T$. Let $\mathbf{Z}=\{\mathbf{x}_i^T\mathbf{a}\mid \mathbf{x}_i \in \mathbf{X}\}$ denote the projection of all the input vectors onto $\mathbf{a}$. Next, consider the ordered set $\{z_1,\ldots, z_{S-1}\}$, where $\min \mathbf{Z}<z_1$ and $z_{S-1}<\max \mathbf{Z}$, and $z_{\ell}<z_{\ell+1}$, for $\ell\in[S-1]$. 

Given the set $\{z_1,\ldots, z_{S-1}\}$ and direction $\mathbf{a}$, which is in fact the normal vector of all of the cutting hyperplanes, we define 
the $\ell^{\text{th}}$ cutting hyperplane orthogonal to $\mathbf{a}$ as $H_{\ell,\mathbf{a}}=\{\mathbf x\in \Omega \mid\ a_1x_1+\ldots+a_px_p=z_{\ell}\}$ for $\ell\in [S-1]$. We use the data points close to $H_{\ell,\mathbf{a}}$ to locate the control points. To this end, we first define $\boldsymbol{\Delta}_{\ell}=\{\mathbf{x}_i\in \mathbf{X}|\;|\mathbf{x}_i^T\mathbf{a}-z_{\ell}|<\delta \}$ as the set of training data points whose Euclidean distance to  $\mathcal{H}_{\ell,\mathbf{a}}$ is less than a predefined constant $\delta$. Then, calculate the maximum and minimum of the $k^\text{th}$ dimension of the data points in $\boldsymbol{\Delta}_{\ell}$, respectively,
\begin{eqnarray}\label{VerCalc}
\tau_{1,k,\ell}=\smash{\displaystyle\max_{\mathbf{x}_i\in \boldsymbol{\Delta}_{\ell}}} \mathbf{x}_i^T\mathbf{e}_k \qquad \text{and}\qquad\tau_{0,k,\ell}=\smash{\displaystyle\min_{\mathbf{x}_i\in \boldsymbol{\Delta}_{\ell}}} \mathbf{x}_i^T\mathbf{e}_k,
\end{eqnarray}
where $\mathbf{e}_k$ is the unit vector along the $k^\text{th}$ primary axis of the space for $k\in[p]$. As such, the set $\mathbf{V}_\ell=\left\{\left[\tau_{b,1,\ell},\ldots,\tau_{b,p,\ell}\right]^T| b=0,1\right\}$ characterizes the vertices of the hyper-rectangle inscribing $\boldsymbol{\Delta}_{\ell}$. Next, we uniformly sample $Q>0$ points from $\mathbf{V}_\ell$ and denote the set of all these points as $\mathbf U_{\ell}$. We obtain the set of control points on $H_{\ell,\mathbf{a}}$ denoted as $\mathbf{C}_{\ell}$ by projecting the points in $\mathbf U_{\ell}$ on $\mathcal{H}_{\ell,\mathbf{a}}$,
\begin{eqnarray}
\mathbf{C}_{\ell}=\{(z_{\ell}-\mathbf{u}^T\mathbf{a})\mathbf{a}+\mathbf{u}\mid\forall \mathbf{u} \in \mathbf{U}_\ell\}.
\end{eqnarray}

There are several ways to choose the width of each subdomain, i.e., $z_{\ell+1}-z_{\ell}$ for $\ell \in [S-1]$. One way is to choose a fixed width for the subdomains; however, this approach results in subdomains with different numbers of local data points depending on their distribution on the domain. Also \emph{adaptive mesh generation} techniques~\citep{becker2001optimal} can be used to vary the widths to balance the error among the subdomains. In Section~\ref{sec_results}, we use varying widths for the subdomains to balance the numbers of local data points across the subdomains. This approach helps us to control the computation time of the algorithm, because it is evenly distributed among the subdomains.

Furthermore, to impose connectivity on the optimization procedure discussed in Section~\ref{sub_sec_mean_pred}, we need to specify the boundary values for each control point $\mathbf{c} \in \mathbf{C}_{\ell}$. To this end, we fit a boundary GPR over the hyper-rectangle defined by $\mathbf{V}_{\ell}$ using the data points in $\boldsymbol{\Delta}_{\ell}$. We then use the predictive mean function of this GPR to determine the boundary values. Letting $\mathcal{R}_{\ell}(.)$ denote as the predictive mean function of the GPR constructed by $\boldsymbol{\Delta}_{\ell}$, the boundary value for each $\mathbf{c} \in \mathbf{C}_{\ell}$ is
\begin{eqnarray}
\mathcal{R}_{\ell}(\mathbf{c})=\mathbf{k}_{\mathbf{c}\boldsymbol{\Delta}_{\ell}}(\mathbf{K}_{\boldsymbol{\Delta}_{\ell}\boldsymbol{\Delta}_{\ell}}+\sigma^{2}_{\ell}\mathbf{I}_{\ell})^{-1}\mathbf{y}_{\boldsymbol{\Delta}_{\ell}},\label{BoundaryFunc}
\end{eqnarray}
where $\mathbf{k}_{\mathbf{c}\boldsymbol{\Delta}_{\ell}}$ is the covariance vector between the control point $\mathbf{c} \in \mathbf{C}_{\ell}$ and the neighboring data points in $\boldsymbol{\Delta}_{\ell}$, and $\mathbf{K}_{\boldsymbol{\Delta}_{\ell}\boldsymbol{\Delta}_{\ell}}$ is the covariance matrix between the neighboring data points in $\boldsymbol{\Delta}_{\ell}$ themselves.
In Section~\ref{sub_sec_mean_pred}, with a slight abuse of notation, we denote $\mathcal{R}(.)$ as a function that takes a control point as an input and returns $\mathcal{R}_{\ell}(.)$, depending on the location of the control point. Note that since the set of neighboring data points $\boldsymbol{\Delta}_{\ell}$
is a small set, we use a full GPR to obtain functions~\ref{BoundaryFunc}.

\subsection{Control points density}\label{sec_seb_cpdensity}

As discussed in Section~\ref{ChoicesOFParams}, we use a density parameter and the dimension of the boundary space, i.e., $q$ and $p-1$, to determine the number of control points to be uniformly located on each boundary. Notably, our experiments show that setting $q$ to small values usually results in satisfactory performance, while increasing it does not significantly affect the prediction accuracy, but increases the computation burden, particularly in higher dimensional domains. The results of testing SPLK on our four datasets with varying values of $q$ and all other parameters fixed are reported in Table~\ref{table:lambda}. An increase in the value of $q$ slightly improves the prediction accuracy in terms of NLPD and MSE. Moreover, as the dimension of the domain of data increases, an increase in the value of $q$ results in much longer computation time.

\begin{table}\footnotesize
 \begin{tabular}{p{3cm} p{3cm} p{3cm} p{3cm} p{3cm}}
Dataset&$q$&Time&MSE&NLPD\\
\hline\hline
&3&145.50&12.18&2.61\\
TCO&4&145.61&12.15&2.61\\
&5&146.06&11.98&2.60\\
\hline
&3&134.48&25.50&2.60\\
Levitus&4&134.47&25.44&2.59\\
&5&135.36&25.25&2.59\\
\hline
&3&157.62&0.42&4.01\\
Dasilva&4&159.98&0.38&3.30\\
&5&167.79&0.38&3.05 \\
\hline
&2.2&147.53&17.41&2.67\\
Protein&2.5&202.09&17.39&2.66\\
&3&3651.33&17.38&2.65\\
\hline
\end{tabular}
\caption{Effect of $q$ on efficiency of SPLK. $S=30$ and $\kappa=4$ across all the datasets}
\label{table:lambda}
\end{table}

\subsection{Hyperparameter learning}\label{sub_sec_numb_hyp}
 Maximizing the marginal likelihood of the training data, $p(\mathbf{y})$, is a popular method for learning the hyperparameters of a model~\citep{Ras}. In SPLK, instead of one global marginal likelihood function, there are $S$ local functions $p(\mathbf{y}_s)$, each of which can be trained independently. Recall that our local predictors are in fact SPGP predictors that consider pseudo-inputs as parameters of the model. Therefore, we have two types of parameters: one is the location of local pseudo-inputs and the other is the hyperparameters of the underlying covariance function. Maximizing the logarithm of the local SPGP marginal likelihood functions using gradient descent with respect to local pseudo-inputs and hyperparameters provides local optimal locations. Specifically, the logarithm of the marginal likelihood of SPLK's $s^{\text{th}}$ local model is 
\begin{eqnarray}
\log(p(\mathbf y_s))=-\frac{1}{2} \log | \mathbf{G}_s|-\frac{1}{2}\mathbf y^T_s \mathbf{G}_s^{-1} \mathbf y_s-\frac {n_s}{2} \log{2\pi},\label{eq:log_lik}
\end{eqnarray}
where $\mathbf{G}_s$ is the same as that of Section~\ref{sub_sec_mean_pred}.

Moreover, we use anti-isotropic squared exponential function as the choice of our local covariance functions, 
\begin{eqnarray}
\mathcal{\phi}(\mathbf{x},\mathbf{x}')=C\exp\big(-(\mathbf{x}-\mathbf{x}')^T\boldsymbol{\Gamma }(\mathbf{x}-\mathbf{x}')\big),
\end{eqnarray}
where $\boldsymbol{\Gamma}$ is a diagonal matrix with length-scale parameters $\gamma_1,\ldots,\gamma_p$ on the diagonal. This covarinace function automatically determines the significance of predictors after training its parameters by minimizing local likelihood function~\eqref{eq:log_lik}.

\section{A simulation study on the performance of SPLK}\label{App_Dasilva_Simulation}
In this section, we conduct a simulation study to further investigate the performance of SPLK comparing to the other competing algorithms in terms of MSE. As mentioned in Section~\ref{Alg_Comp_sec}, when the rates of covariance decay highly vary in different directions (similar to the Dataset Dasilva), SPLK can perform better than the competing algorithms considered in this study. This is because SPLK partitions the domain of data orthogonal to the direction of the fastest rate of covariance decay, which potentially reduces the degree of mismatch on the boundaries compared with the other directions.

To test this claim we generate 10,000 samples from a Gaussian process with  covariance function~\eqref{SEKernel} and highly different length scale parameters $\gamma_1=50$, $\gamma_2=10$, and $\gamma_3=0.001$. To this end, we first generate 10,000 vectors, $\mathbf{x}_i$, uniformly from the cube $[0,5]\times[0,5]\times[0,5]$ and form the covariance matrix $\mathbf{K}_{\mathbf{X}\mathbf{X}}$. Then we draw 10,000 responses, $y_i$, using $\mathbf{K}_{\mathbf{X}\mathbf{X}}$ and add a noise to each response from distribution $\mathcal{N}(0,4)$. Finally, we use 9,000 of these samples for training and 1,000 fo r testing.

\begin{figure}[ht]
\begin{center}
	\begin{subfigure}{0.45\textwidth}
		\includegraphics[height=4cm,width=7.5cm]{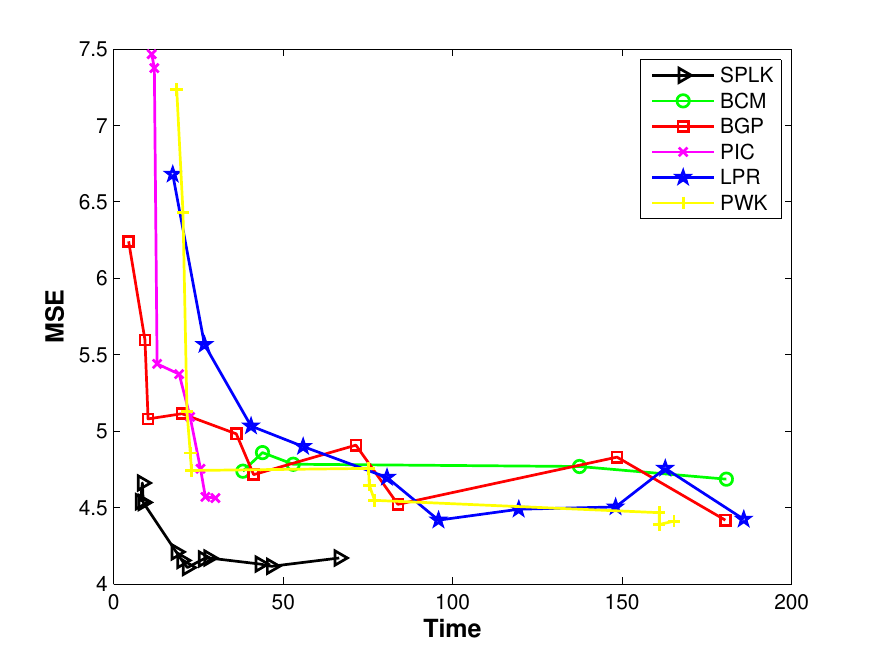}
		\caption {MSE vs. computation time}
\label{fig:TimeVsMSESim}
	\end{subfigure}
	\begin{subfigure}{0.45\textwidth}
		\includegraphics[height=4cm,width=7.5cm]{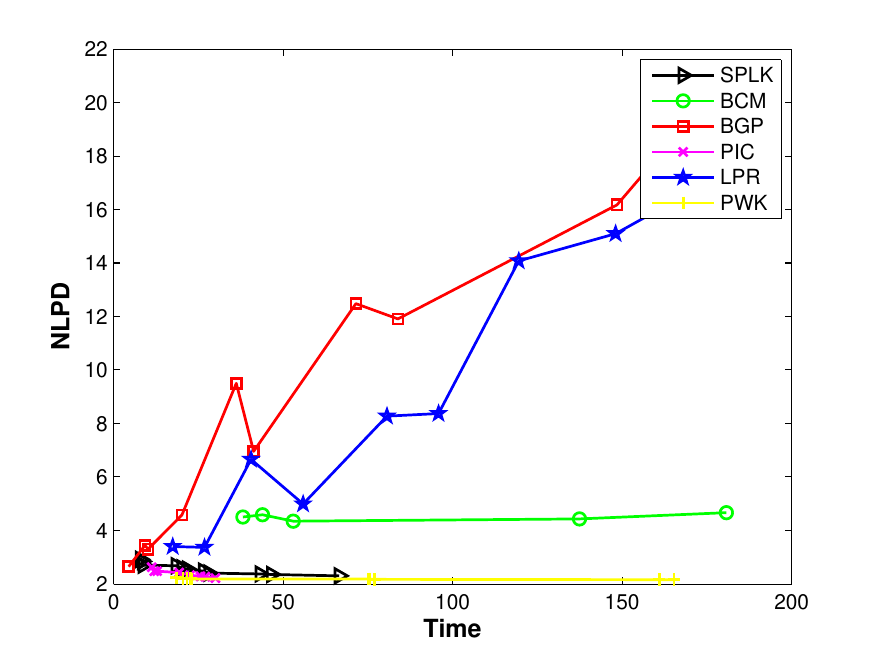}
		\caption {NLPD vs. computation time}
\label{fig:TimeVsNLPDSim}
	\end{subfigure}
\end{center}
\caption{MSE and NLPD versus computation time. For SPLK, $q=3$ and $k \in \{2,4,6,8\}$. The value of parameter $S$ is selected from the set $ \{8,16,32,64,128,256\}.$}
\label{fig:MSENLPDTimeSim}
\end{figure}

For this simulated dataset, SPLK partitions the domain of data from the first direction which has the largest associated length scale parameter. Figures~\ref{fig:TimeVsMSESim} and~\ref{fig:TimeVsNLPDSim} show the performance of all the competing algorithms in terms of MSE and NLPD versus computation time. As expected, due to the designed covariance structure, i.e., highly varying rates of covariance decay, SPLK outperforms the other competing algorithms in terms of MSE, while performs as well as PWK and PIC in terms of NLPD.

\end{appendices}

\end{document}